\documentclass[acmsmall]{acmart}

\usepackage{amsthm}

\usepackage[edges]{forest}
\tikzset{%
    parent/.style =          {align=center,text width=3.5cm,rounded corners=3pt},
    child/.style =           {align=center,text width=2.5cm,rounded corners=3pt},
    grandchild/.style =      {align=center,text width=2cm,rounded corners=3pt},
    greatgrandchild/.style = {align=center,text width=1.5cm,rounded corners=3pt},
    referenceblock/.style =  {align=center,text width=1.5cm,rounded corners=2pt}
}

\usepackage{lscape}
\usepackage[utf8]{inputenc}
\usepackage{multirow}
\usepackage{soul}
\usepackage {lineno}
\usepackage{subfig}
\usepackage{longtable}
\usepackage{xcolor}
\usepackage{enumerate}
\usepackage{mathdots}
\usepackage{cancel}
\usepackage{color}
\usepackage{siunitx}
\usepackage{array}
\usepackage{gensymb}
\usepackage{tabularx}
\usepackage{extarrows}
\usetikzlibrary{fadings}
\usetikzlibrary{patterns}
\usetikzlibrary{shadows.blur}
\usetikzlibrary{shapes}
\usepackage{hyperref}  
\usepackage[normalem]{ulem}
\useunder{\uline}{\ul}{}
\usepackage{tikz}
\def\checkmark{\tikz\fill[scale=0.4](0,.35) -- (.25,0) -- (1,.7) -- (.25,.15) -- cycle;} 

\newtheorem*{definition}{Unified Definition}

\newcommand{\mycomment}[1]{}

\AtBeginDocument{%
  \providecommand\BibTeX{{%
    \normalfont B\kern-0.5em{\scshape i\kern-0.25em b}\kern-0.8em\TeX}}}

\setcopyright{rightsretained}
\copyrightyear{2025}
\acmYear{2025}
\acmDOI{XXXXXXX.XXXXXXX}

\acmJournal{CSUR}
\acmVolume{67}
\acmNumber{1}
\acmArticle{1}
\acmMonth{1}

\acmSubmissionID{CSUR-2023-1001}

\begin{document}

\title{A Survey on Online Aggression: Content Detection and Behavioral Analysis on Social Media}


\author{Swapnil Mane}
\email{mane.1@iitj.ac.in}
\orcid{0000-0002-8234-4557}
\author{Suman Kundu}
\email{suman@iitj.ac.in}
\orcid{0000-0002-7856-4768}
\affiliation{%
  \institution{Indian Institute of Technology Jodhpur}
  \state{Rajasthan}
  \country{India}
  \postcode{342030}
}

\author{Rajesh Sharma}
\affiliation{%
  \institution{Institute of Computer Science, University of Tartu}
  \country{Estonia}
  }
\email{rajesh.sharma@ut.ee}
\orcid{0000-0003-3581-1332}

\renewcommand{\shortauthors}{Mane et al.}

\begin{abstract}
The proliferation of social media has increased cyber-aggressive behavior behind the freedom of speech, posing societal risks from online anonymity to real-world consequences. This article systematically reviews Aggression Content Detection and Behavioral Analysis to address these risks. Content detection is vital for handling explicit aggression, and behavior analysis offers insights into underlying dynamics. The paper analyzes diverse definitions, proposes a unified cyber-aggression definition, and reviews the process of Aggression Content Detection, emphasizing dataset creation, feature extraction, and algorithm development. Additionally, examines Behavioral Analysis studies that explore influencing factors, consequences, and patterns of online aggression. We cross-examine content detection and behavioral analysis, revealing the effectiveness of integrating sociological insights into computational techniques for preventing cyber-aggression. We conclude by identifying research gaps that urge progress in the integrative domain of socio-computational aggressive behavior analysis.
\end{abstract}





\received{20 November 2024}
\received[revised]{5 September 2024}
\received[accepted]{15 December 2024}

\maketitle

\section{Introduction}
\label{Introduction}
In the age of rapid technological advancement, the landscape of global communication has undergone a profound transformation. 
Social media platforms serve a crucial role in shaping human interaction and information dissemination \cite{souza2020, saha2023}. These platforms transcend geographical boundaries, enabling individuals to connect, share experiences, and engage in discussions on a global scale. However, this digital evolution has also given rise to a complex challenge: the prevalence of aggressive behavior and hostile communication within these virtual spaces \cite{bhattacharya2020developing}. 
Online aggression encompasses a diverse range of hostile behaviors, such as cyberbullying, online harassment, and the propagation of offensive and hate speech. This broad spectrum of aggressive behavior manifests online, spanning from subtle forms of verbal aggression to explicit instances of hate speech and threats \cite{mladenovic2021cyber, chatzakou2017mean}.
One of the reasons for this behavior is the shroud of anonymity provided by the digital environment, which has encouraged individuals to express opinions and engage in behavior that might be restrained in traditional face-to-face interactions \cite{bazarova2013managing}. This phenomenon has created a digital environment where aggression manifests itself in diverse and often insidious ways, challenging the very essence of open and constructive communication.

Detecting aggressive content on social media is the most important first step in developing any aggressive activity preventive system. Similarly, behavioral analysis of user's aggressiveness is equally important in order to identify causes, contextual factors, and patterns associated with such behavior. Both issues have a substantial impact on society, hence finding solutions to these problems is essential. It has been reported that aggressive content can exert detrimental psychological effects on both individual users and the broader online community \cite{bouhnik2014whatsapp}. Exposure to aggressive language can evoke feelings of distress, anxiety, and fear, ultimately compromising the sense of safety that should ideally characterize social media interactions. To summarize, the unchecked propagation of aggressive behavior can result in the erosion of the digital environment's potential as a space for meaningful discourse and collective problem-solving \cite{rawat2023modelling}.

Researchers have increasingly turned to computational methods for aggressive content detection \cite{chatzakou2019detecting, mladenovic2021cyber}.
The investigations have been conducted on prominent platforms like Facebook \citep{chen2017aggressivity, kumar2018aggression, kumari2022multi, sharif2022tackling}, Instagram \cite{kumari2022multi}, X formerly known as Twitter \citep{kumar2018aggression, chen2017aggressivity, chatzakou2019detecting, sadiq2021aggression, torregrosa2022mixed, kumari2022multi, gattulli2022cyber}, YouTube \cite{bhattacharya2020developing, kumar2020evaluating, sharif2022tackling}, and From-spring.me \cite{Reynolds2011}. These approaches leverage the power of natural language processing (NLP), machine learning (ML), and deep learning (DL), including transformers to automatically identify and categorize aggressive posts \cite{burnap2015cyber}. By promptly flagging aggressive content, these techniques strive to mitigate the harmful consequences of such behavior and promote a more positive online culture. Simultaneously, researchers recognize that a comprehensive behavioral understanding of users' aggressiveness necessitates a deeper analysis of the underlying user behaviors contributing to these interactions. 
Studies on behavioral analysis of user aggressiveness have been carried out on platforms like X \cite{chatzakou2017mean, aragon2019overview, poiitis2021aggression, torregrosa2022mixed, mane2023you}, Gab.com \cite{mathew2019spread}, and others \cite{borraccino2022problematic, eraslan2019social, balci2015automatic}. A holistic perspective requires exploring the triggers, factors, and patterns associated with aggressive behavior. This nuanced comprehension can inform the development of more effective prevention and intervention strategies, addressing the issue at its roots.
Despite considerable progress in aggressive content detection and behavioral analysis of user aggressiveness, a notable gap persists in integrating these two domains \cite{poiitis2021aggression, terizi2021modeling}. 
This literature review aims to bridge this gap by systematically analyzing prior research on explicitly aggressive content detection methods and behavioral analysis of aggressive users within the realm of social media platforms. By synthesizing these two critical aspects, the review seeks to shed light on a combination of these two aspects which can yield a more comprehensive understanding of online aggression and its underlying dynamics.
The increasing number of publications over time indicates the current significance and relevance of the topic.
In this survey, we have focused on research papers published over the past six years, starting from 2018.

\paragraph{Motivation and contributions}
\label{contributions}
Aggressive behavior analysis on social media has received significant attention that highlights its impact on individuals' mental health and social relationships. This impact manifests in high levels of stress, anxiety, depression, and even suicidal thoughts \citep{Hinduja2018, vazsonyi2012cyberbullying, pantic2012association}. The prevalence of such behavior increases anonymity, social division, social prejudice, and polarization within communities \citep{Hinduja2018, zimmerman2016online}. Particularly targeting vulnerable groups, including children and religious minorities, who face elevated risks of online victimization with severe offline consequences \citep{blachnio2018facebook, livingstone2011risks}. Despite increased scrutiny, there remains a critical gap in research focusing on low-resource languages, such as Punjabi, Marathi, Tamil, Urdu, Italian, Slovenian, Latvian, Turkish, Ukrainian, Vietnamese, etc. The lack of attention to these languages, characterized by limited linguistic resources, further highlights the need for intervention \citep{ranasinghe2021multilingual,saeed2023detection,peng2003language, corcoran2015cyberbullying}. Additionally, the global response to online aggression is reflected in legal regulations such as Germany's NetzDG \cite{lee2022germany}, India's Information Technology Rules\footnote{\url{https://meity.gov.in/writereaddata/files/Intermediary_Guidelines_and_Digital_Media_Ethics_Code_Rules-2021.pdf}}, the UK's Online Harms White Paper\footnote{ \url{https://www.gov.uk/government/consultations/online-harms-white-paper}}, and Singapore's POFMA\footnote{ \url{https://www.mlaw.gov.sg/news/others/pofmb-videos/}}, highlights the increasing recognition by governments worldwide of the urgent need to address and regulate aggressive behavior on digital platforms.
This literature review contributes to the growing field of aggressive behavior analysis, addressing persistent gaps in current research by comprehensively examining approaches and providing potential research directions. The following are the highlights of this work:
\begin{enumerate}
    \item Proposed a unified and comprehensive definition of cyber-aggression by analyzing the diverse range of existing definitions.
    \item Thoroughly examines the entire process of Aggression Content Detection, including dataset creation, feature selection, extraction, and detection algorithm development.
    \item Reviews studies related to the Behavioral Analysis of Aggression, shedding light on the influencing factors, consequences, and patterns associated with cyber-aggressive behavior, thus contributing to a deeper understanding of this phenomenon.
    \item Emphasizes the significance of incorporating sociological findings into automatic computational prevention systems for cyber-aggressive behavior, highlighting the need for a multidisciplinary approach to address this complex issue.
\end{enumerate}

The structure of the paper is as follows. Section \ref{2_definations} analyzed the theoretical aspects of the topic, including exploration of different definitions, while
Section \ref{3_related_work} presents an analysis of a previous survey on the same topic. 
Section \ref{4_methodology} outlines the methodology used for the present systematic literature review. Sections \ref{5_detection} and \ref{6_aggressive_behaviour} review the aggressive content detection approaches and the behavioral analysis studies respectively.
In Section \ref{socio-comptational}, we presented the integration of sociological insights with computational techniques to understand and mitigate online aggression.
Research challenges and directions are highlighted in Section \ref{7_research_gap}. Finally, Section \ref{8_conclusion} provides a comprehensive conclusion to this survey by summarizing the key contributions.

\section{Understanding Cyber-Aggression}
\label{2_definations}
Interactions on social media have given rise to harmful and widespread aggressive behaviors with severe consequences. This section provides a thorough analysis of cyber-aggression, aiming to establish a comprehensive definition for its diverse expressions in virtual environments.
The term ``cyber-aggression'' has been given various interpretations within scholarly discourse.  The initial definitions primarily focused on instances of hostility, harassment, or harm that occur through digital channels  \cite{roy2018ensemble, diaz2020automatic}. While these initial definitions offer a basic understanding, they may unintentionally limit the broad scope of cyber-aggression to only overtly aggressive behaviors. 
In order to offer a thorough comprehension of cyber-aggression, a collection of definitions from scholarly discourse has been compiled. These definitions collectively illustrate various ways in which aggression can be expressed through digital communication. 
Examples include language targeting individuals for harm \cite{anderson2002human}, verbal attacks, abusive language, and derogatory comparisons \cite{roy2018ensemble}, language inciting violence or causing harm \cite{diaz2020automatic}, content encompassing aggression, credible threats, and misinformation \cite{facebook}, text glorifying violent actions and fostering hostility \cite{fortuna2018survey}, communication undermining based on attributes \cite{nobata2016abusive}, and the intent to cause harm through verbal, physical, or psychological means \cite{baron1994human, buss1962psychology}. 
This further include deliberate electronic harm perceived as offensive \cite{UT_ref, kowalski2014bullying}, internet-driven harm \cite{grigg2010cyber, tonglin2018effect, zhao2012reliability}, peer-to-peer online aggression \cite{igiDef}, digital technology fostering hostility \cite{patchin2015measuring}, and digital aggression and harm via online platforms \cite{vollink2015cyberbullying, al2017cyber}.

The definitions of cyber-aggression highlight its complex nature, which encompasses a range of behaviors intended to cause harm, abuse, derogatory remarks, threats, and hostility within the digital domain. Various definitions of cyber-aggression illustrate online aggression in distinct ways, often tailored to specific contexts or applications, with a particular focus on explicit acts of aggression. Nevertheless, it is crucial to acknowledge that contemporary online expressions of aggression often adopt a covert tone, including humor and sarcasm. This highlights the significance of taking into account covert or satirical manifestations of aggression. However, this perspective presents the difficulty of accurately identifying aggression in the context of online interactions. Additionally, other perspectives highlight the power dynamics that are inherent in cyber-aggression \cite{kowalski2013psychological}.
The dimensions of cyber-aggression encompass a range of expressions, spanning from explicitly aggressive behavior to subtler, covert behaviors. In this dimension, \textit{explicit aggression} involves direct and overt harmful behaviors, while \textit{passive-aggressive behavior} employs indirect expressions of hostility. \textit{Microaggressions} are subtle actions that undermine or diminish individuals based on their characteristics. \textit{Exclusionary practices} involve deliberate isolation, and \textit{identity-based attacks} target personal attributes. \textit{Manipulative coercion} employs emotional tactics, \textit{trolling, and provocation} incites reactions, and \textit{revengeful actions} aim at retribution. These dimensions can manifest through the use of hostile, offensive language, insults, and threats intended to harm. The scope of cyber-aggression, therefore, includes behaviors that are both overt and covert, aligning with the broader categories of cyberbullying, hate speech, and trolling.
Based on this comprehensive understanding of cyber-aggression. This study proposed the following unified definition:
\begin{definition}
  Cyber-aggression refers to harmful intentional online behavior, irrespective of whether it is overt or covert. It includes the use of hostile, offensive, or abusive language, insults, threats, and abusive comments intended to cause discomfort, distress, or harm to individuals or communities.
\end{definition}

This definition captures the multidimensional nature of online aggressive behaviors, encompassing both explicit attacks and subtle actions that may result in psychological harm. As demonstrated in Table \ref{tab:comp-analysis}, our definition addresses key parameters often overlooked in other definitions, such as the inclusion of both overt and covert forms of aggression, acknowledgment of psychological impact, and applicability across various online contexts and cultures. Further, we validated the proposed definition with an inter-rater reliability score of 0.78. This was achieved using Krippendorff's alpha metric among 27 diverse annotators.

\begin{table}[h!]
\centering
\caption{Comparative analysis of cyber-aggression definitions across key parameters.}
\label{tab:comp-analysis}
\resizebox{\textwidth}{!}{%
\begin{tabular}{lcccccccc}
\hline
\multicolumn{1}{c}{\textbf{\begin{tabular}[c]{@{}c@{}}Definition \\ Source\end{tabular}}} & \textbf{Intentionality} & \textbf{\begin{tabular}[c]{@{}c@{}}Overt/Covert \\ Inclusion\end{tabular}} & \textbf{\begin{tabular}[c]{@{}c@{}}Psychological \\ Impact\end{tabular}} & \textbf{\begin{tabular}[c]{@{}c@{}}Cultural \\ Sensitivity\end{tabular}} & \textbf{\begin{tabular}[c]{@{}c@{}}Contextual \\ Sensitivity\end{tabular}} & \textbf{\begin{tabular}[c]{@{}c@{}}Platform \\ Specificity\end{tabular}} & \textbf{\begin{tabular}[c]{@{}c@{}}Power \\ Dynamics\end{tabular}} & \textbf{\begin{tabular}[c]{@{}c@{}}Comprehensive \\ Scope\end{tabular}} \\ \hline
Roy et al. \cite{roy2018ensemble} & $\checkmark$ & $\times$ & $\checkmark$ & $\times$ & $\times$ & $\times$ & $\times$ & $\times$ \\ 
Diaz et al. \cite{diaz2020automatic} & $\checkmark$ & $\times$ & $\checkmark$ & $\times$ & $\checkmark$ & $\times$ & $\times$ & $\times$ \\ 
Nobata \& Tetreault. \cite{nobata2016abusive} & $\checkmark$ & $\checkmark$ & $\times$ & $\times$ & $\checkmark$ & $\times$ & $\times$ & $\times$ \\
Baron \& Richardson \cite{baron1994human} & $\checkmark$ & $\times$ & $\checkmark$ & $\times$ & $\times$ & $\times$ & $\times$ & $\times$ \\ 
Smith et al. \cite{smith2008cyberbullying} & $\checkmark$ & $\checkmark$ & $\checkmark$ & $\times$ & $\times$ & $\times$ & $\times$ & $\times$ \\ 
Grigg \cite{grigg2010cyber} & $\checkmark$ & $\times$ & $\checkmark$ & $\times$ & $\checkmark$ & $\times$ & $\times$ & $\times$ \\ 
Patchin \& Hinduja \cite{patchin2015measuring} & $\checkmark$ & $\times$ & $\checkmark$ & $\times$ & $\checkmark$ & $\checkmark$ & $\checkmark$ & $\times$ \\ 
Vollink \cite{vollink2015cyberbullying} & $\checkmark$ & $\times$ & $\checkmark$ & $\times$ & $\checkmark$ & $\checkmark$ & $\checkmark$ & $\times$ \\ 
Hadeel \cite{al2017cyber} & $\checkmark$ & $\times$ & $\checkmark$ & $\checkmark$ & $\times$ & $\times$ & $\checkmark$ & $\times$ \\ 
\textbf{Ours} & $\checkmark$ & $\checkmark$ & $\checkmark$ & $\checkmark$ & $\checkmark$ & $\checkmark$ & $\checkmark$ & $\checkmark$ \\ \hline
\end{tabular}%
}
\end{table}

\section{Related Work}
\label{3_related_work}
This section reviews existing literature surveys that have explored various aspects of aggressive content detection and behavioral analysis of user aggressiveness, particularly within social media platforms. 
\citet{langham2018classification} reviews work on aggression detection and underscores the necessity of analyzing aggressive language and the homogeneity of users, rather than focusing solely on hate speech. Similarly, \citet{mladenovic2021cyber} provides an in-depth examination of the detection of objectionable content, including aggression, bullying, and grooming. They analyze different datasets, feature types, and detection methods.  
Expanding on the relationship between online platforms and aggression, \citet{anderson2001effects} investigate the impact of violent video games. They reveal a correlation between exposure to such games and increased aggressive behavior, thoughts, and affect. A comprehensive analysis of cyber-based peer-to-peer aggression is offered in \cite{corcoran2015cyberbullying}. It compares it with traditional bullying and highlights the challenges of applying conventional bullying criteria in the digital realm. 
The survey by \citet{martins2020effects} further explores the influence of media use on social aggression in children and adolescents, contributing valuable insights to the field of media psychology.

This paper offers an integrative analysis that bridges the gap between content detection and the behavioral analysis of cyber-aggression on social media platforms. Unlike prior surveys that focus solely on either the detection of aggressive content or the behavioral aspects of user aggressiveness, our survey combines both dimensions. This approach provides a more holistic understanding of the phenomenon. Additionally, we have conducted a thorough review of individual dimensions as well. We introduce a unified definition of cyber-aggression and emphasize the importance of incorporating sociological insights into computational models. This integrated perspective not only enhances the depth of analysis but also proposes new directions for research and practical applications in mitigating cyber-aggression.

\section{Methodology for Systematic Literature Review}
\label{4_methodology}
In the examination of online aggression detection and behavioral analysis, we conducted a systematic literature review following established guidelines \cite{Kitchenham2007, Moher2009}. We queried a diverse range of literature sources to ensure comprehensive coverage, including ACM Digital Library, IEEE Xplore, SpringerLink, ScienceDirect, Wiley Online Library, ACL Anthology, PubMed Central, DBLP Computer Science Bibliography, Semantic Scholar, and Google Scholar. These sources encompass repositories specializing in computer science, software engineering, sociology, and interdisciplinary research. Our search strategy employed a multi-phase approach: in phase 1, we conducted an automated search using a carefully crafted search string and limited it from 2018 to 2024. Phase 2 involved forward and backward snowballing to identify additional relevant studies that may have been missed in the initial search \cite{Wohlin2014}.
We applied study selection criteria to ensure the relevance and quality of the studies included in our review. Inclusion criteria required that studies be published in English between January 2018 and February 2024 and present computational models for identifying (or analyzing) online aggression using data from social media platforms or online communities. Exclusion criteria led to the exclusion of duplicate publications or those with substantial overlap with included studies, as well as studies lacking clear specifications of the target population, platform, or data source. We also excluded studies focused primarily on cyberbullying or general emotion detection without specific emphasis on aggression, purely theoretical works without empirical validation, and studies centered solely on offline aggression or non-digital contexts.

The screening and selection process involved three stages. First, we conducted a title and abstract screening to assess each study against the inclusion and exclusion criteria. Studies that passed this initial stage then underwent a full-text review, where we performed an evaluation of the full texts. 
For data extraction and synthesis, we created a standardized form to systematically extract study metadata, objectives, methods, models, performance metrics, key findings, limitations, and future research directions. To mitigate potential biases, we employed several strategies: all reviewers participated in both the screening, and preprints were included to ensure comprehensive coverage.
This methodology allows us to present a thorough, unbiased, and replicable review of the literature.

\begin{figure}[!t]
\centering
\includegraphics[scale=0.45]{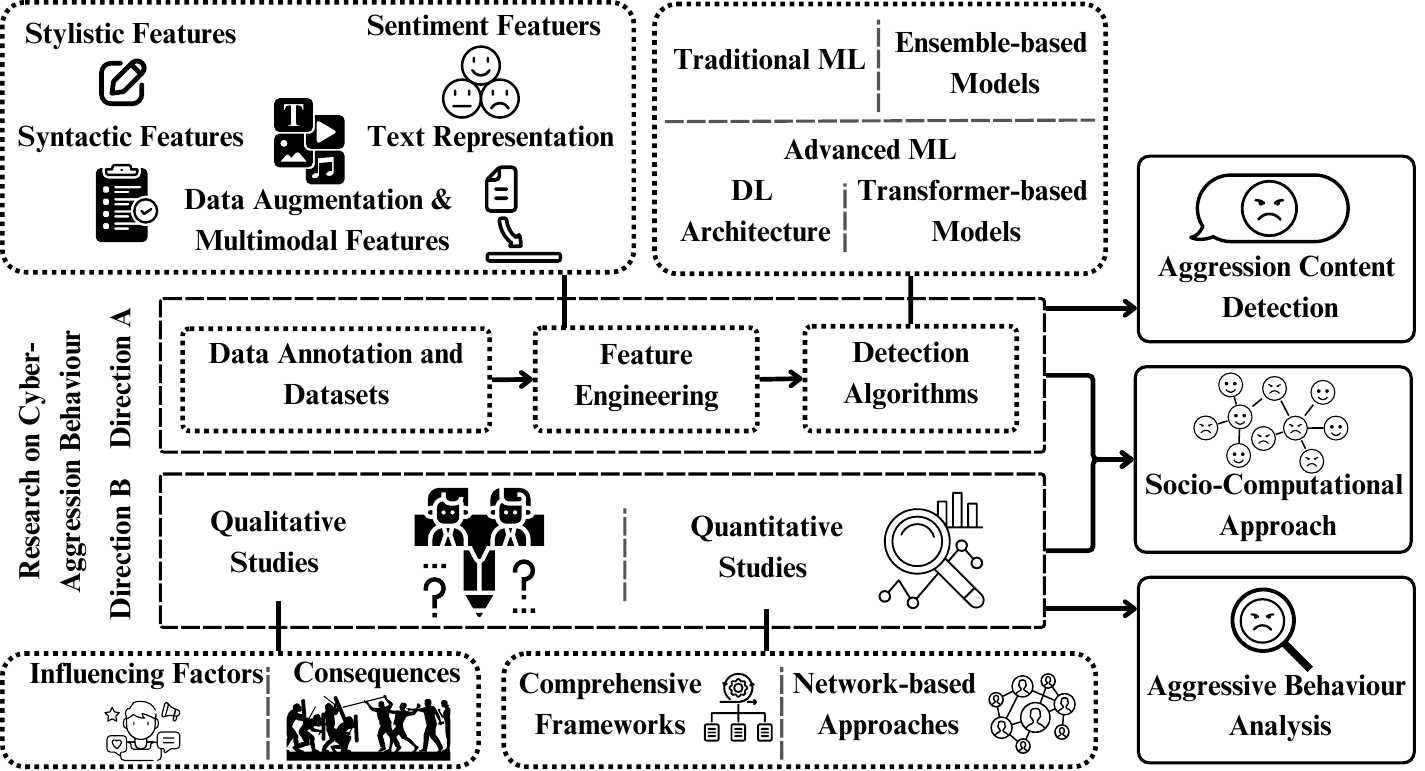}
\caption{Overview of our systematic literature review on cyber-aggression, introducing a new socio-computational approach to address the problem more effectively.}
\label{fig:methodology}
\vspace*{-1.3\baselineskip}
\end{figure}

Further, our review distinguished two primary research directions (as shown in Figure \ref{fig:methodology}): (A) Aggression content detection, rooted in computer science, and Aggressive Behavior Analysis, drawing from both qualitative and quantitative methods in sociology and computer science. The aggression content detection studies are categorized based on their computational approaches to identifying aggressive content online, reflecting their primary contributions to the field.
Data Annotation and Datasets involve research focused on developing annotation schemes and creating datasets. Feature engineering methods are classified based on the features used for aggression detection, including stylistic features, syntactic features, sentiment features, text representation, and data augmentation and multimodal features. Detection algorithms are categorized into traditional machine learning approaches, advanced machine learning methods, which include deep learning architectures and transformer-based models, and ensemble-based models, which combine multiple algorithms to enhance accuracy and robustness in aggression detection.
(B) Behavioral analysis approaches are categorized into qualitative and quantitative studies: Qualitative studies involve research employing interpretive methods to understand the nature of online aggression, including influencing factors and consequences. Quantitative Studies involve research using statistical and computational methods to model and predict aggressive behavior, including comprehensive frameworks and models and network- and user-based approaches.

\section{Aggression Content Detection}
\label{5_detection}
In this section, we provide an overview of the research papers that have focused on the domain of aggression detection.  
The following subsections provide in-depth analytical insights by delving deeply into the annotation, available datasets, effective features, and algorithms used for aggression detection.

\subsection{Data Annotation and Datasets}
This section explores the annotation strategies employed in creating ground truth data, as well as the available datasets across various online social media platforms and languages.

\subsubsection{Annotation}

While some studies employ a straightforward binary classification of aggressive/non-aggressive content \citep{chen2017aggressivity, sadiq2021aggression, gattulli2022cyber, sharif2022tackling, Aragn2020OverviewOM}, others have developed more nuanced multi-level approaches. These approaches differentiate between overt, covert, and Non-aggression \citep{kumaretal2022comma, kumar2018aggression, bhattacharya2020developing, rawat2023modelling}, or categorize content based on aggression intensity levels \citep{kumari2019aggressive, kumari2022multi}. To capture finer details of aggressive behavior, researchers have introduced subtags and attributes such as physical threat, sexual threat, and discursive effects like attack, defense, counterspeech, and gaslighting \citep{kumaretal2022comma, kumar2018aggression}. Some studies have even extended their focus to user-level annotations, categorizing individuals as bullies, aggressors, normal users, or spammers \citep{chatzakou2019detecting}.
Ensuring reliability in annotation is crucial for developing robust datasets. Researchers employ various metrics to measure inter-rater agreement (IAA), including Cohen's kappa for two annotators \citep{SaneCorpusCA, sharif2022tackling, rawat2023modelling} and Fleiss' kappa or Krippendorff's alpha for multiple annotators \citep{chatzakou2019detecting, Srivastava2020AMV, kumaretal2022comma, kumar2018aggression, bhattacharya2020developing}.
As shown in Table \ref{tab:dataset}, IAA scores in these studies range from moderate to almost perfect, indicating the overall robustness of the annotation process. Some researchers, like \cite{kumaretal2022comma}, have implemented multi-phase annotation processes, demonstrating substantial improvement in IAA scores after discussions and refinements.
Recognizing the potential influence of annotators' backgrounds on the annotation process \citep{benderfriedman2018data, Binns2017LikeTL}, researchers have implemented strategies to mitigate bias. These include selecting annotators from diverse racial, residential, and religious backgrounds \citep{sharif2022tackling}, ensuring balanced demographics among annotators \citep{kumari2022multi}, and employing multiple annotators to reduce individual biases \citep{gattulli2022cyber, chatzakou2019detecting, rawat2023modelling}. Despite these efforts, challenges persist in annotating subjective aspects of aggression, such as discursive effects, often necessitating revisions to annotation guidelines \citep{chatzakou2019detecting, kumaretal2022comma}. 
The subjective nature of aggression makes consistent annotation challenging, even for human annotators. This subjectivity can lead to disagreements among annotators and potential biases in the resulting datasets.

\begin{figure}[]
\centering
\includegraphics[scale=0.40]{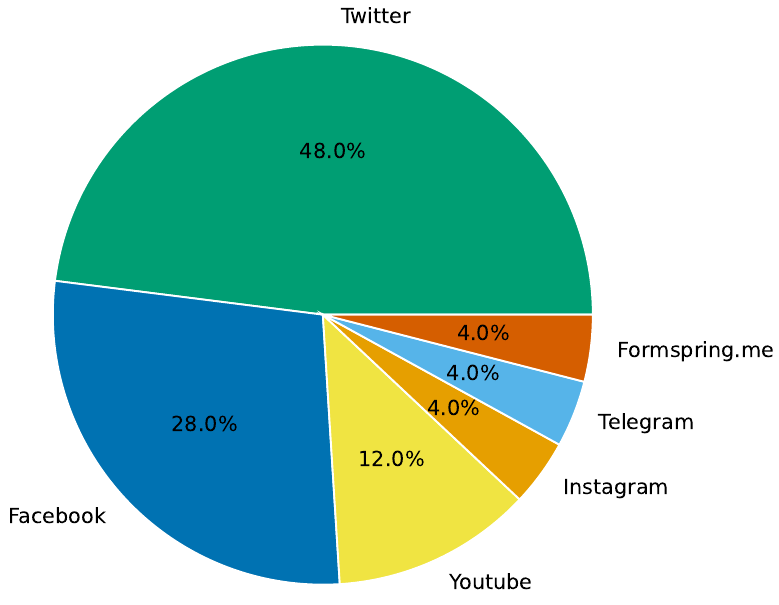}
\caption{Distribution of aggression detection datasets across OSM platforms.}
\label{platform_proportion}
\vspace*{-1\baselineskip}
\end{figure}

\subsubsection{Datasets}
\begin{figure}[h!]
    \centering
    \subfloat[\centering Distribution of language-specific datasets in aggression detection.]{{\includegraphics[scale=0.35]{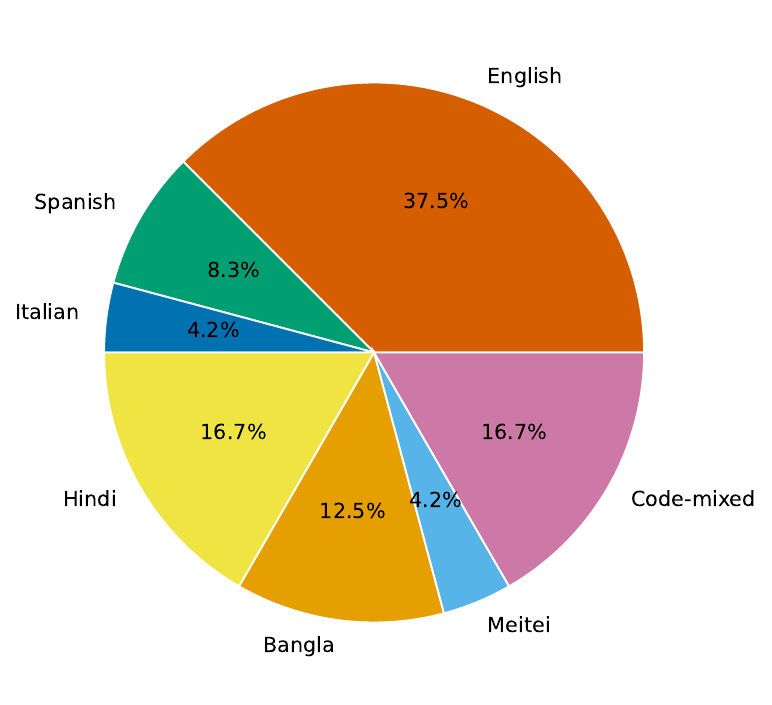} }}%
    \qquad
    \subfloat[\centering {Distribution of data instances in language-specific datasets.}]{{\includegraphics[scale=0.35]{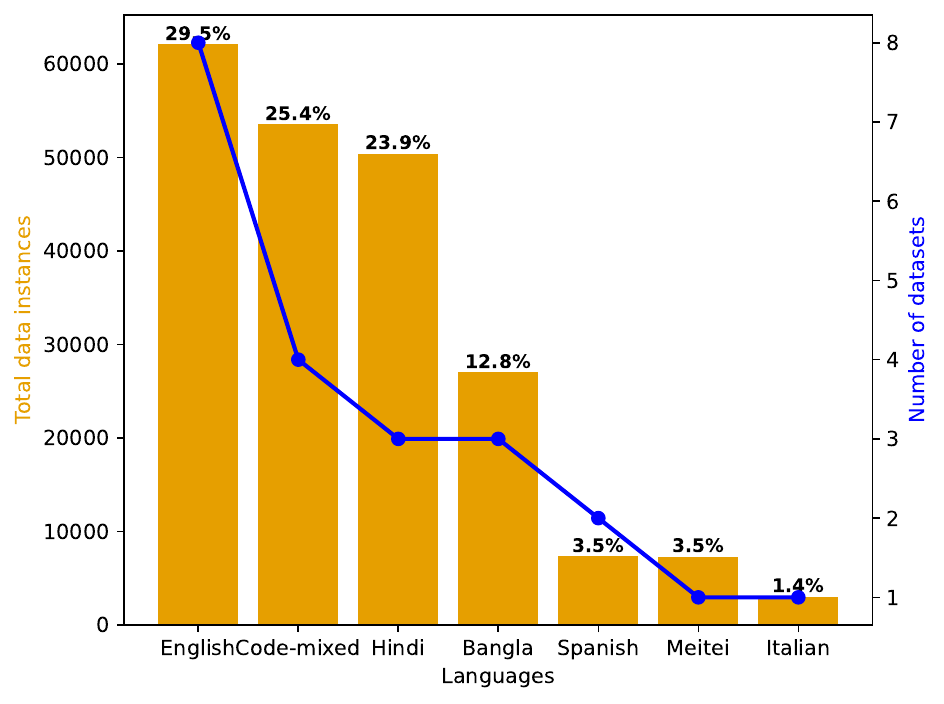} }}%
    \caption{Analysis of language-specific datasets.}%
    \label{fig:lang_proportion}%
    \vspace*{-1.3\baselineskip}
\end{figure}

The landscape of online aggression research datasets is diverse and evolving, encompassing multiple platforms and languages. Table \ref{tab:dataset} provides a comprehensive overview of key datasets in this field, highlighting their characteristics, limitations, and availability.
Online Social Media (OSM) platforms serve as primary data sources, with \cite{chen2017aggressivity, rawat2023modelling, kumar2018aggression, bhattacharya2020developing, kumaretal2022comma, Kumari2021MultimodalCD, gattulli2022cyber, chatzakou2019detecting, sadiq2021aggression, mane2023you}, Facebook \cite{kumar2018aggression, bhattacharya2020developing, kumaretal2022comma, sharif2022tackling, Kumari2021MultimodalCD}, and YouTube \cite{bhattacharya2020developing, kumaretal2022comma, sharif2022tackling} being the most prevalent (Figure \ref{platform_proportion}). Notable datasets include those from Kumar et al. (2018) \cite{kumar2018aggression}, Chatzakou et al. (2019) \cite{chatzakou2019detecting}, and Bhattacharya et al. (2020) \cite{bhattacharya2020developing}, which have been widely adopted in subsequent research.
While English remains the dominant language (Figure \ref{fig:lang_proportion}a), there is a growing focus on low-resource and code-mixed languages. For instance, Kumar et al. (2022) \cite{kumaretal2022comma} developed a multilingual dataset covering English, Hindi, Bangla, Meitei, and code-mixed variants. This trend toward linguistic diversity is crucial for developing more inclusive and generalizable aggression detection models.
Data collection methodologies vary among researchers, with many employing API-based collection methods \citep{rawat2023modelling, kumaretal2022comma, gattulli2022cyber, chatzakou2019detecting, mane2023you} or web crawlers \citep{kumaretal2022comma, sharif2022tackling}. Researchers often tailor their collection strategies to specific objectives, such as Rawat et al. (2023) \cite{rawat2023modelling} focusing on aggression during Indian general elections.
These datasets exhibit key characteristics that are crucial for aggression detection research. The sizes vary significantly, ranging from a few thousand to over 50,000 instances, with different levels of class balance between aggressive and non-aggressive content. Annotation schemes typically involve binary (aggressive/non-aggressive) or ternary (including covert aggression) classifications, with IAA scores varying widely from 0.27 to 0.93, highlighting the subjective nature of aggression annotation. Some datasets, like Kumari et al. (2022) \cite{Kumari2021MultimodalCD}, address the multimodal nature of online communication by incorporating image data alongside text. As shown in Table \ref{tab:dataset}, many of these datasets are publicly available, enabling reproducibility and facilitating comparative studies, though some remain inaccessible due to privacy concerns or platform restrictions.
Existing datasets have limitations that impact their effectiveness in aggression detection research. There remains a significant language bias, with a predominant focus on English and a few other major languages, limiting the applicability of models to diverse linguistic contexts. Additionally, most datasets are platform-specific, which can restrict the generalizability of models trained on them to other platforms. The rapid evolution of online language and behavior also poses a challenge, as datasets can quickly become outdated, reducing their relevance over time. Furthermore, inconsistencies in annotation schemes and IAA scores across datasets complicate cross-dataset comparisons, making it difficult to benchmark and evaluate models comprehensively.

While our analysis reveals a growing diversity in dataset languages and platforms, the field still faces challenges in establishing truly comprehensive, multilingual datasets for online aggression detection. The concept of ``one dataset for all models'' is appealing but presents significant technical and logistical challenges. These include addressing linguistic and cultural differences, ensuring consistent annotation, navigating ethical concerns related to data collection, and managing the scalability of maintaining and updating the dataset.
Despite these challenges, efforts towards more inclusive datasets are crucial. A study by \citet{ousidhoum-etal-2020-comparative} on multilingual hate speech detection demonstrates the potential of cross-lingual approaches. Their dataset, covering six languages and various types of toxic speech, serves as a step towards more comprehensive resources. Similarly, the HASOC dataset \cite{mandl2019overview} provides a valuable resource for hate speech and offensive content identification in Indo-European languages.
These initiatives show that creating more inclusive datasets is possible. However, significant work is still needed to achieve a truly unified global dataset for online aggression detection.
The size distribution across languages (Figure \ref{fig:lang_proportion}b) reveals that code-mixed datasets are comparable in size to English datasets, aligning with observations of increased aggressive behavior in code-mixed contexts.

\begin{table}[h!]
\centering
\caption{Presents a comprehensive overview of various datasets, including the year of creation, data source, languages covered, total data instances, non-aggressive and aggressive instances, number of classes, IAA if available, dataset accessibility status, and corresponding citations.}
\label{tab:dataset}
\resizebox{\textwidth}{!}{%
\begin{tabular}{rrlllllllcc}
\hline
\textbf{Ref.} & \multicolumn{1}{c}{\textbf{Year}} & \multicolumn{1}{c}{\textbf{Source}} & \multicolumn{1}{c}{\textbf{Language}} & \multicolumn{1}{c}{\textbf{Total}} & \multicolumn{1}{c}{\textbf{Non-Aggressive}} & \multicolumn{1}{c}{\textbf{Aggressive}} & \multicolumn{1}{c}{\textbf{Classes}} & \multicolumn{1}{c}{\textbf{IAA (0-1)}} & \textbf{Avail.?} & \textbf{Cites} \\ \hline
\cite{chen2017aggressivity} & 2017 & Tw and Fb & En & \multicolumn{1}{c}{-} & \multicolumn{1}{c}{-} & \multicolumn{1}{c}{-} & NAG, AG & \multicolumn{1}{c}{-} & N & $\geq8$ \\ \hline
\cite{kumar2018aggression} & 2018 & Tw and Fb & En, Hi & \begin{tabular}[c]{@{}l@{}}Tw: 18000\\ Fb: 21000\end{tabular} & \begin{tabular}[c]{@{}l@{}}Tw: 8982\\ FB: 8946\end{tabular} & \begin{tabular}[c]{@{}l@{}}Tw: \{CAG: 7938, \\ OAG: 1080\}\\ Fb: \{CAG: 6279, \\ OAG: 5775\}\end{tabular} & \begin{tabular}[c]{@{}l@{}}OAG, CAG,\\ NAG\end{tabular} & KA: 0.49 & Y & $\geq157$ \\ \hline
\cite{chatzakou2019detecting} & 2019 & Tw-user & En & 4954 & 3562 & \begin{tabular}[c]{@{}l@{}}AG: 59, Bully: 24, \\ Spam: 1300\end{tabular} & \begin{tabular}[c]{@{}l@{}}AG, Bully, \\ Spam, NAG\end{tabular} & FK: 0.67 & Y & $\geq97$ \\ \hline
\cite{bhattacharya2020developing} & 2020 & \begin{tabular}[c]{@{}l@{}}YT, Fb, \\ and Tw\end{tabular} & \begin{tabular}[c]{@{}l@{}}En, Hi, \\ and Ba\end{tabular} & 20000 & \begin{tabular}[c]{@{}l@{}}Ba : 3312\\ Hi: 3148\\ En: 4901\end{tabular} & \begin{tabular}[c]{@{}l@{}}Ba: \{CAG: 1341, \\ OAG: 1318\}\\ Hi: \{CAG: 1231, \\ OAG: 1802\}\\ En: \{CAG: 794, \\ OAG: 834\}\end{tabular} & \begin{tabular}[c]{@{}l@{}}OAG, CAG,\\ NAG\end{tabular} & KA: 0.75 & Y & $\geq71$ \\ \hline
\cite{Srivastava2020AMV} & 2020 & Tw & En & 1001 & 597 & \begin{tabular}[c]{@{}l@{}}CAG: 140, \\ OAG: 264\end{tabular} & \begin{tabular}[c]{@{}l@{}}OAG, CAG,\\ NAG\end{tabular} & FK: 0.64 & Y & $\geq1$ \\ \hline
\cite{sadiq2021aggression} & 2021 & Tw & En & 20001 & 12,179 & 7822 & AG, NAG & \multicolumn{1}{c}{-} & Y & $\geq89$ \\ \hline
\cite{torregrosa2022mixed} & 2022 & Tw & Es & \multicolumn{1}{c}{-} & \multicolumn{1}{c}{-} & \multicolumn{1}{c}{-} &  & \multicolumn{1}{c}{-} & N & $\geq4$ \\ \hline
\cite{gattulli2022cyber} & 2022 & Tw & It & 3028 & 1514 & 1514 & AG, NAG & \multicolumn{1}{c}{-} & N & $\geq2$ \\ \hline 
\cite{sharif2022tackling} & 2022 & Fb and YT & Ba & 14158 & 7351 & 6807 & AG, NAG & CK: 0.87 & Y & $\geq26$ \\ \hline
\cite{Aragn2020OverviewOM} & 2022 & Tw & Es & 7332 & 5222 & 2110 & AG, NAG & \multicolumn{1}{c}{-} & N & $\geq25$ \\ \hline
\cite{kumaretal2022comma} & 2022 & \begin{tabular}[c]{@{}l@{}}Tw, Fb\\ YT, and Tel\end{tabular} & \begin{tabular}[c]{@{}l@{}}En, Hi, Ba, \\ Mn, and Cm\end{tabular} & 59152 & \begin{tabular}[c]{@{}l@{}}Mn: 3221\\ Ba: 5875\\ Hi: 3684\\ En: 5798\\ Cm: 652\end{tabular} & \begin{tabular}[c]{@{}l@{}}OAG: \{Mn: 7677, \\ Ba: 7523, Hi: 9866,\\ En: 2959, Cm: 1176\},\\ CAG: \{Mn: 4103, \\ Ba: 2360, Hi: 2482, \\ En: 1299, Cm: 477\}\end{tabular} & \begin{tabular}[c]{@{}l@{}}NAG, OAG, \\ CAG\end{tabular} & \begin{tabular}[c]{@{}l@{}}KA: \\ 0.27 - 0.93\end{tabular} & Y & $\geq14$ \\ \hline
\cite{Kumari2021MultimodalCD} & 2022 & \begin{tabular}[c]{@{}l@{}}Tw, Fb, \\ and Insta\end{tabular} & En with Img & 3600 & 1804 & \begin{tabular}[c]{@{}l@{}}MAG: 1327, \\ HAG: 469\end{tabular} & \begin{tabular}[c]{@{}l@{}}HAG, MAG, \\ NAG\end{tabular} & - & N & $\geq6$ \\ \hline
\cite{mane2023you} & 2023 & Tw & En & 6000 & 3398 & 2602 & AG, NAG & FK: 0.78 & Y & $\geq0$ \\ \hline
\cite{rawat2023modelling} & 2023 & Tw & Cm-Hi & 2000 & 992 & CAG: 519, OAG: 489 & \begin{tabular}[c]{@{}l@{}}OAG, CAG,\\ NAG\end{tabular} & CK: 0.76 & Y & $\geq0$ \\ \hline
\end{tabular}%
}
\end{table}

\subsection {Feature Engineering}


The identification of appropriate features is crucial in the advancement of machine learning algorithms in the field of aggression detection. Feature engineering, which involves identifying and leveraging specific properties, is critical to enhancing the performance of aggression detection models. This section reviews the various features utilized by researchers in the field. Researchers have utilized many features to handle nuances in text, even when dealing with low-resource languages. 
They have also used a combination of different features to capture the multidimensional aspects of a text. A detailed breakdown of features used for aggression detection is provided in Table \ref{tab:detection_table}.

\subsubsection{Stylistic Features}
Stylistic features capture the nuances of writing style, expression, and tone that often characterize aggressive content.
Research has consistently shown that aggressive texts tend to exhibit distinct stylistic patterns. Several studies have observed that such content often comprises shorter, more fragmented sentences \citep{herodotou2021catching, pascucci2020role, golem2018combining}. This linguistic brevity is frequently accompanied by simpler, more repetitive language, as evidenced by metrics such as the type-token ratio (TTR) \citep{pascucci2020role}. The TTR, which measures lexical diversity by comparing unique words to total words, typically indicates lower diversity in aggressive texts. Additionally, word length distribution analyses reveal a prevalence of short, simple words in aggressive content \citep{herodotou2021catching, singh2018aggression, pascucci2020role, maitra2018k}. \textit{While these features provide valuable indicators, it is important to note that brevity and simplicity are not exclusive to aggressive content, necessitating careful consideration of context in their interpretation.}
Punctuation and formatting elements have emerged as potent stylistic markers of aggression. Excessive use of exclamation marks \citep{risch2018aggression}, asterisks \citep{risch2018aggression}, capitalization \cite{singh2018aggression, maitra2018k, fortuna2018merging, ramiandrisoa2018irit, golem2018combining}, and question marks \citep{herodotou2021catching, huerta2022verbal, gattulli2022cyber, singh2018aggression, SaneCorpusCA} often correlates with aggression.\textit{ However, the interpretation of these elements can be highly context-dependent and may vary across different online platforms and communities.}
Readability indices, such as the Gunning Fog Index and Flesch Reading Ease, further support the observation that aggressive content tends to be less complex and more straightforward \citep{pascucci2020role, ramiandrisoa2018irit}. The presence of figurative language and metaphors has also been noted as expressive of emotional states, including aggression \citep{pascucci2020role}. Moreover, the occurrence of swear words \cite{ramiandrisoa2018irit, golem2018combining} and abusive/aggressive tokens \citep{singh2018aggression, gomez2018machine, gattulli2022cyber, si2019aggression, Mundra2021EvaluationOT} serves as a strong indicator of aggressive intent. \textit{However, the evolving nature of online language and the creative use of euphemisms to evade detection pose ongoing challenges in relying solely on these linguistic markers.}
Beyond traditional linguistic features, researchers have explored the inclusion of platform-specific elements such as URLs \cite{singh2018aggression, herodotou2021catching}, hashtags \cite{singh2018aggression, SaneCorpusCA, herodotou2021catching, maitra2018k, ramiandrisoa2020aggression}, mentions \cite{SaneCorpusCA, herodotou2021catching}, and numbers \cite{singh2018aggression, golem2018combining} in their analyses. Integrating these elements with linguistic and contextual features has contributed to developing more robust machine learning models for aggression detection. \textit{However, the platform-specific nature of these features may limit their generalizability across different online environments.}
Researchers have often combined multiple stylistic features to enhance the nuance of linguistic style analysis. For instance, \cite{herodotou2021catching} incorporated metrics such as mean word count per sentence and mean word length into their framework. Another innovative approach leveraged COGIT, Expert System's semantic intelligence software, for word-sense disambiguation \cite{pascucci2020role}, adding depth to the analysis of aggressive content.

\subsubsection{Syntactic Features}
Syntactic features focus on the structural aspects of language, providing insights into how sentences are constructed to convey aggression. 
A fundamental approach in syntactic analysis is the examination of parts of speech (POS) distribution. Several studies have demonstrated the efficacy of analyzing the frequency and patterns of nouns, verbs, adjectives, and named entities in identifying aggressive content \citep{huerta2022verbal, herodotou2021catching, Iqbal2019UsingCL, gomez2018machine, fortuna2018merging, si2019aggression, Tommasel2019AnES, ramiandrisoa2018irit, golem2018combining, pascucci2020role}.
For instance, a higher occurrence of aggressive verbs or emotionally charged adjectives can serve as a strong indicator of aggressive language. \textit{However, it is important to note that relying solely on POS distribution may overlook context-dependent aggression, where seemingly neutral words are used aggressively.}
N-gram language models have emerged as a powerful tool for capturing more nuanced linguistic patterns. By analyzing combinations of words or characters, these models provide a comprehensive understanding of the text's underlying structure and potential aggressive content \citep{Iqbal2019UsingCL, gomez2018machine, SaneCorpusCA, Baruah2020AggressionII, Tanase2020DetectingAI, pascucci2020role, risch2018aggression}. \textit{The strength of n-gram models lies in their ability to capture local context, but they may struggle with long-range dependencies or idiomatic expressions of aggression.}
Researchers have also recognized the importance of formatting features in aggression detection. The presence of parentheses, emoticons, or emojis can offer valuable cues about emotional expression and potential aggression \citep{huerta2022verbal, si2019aggression, ramiandrisoa2018irit, ramiandrisoa2020aggression}. \textit{While these features can enhance detection accuracy, their interpretation can be highly context-dependent and culturally variable, necessitating careful consideration in model development.}
The Linguistic Inquiry and Word Count (LIWC2007) tool has been adapted to focus on specific word categories associated with emotions and self-references, providing a more nuanced exploration of the emotional context within aggressive texts \citep{Iqbal2019UsingCL, Samghabadi2020AggressionAM}. 
Emotional trait analysis has been incorporated into several studies, focusing on classifying discrete emotions such as anger, disgust, and fear \citep{huerta2022verbal, oravsan2018aggressive, ramiandrisoa2018irit, khan2022aggression}. This granular approach to emotion classification enhances the understanding of the affective aspects of aggression in texts. \textit{However, the complex and often ambiguous nature of online communication can pose challenges in accurately categorizing emotions.}
The role of negation in modulating the expression of aggression has been acknowledged in the literature \cite{ramiandrisoa2018irit}. Identifying and correctly interpreting negation is crucial for accurate aggression detection, as it can significantly alter the meaning and intent of a statement. Nevertheless, the subtle ways in which negation can be used in online discourse may sometimes elude current detection methods.
To further refine aggression detection models, researchers have introduced additional features such as affective and dimensional attributes (e.g., valence, pleasantness, and attention) \cite{huerta2022verbal}. \textit{These features aim to capture more subtle aspects of aggressive language but may require extensive validation across different contexts and cultures.}
Some studies incorporate gender-related information into feature vectors. \citet{waseem2016you} introduced gender probability, calculated using a Twitter-based lexicon \cite{sap2014developing}, which was then transformed into binary gender categories. \textit{While this approach may offer insights into gender-related patterns of aggression, it raises important ethical considerations regarding privacy and potential bias in aggression detection systems.}

\subsubsection{Sentiment Features}
Sentiment features play a crucial role in distinguishing the emotional tone of content, determining whether it conveys positive, negative, or neutral sentiments \cite{Datta2020SpyderAD, huerta2022verbal, herodotou2021catching, gattulli2022cyber, Iqbal2019UsingCL, maitra2018k, Tommasel2019AnES, oravsan2018aggressive, fortuna2018merging, si2019aggression, agbaje2022neural, ramiandrisoa2018irit, tommasel2018textual, golem2018combining}. Sentiment analysis was used to obtain fine-grained sentiment distributions for textual comments \cite{Samghabadi2020AggressionAM}. They employed sentiment distribution mean and standard deviation for every message as feature vectors for aggression detection. Several researchers combined sentiment features with other textual qualities. Researchers have also explored the integration of sentiment features with other textual attributes to enhance the accuracy of aggression detection. For instance,  Emoti-KATE \cite{maitra2018k}, a K-competitive autoencoder, combines sentiment scores with other features such as word count, capitalization, and hashtag analysis. A significant limitation observed in the literature is the language-dependent nature of many sentiment analysis techniques. For example, the approach presented by \citet{Samghabadi2020AggressionAM} was effective for English language corpora, but it encountered difficulties when applied to Hindi texts. \textit{This underscores the importance of considering linguistic and cultural nuances across different languages and communities in aggression detection.}

\subsubsection{Text Representation}
Traditional text representation methods, such as one-hot encoding and TF-IDF, have provided foundational approaches to capturing textual information. One-hot encoding, which transforms words into binary vectors, has proven effective in identifying isolated words or phrases indicative of aggression \citep{Kumari2021BilingualCD, srivastava2018identifying, madisettyaggression, Raiyani2018FullyCN, kumari2020ai}. Similarly, TF-IDF has been widely employed to highlight distinctive, aggressive terms within specific texts by assigning weights based on term frequency and inverse document frequency \citep{Baruah2020AggressionII, Datta2020SpyderAD, Chen2018VerbalAD, Tanase2020DetectingAI, tommasel2018textual, Shrivastava2021EnhancingAD}.\textit{ While these methods offer simplicity and interpretability, they often fall short in capturing the contextual nuances and semantic relationships crucial for detecting subtle forms of aggression.}
Word2Vec \citep{mikolov2013efficient}, GloVe \citep{pennington2014glove}, and FastText \citep{bojanowski2017enriching} have been extensively utilized to capture semantic information and recognize subtle cues in aggressive language. Word2Vec, for instance, has demonstrated effectiveness in identifying aggressive language that shares semantic characteristics with known aggressive terms, even when not explicitly labeled \citep{Tommasel2019AnES, khan2022aggression, sharif2022tackling, ramiandrisoa2020aggression, raman2022hate}. GloVe embeddings have provided valuable contextual understanding, enabling the capture of nuances in aggression by considering word relationships within text segments \citep{Tommasel2019AnES, oravsan2018aggressive, roy2018ensemble, orabi2018cyber, tommasel2018textual, golem2018combining, patwa2021hater, Khandelwal2020AUS, kumari2022multi}.
FastText, with its ability to capture subword information, has shown particular promise in handling the colloquial and often misspelled language characteristic of online aggression \citep{modha2022empirical, Srivastava2019DetectingAA, galery2018aggression, modha2018filtering, aroyehun2018aggression, kumari2020ai, Khandelwal2020AUS, risch2018aggression, mane2023you}. Its effectiveness has been further enhanced when combined with specialized lexicons like HurtLex \cite{galery2018aggression}, demonstrating the potential of hybrid approaches in improving detection accuracy.
Document-level embeddings, such as Doc2Vec \cite{le2014distributed}, have offered a broader perspective by encapsulating the semantic content of entire text segments, providing insights into the overall tone and sentiment crucial for aggression detection \citep{ramiandrisoa2018irit}. This approach addresses some limitations of word-level embeddings by considering the broader context of communication. 
Recent advancements in natural language processing have ushered in a new era of contextual embeddings, primarily through transformer-based models like BERT, RoBERTa, xlm-RoBERTa, etc. These models provide dynamic, context-sensitive token representations that capture the nuanced meanings of words across different contexts. While primarily used for fine-tuning in aggression detection tasks, their potential to provide rich, adaptable text representations marks a significant leap forward in the field.

\begin{figure}[h!]
    \centering
    \subfloat[\centering Utilization of traditional text embeddings and representations.]{{\includegraphics[scale=0.30]{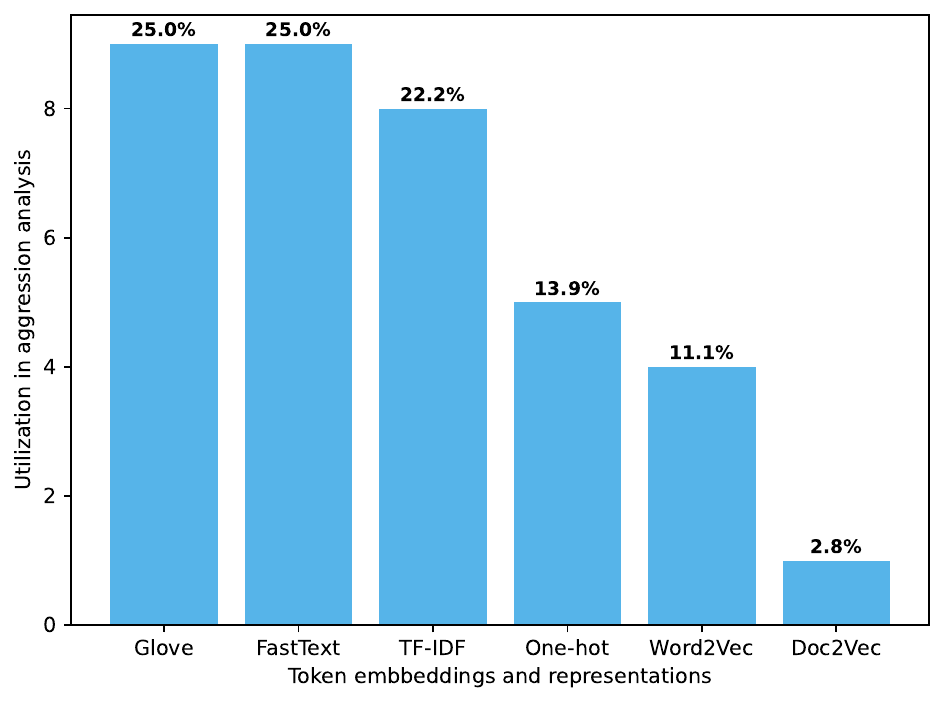} }}%
    \qquad
    \subfloat[\centering Utilization of types of features.]{{\includegraphics[scale=0.40]{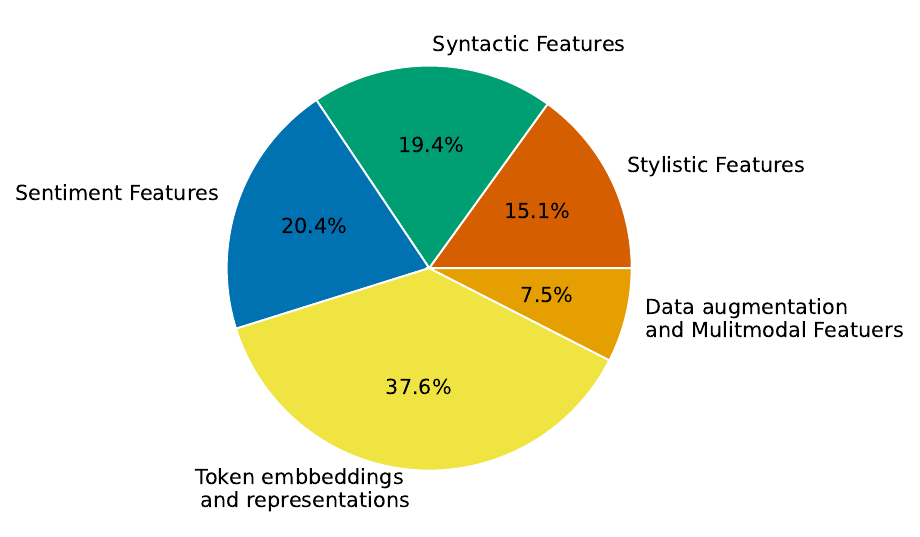} }}%
    \caption{Distribution analysis of features utilized in previous studies for aggression analysis.}%
    \label{fig:featuers_data_proportion}%
\end{figure}

\subsubsection{Data augmentation and Multimodal Features}
To overcome the issues of data unavailability and class imbalance, various data augmentation techniques have been employed. These include machine translation-based data augmentation \citep{akter2022deep, risch2018aggression, ta2022multi}, the use of language models like GPT-2 for synthetic data generation \citep{Shrivastava2021EnhancingAD}, and pseudo-labeling strategies \citep{aroyehun2018aggression}. These methods aim to expand the diversity and representation of the training data, enabling models to better generalize and handle the nuances of aggressive content.
Beyond textual features, researchers have also investigated the integration of multimodal information for aggression detection. These approaches combine linguistic cues with visual, network, and social media user profile features \citep{herodotou2021catching, kumari2022multi}. For instance, user-level features such as in-degree and out-degree ego networks, account age, number of posts, and subscriptions have been incorporated to better understand user behavior and its association with aggressive tendencies. Hybrid feature sets, leveraging both image and text data, have been optimized using techniques like binary firefly optimization (BFFO), with VGG16 for image features and GloVe for text features \citep{kumari2022multi}. \textit{The integration of non-textual data sources, such as images and user profiles, requires specialized preprocessing and feature engineering, adding complexity to the overall detection pipeline. Furthermore, the interpretability and explainability of multimodal models can be more challenging, as the interplay between different modalities may not always be straightforward.}

\begin{table}[]
\centering
\caption{Comparison of feature types for aggression detection.}
\label{tab:fet-comp}
\resizebox{\textwidth}{!}{%
\begin{tabular}{lllll}
\hline
\multicolumn{1}{c}{\textbf{Feature Type}} & \multicolumn{1}{c}{\textbf{Performance}} & \multicolumn{1}{c}{\textbf{Advantages}} & \multicolumn{1}{c}{\textbf{Limitations}} & \multicolumn{1}{c}{\textbf{Best Use Cases}} \\ \hline
Stylistic & Moderate (F1: 65-80\%) & \begin{tabular}[c]{@{}l@{}}- Captures writing nuances\\ - Effective for explicit aggression\\ - Computationally efficient\end{tabular} & \begin{tabular}[c]{@{}l@{}}- May miss context-dependent aggression\\ - Can be platform-specific\\ - Vulnerable to evasion tactics\end{tabular} & \begin{tabular}[c]{@{}l@{}}- Social media posts\\ - Short-form content\end{tabular} \\ \hline
Syntactic & Good (F1: 70-85\%) & \begin{tabular}[c]{@{}l@{}}- Reveals sentence structure patterns\\ - Effective for implicit aggression\\ - Language-agnostic potential\end{tabular} & \begin{tabular}[c]{@{}l@{}}- Computationally intensive\\ - May struggle with informal language\\ - Requires sophisticated NLP tools\end{tabular} & \begin{tabular}[c]{@{}l@{}}- Formal texts\\ - Cross-lingual detection\end{tabular} \\ \hline
Sentiment & Moderate (F1: 60-75\%) & \begin{tabular}[c]{@{}l@{}}- Captures emotional tone\\ - Useful for detecting hostile intent\end{tabular} & \begin{tabular}[c]{@{}l@{}}- Often language-dependent\\ - May oversimplify complex emotions\\ - Can be misled by sarcasm\end{tabular} & \begin{tabular}[c]{@{}l@{}}- Review platforms\\ - Opinion pieces\end{tabular} \\ \hline
\begin{tabular}[c]{@{}l@{}}Text \\ Representation\end{tabular} & Very Good (F1: 75-90\%) & \begin{tabular}[c]{@{}l@{}}- Captures semantic relationships\\ - Adaptable to various contexts\\ - Enables deep learning approaches\end{tabular} & \begin{tabular}[c]{@{}l@{}}- Can be computationally expensive\\ - May require large training datasets\\ - Interpretability challenges\end{tabular} & \begin{tabular}[c]{@{}l@{}}- Large-scale text analysis\\ - Multilingual contexts\end{tabular} \\ \hline
\end{tabular}%
}
\end{table}

Our analysis of previous research on features used in the field of aggression detection highlights the prevalent use of traditional text representation methods. It clearly emphasizes FastText and GloVe, which account for approximately 50\% of the selected features. TF-IDF, although less dominant, still has a significant percentage at 22\%, as shown in Figure \ref{fig:featuers_data_proportion}a. Moreover, we observed a clear hierarchy after analyzing the popularity of different features in the reviewed studies. Token embeddings and representations claim the foremost position, accounting for about 37\% of feature selections, followed by emotion features, which contribute about 20\% of the feature landscape. In third place, syntactic features account for approximately 19\% of the total feature usage, as shown in Figure \ref{fig:featuers_data_proportion}b. These observations highlight the importance of token embedding, emotion features, and syntactic properties in detecting affective aggression and provide valuable insights for future research in this domain. Additionally, Table \ref{tab:fet-comp} provides a comparative overview of the major feature types used in aggression detection, highlighting their advantages, limitations, and best use cases.

\begin{figure}[h!]
    \centering
    \subfloat[\centering Distribution of aggression detection studies across languages.]{{\includegraphics[scale=0.35]{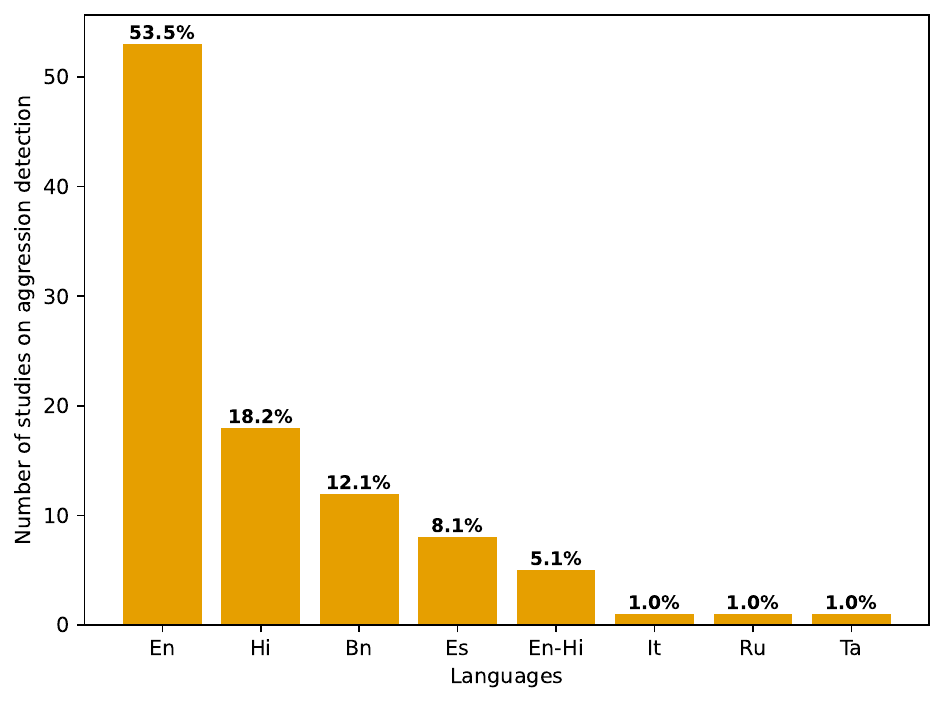} }}%
    \qquad
    \subfloat[\centering Distribution of machine learning algorithms employed in aggression detection studies.]{{\includegraphics[scale=0.35] {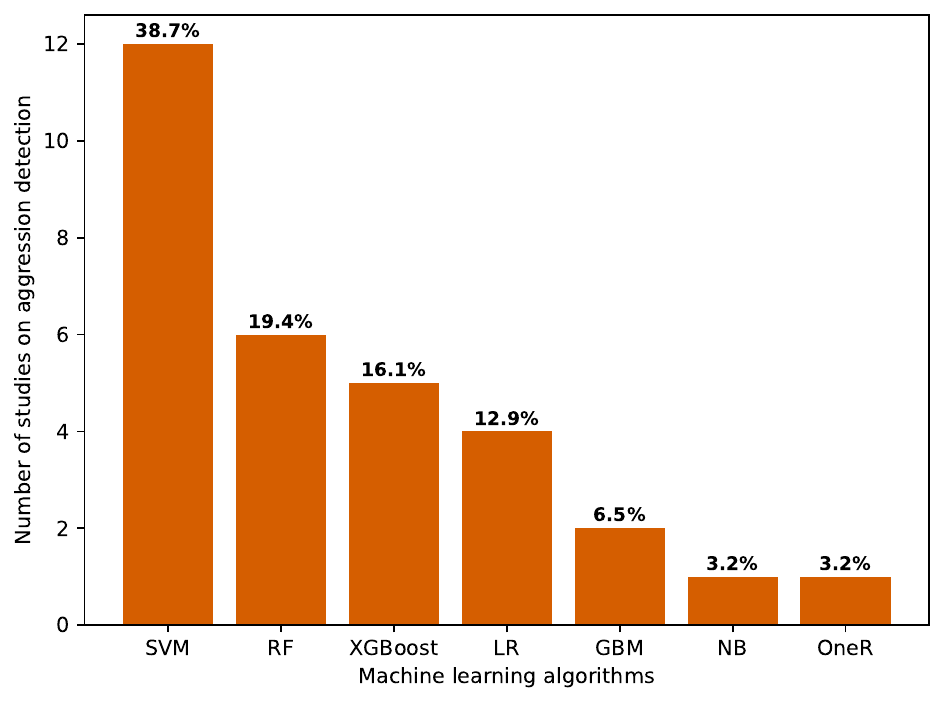} }}%
    \caption{Distribution analysis of detection studies across various languages and ML algorithms.}%
    \label{fig:det_lang_ml_proportion}%
    \vspace*{-1.3\baselineskip}
\end{figure}

\subsection{Detection Algorithms}

In this section, we explore algorithms employed in detecting aggression, ranging from traditional machine learning methods to state-of-the-art deep learning and transformer-based models. Algorithms have been developed to address aggressiveness in multiple languages, such as English (En) \citep{samghabadi2018ritual, ramiandrisoa2018irit, Raiyani2018FullyCN, Nikhil2018LSTMsWA, kumar2018trac, roy2018ensemble, galery2018aggression, maitra2018k, aroyehun2018aggression, srivastava2018identifying, oravsan2018aggressive, madisettyaggression, fortuna2018merging, arroyo2018cyberbullying, risch2018aggression, orabi2018cyber, tommasel2018textual, golem2018combining, modha2018filtering, singh2018aggression, patwa2021hater, modha2022empirical, Tommasel2019AnES, Srivastava2019DetectingAA, ramiandrisoa2020aggression, ramiandrisoa2020irit, liu2020scmhl5, pascucci2020role, altin2020lastus, tawalbeh2020saja, rischbaggingBM, Baruah2020AggressionII, Datta2020SpyderAD, Gordeev2020BERTOA, koufakou2020florunito, kumari2020ai, Mishra2020MultilingualJF, Samghabadi2020AggressionAM, Khandelwal2020AUS, Chen2018VerbalAD, zhao2021comparative, Kumari2021BilingualCD, dutta2021efficient, dutta2021efficient, herodotou2021catching, akter2022deep, Ramiandrisoa2022MultitaskLF, khan2022aggression, kumari2022multi, agbaje2022neural, Iqbal2019UsingCL, ghosh2023transformer, rawat2023modelling}, Hindi (Hi) \citep{ramiandrisoa2018irit, Raiyani2018FullyCN, Nikhil2018LSTMsWA, roy2018ensemble, maitra2018k, risch2018aggression, modha2018filtering, rischbaggingBM, Baruah2020AggressionII, Datta2020SpyderAD, Gordeev2020BERTOA, koufakou2020florunito, kumari2020ai, Mishra2020MultilingualJF, Samghabadi2020AggressionAM, Khandelwal2020AUS, akter2022deep, ghosh2023transformer}, Bangla (Bn) \citep{rischbaggingBM, Baruah2020AggressionII, Datta2020SpyderAD, Gordeev2020BERTOA, koufakou2020florunito, kumari2020ai, Mishra2020MultilingualJF, Samghabadi2020AggressionAM, Khandelwal2020AUS, akter2022deep, sharif2022tackling, ghosh2023transformer}, Spanish (Es) \citep{gomez2018machine, GutirrezEsparza2019ClassificationOC, Tanase2020DetectingAI, GuzmanSilverio2020TransformersAD, tonja2022detection, ta2022multi, ta2022gan, huerta2022verbal}, Italian (It) \citep{gattulli2022cyber}, Russian (Ru) \citep{shulginov2021automatic}, and even in code-mixed versions like Hindi-English (Hi-En) \citep{aroyehun2018aggression, si2019aggression, Shrivastava2021EnhancingAD, Mundra2021EvaluationOT, sengupta2022does} and Tamil-English (Ta-En) \citep{SaneCorpusCA}. 
In the recent past, some events have attracted the attention of the research community towards the detection of aggression. These include the First and Second Workshop on Trolling, Aggression, and Cyberbullying (TRAC-2018\cite{kumar2018aggression} and TRAC-2020 \cite{bhattacharya2020developing}) and the DA-VINCIS competition (Detection of Aggressive and Violent INCIdents from Social Media in Spanish) at IberLEF 2022 \cite{arellano2022overview}.
Further, Table \ref{tab:detection_table} provides a detailed year-wise analysis of aggression detection algorithms, describing their respective datasets, performance metrics, and their status in the research community as estimated by citation counts.
A distribution analysis of previous studies on aggression detection in different languages is presented in Figure \ref{fig:det_lang_ml_proportion}a. Our analysis reveals that the majority of aggression detection studies are in English, representing 53\%. Hindi follows at 18\%, Bangla at 12\%, Spanish at 8\%, and Hindi-English code-mixed at 5\%. This data highlights a clear preference for English in the research on aggression detection.

\subsubsection{Machine learning}
Early research in aggression detection relied on traditional machine learning algorithms. These methods include Logistic Regression (LR) \citep{samghabadi2018ritual, ramiandrisoa2018irit, golem2018combining, Shrivastava2021EnhancingAD}, Gradient Boosting Machine (GBM) \citep{Datta2020SpyderAD, si2019aggression}, Random Forest (RF) \citep{ramiandrisoa2018irit, oravsan2018aggressive, fortuna2018merging, gattulli2022cyber, Iqbal2019UsingCL, kumari2022multi}, eXtreme Gradient Boosting (XGBoost) \citep{Datta2020SpyderAD, si2019aggression, tawalbeh2020saja, herodotou2021catching, dutta2021efficient}, Support Vector Machine (SVM) \citep{oravsan2018aggressive, tommasel2018textual, golem2018combining, arroyo2018cyberbullying, gomez2018machine, si2019aggression, Tommasel2019AnES, SaneCorpusCA, pascucci2020role, Baruah2020AggressionII, Shrivastava2021EnhancingAD, huerta2022verbal}, Naive Bayes (NB) \citep{Shrivastava2021EnhancingAD}, and One Rule (OneR) \citep{GutirrezEsparza2019ClassificationOC}. While these approaches may seem less sophisticated compared to more recent neural network-based methods, they continue to play a crucial role in aggression detection research due to their interpretability and effectiveness in scenarios with limited training data.
The \textit{passive-aggressive learning} algorithm has also been applied with SVM and SGD, adjusting weight vectors that parameterize a linear decision boundary \citep{arroyo2018cyberbullying}.
Real-time Twitter data streams have also been analyzed using streaming ML algorithms, including Hoeffding Tree (HT), Adaptive Random Forest (ARF), and Streaming Logistic Regression (SLR) \citep{herodotou2021catching}.
Further, the distribution analysis of machine learning algorithms employed in previous aggression detection studies is presented in 
Figure \ref{fig:det_lang_ml_proportion}b. Our analysis reveals that SVMs are the most commonly employed machine learning algorithm at 38\%, underscoring their effectiveness in aggression detection. RF is the second choice at 19\%, indicating its popularity for this task. XGBoost and LR follow with 16\% and 12\% usage, respectively, while GBM is used in 6\% of studies. NB and OneR are less frequent choices, both at 3\%.

\subsubsection{Advanced Machine Learning Approaches}
In the evolution of aggression detection, researchers have increasingly adopted advanced machine learning techniques, such as deep learning models. This section is divided into deep learning architectures and transformer-based models, reflecting their historical development and the significant leap in performance that transformers have brought to the field.
Importantly, these approaches are not mutually exclusive.

\begin{figure}[]
    \centering
    \subfloat[\centering Utilization of Deep learning algorithms.]{{\includegraphics[scale=0.30]{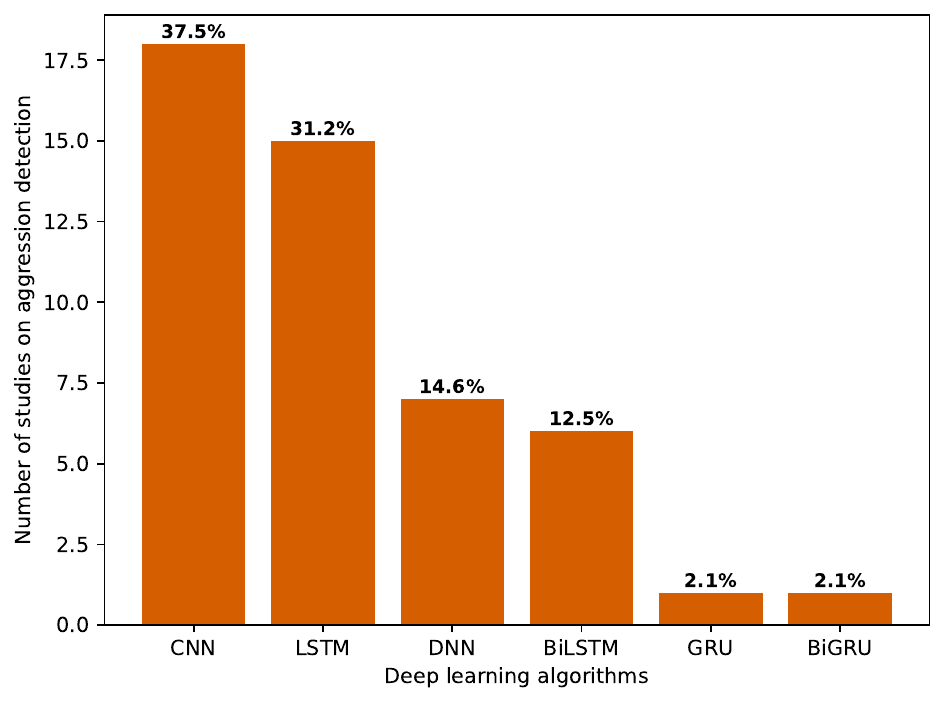} }}%
    \qquad
    \subfloat[\centering Utilization of Transformer-based algorithms.]{{\includegraphics[scale=0.30]{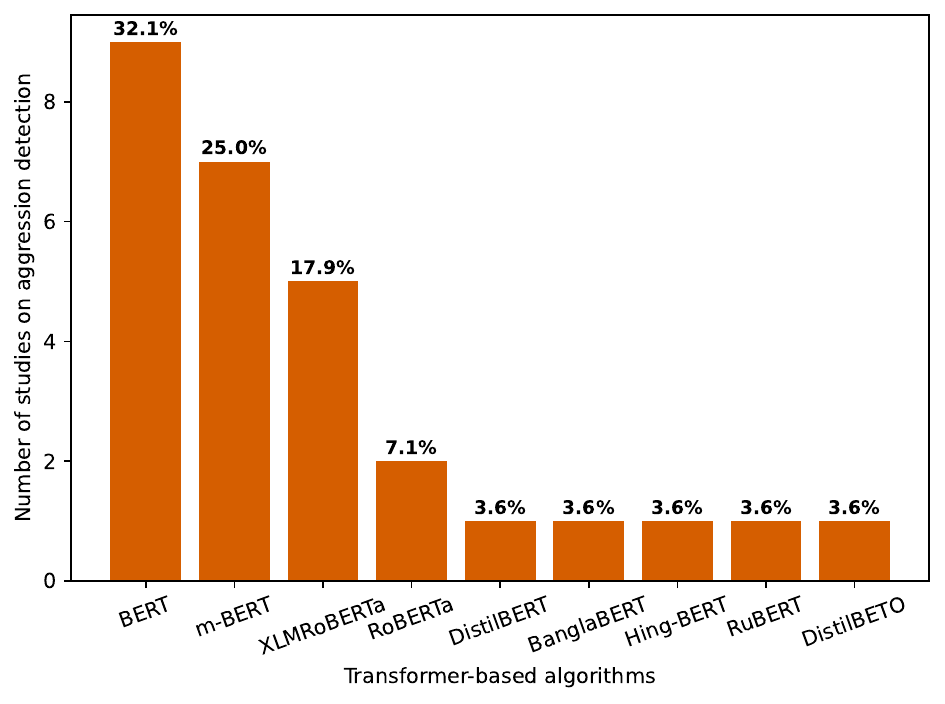} }}%
    \caption{Comparative percentage distribution of DL and transformer-based algorithms used in studies focused on aggression detection.}%
    \label{fig:det_dl_tf_proportion}%
    \vspace*{-1.3\baselineskip}
\end{figure}

\paragraph{Deep Learning Architectures}
Deep learning models are capable of capturing complex patterns in textual data \citep{khan2022aggression, Raiyani2018FullyCN, maitra2018k, tommasel2018textual, sadiq2021aggression}. Convolutional neural network (CNN) have been used in extracting features from textual data for aggression detection \citep{roy2018ensemble, aroyehun2018aggression, madisettyaggression, orabi2018cyber, modha2018filtering, singh2018aggression, patwa2021hater, modha2022empirical, kumari2019aggressive, kumari2020ai, Mundra2021EvaluationOT, sharif2022tackling, sengupta2022does, kumari2022multi, Chen2018VerbalAD, Shrivastava2021EnhancingAD, Khandelwal2020AUS, raman2022hate}. 
Long Short-Term Memory (LSTM) models are used for sequence modeling tasks in aggression detection \citep{Nikhil2018LSTMsWA, aroyehun2018aggression, madisettyaggression, tommasel2018textual, golem2018combining, modha2018filtering, aroyehun2018aggression, koufakou2020florunito, kumari2020ai, Kumari2021BilingualCD, Mundra2021EvaluationOT, sengupta2022does, Shrivastava2021EnhancingAD, Khandelwal2020AUS, Karan2023}. Bidirectional LSTMs (BiLSTMs) further enhance performance by considering both past and future context \citep{kumar2018trac, madisettyaggression, golem2018combining, Srivastava2019DetectingAA, altin2020lastus, sharif2022tackling, mane2023you}. Other variants of RNN (Recurrent Neural Network) have also been utilized, such as Gated Recurrent Unit (GRU) \citep{risch2018aggression, raman2022hate} and Bidirectional GRU (BiGRU)\citep{galery2018aggression}.
Capsule networks have also been explored for aggression detection. In \cite{srivastava2018identifying}, a focal loss-based capsule network is used, while in \cite{patwa2021hater} a capsule network followed by a CNN is utilized to enhance the aggression detection. Researchers have also explored multimodal approaches, combining text features with LSTMs and image features with CNNs to improve detection accuracy \cite{sengupta2022does, kumari2022multi}.
RNNs and their variants, like LSTMs, have issues like the vanishing gradient problem. This limitation is addressed by attention mechanisms and transformer-based models. These are effective for processing longer text sequences and capturing complex, context-dependent patterns in aggressive language.
Figure \ref{fig:det_dl_tf_proportion}a reveals the distribution of deep learning algorithms in aggression detection studies. CNNs are the most favored at 37\%, emphasizing their proficiency in processing textual cues. LSTMs are the second choice at 31\%, indicating the importance of temporal dynamics. DNNs, used in 14\% of the studies which offer a more generalized approach. BiLSTMs at 12\% capture contextual nuances effectively. GRUs and BiGRUs, together at 2\%, are less common, possibly due to their reduced complexity. These insights aid researchers and practitioners in the selection of algorithms for aggression detection.

\paragraph{Transformer-based Models}
The introduction of transformer-based models \citep{vaswani2017}, particularly BERT (Bidirectional Encoder Representations from Transformers) and its variants, has marked a significant advancement in natural language processing tasks, including aggression detection. These models utilize self-attention mechanisms and pre-training on large datasets to capture contextual information more effectively than previous approaches. As a result, BERT-based models have detected various forms of aggressive language effectively \citep{ramiandrisoa2020aggression, liu2020scmhl5, risch2018aggression, Baruah2020AggressionII, Mishra2020MultilingualJF, Samghabadi2020AggressionAM, GuzmanSilverio2020TransformersAD, dutta2021efficient, akter2022deep}. The multilingual BERT (m-BERT) extends the utility of these models across linguistic boundaries \citep{risch2018aggression, Gordeev2020BERTOA, Mishra2020MultilingualJF, Samghabadi2020AggressionAM, akter2022deep, sharif2022tackling, ta2022multi}. Distilled versions of BERT (DistilBERT) offer a more lightweight alternative \citep{sharif2022tackling}. Robustly Optimized BERT Approach (RoBERTa) \citep{Ramiandrisoa2022MultitaskLF, rawat2023modelling}, Cross-lingual Language Model RoBERTa (XLMRoBERTa) \citep{Baruah2020AggressionII, Tanase2020DetectingAI, zhao2021comparative, sharif2022tackling, ghosh2023transformer}, and language-specific variants of BERT, such as BERT pre-trained on Bangla (BanglaBERT) \citep{sharif2022tackling}, BERT pre-trained on code-mixed Hindi-English (Hing-BERT) \citep{rawat2023modelling}, BERT pre-trained on Russian (RuBERT) \citep{shulginov2021automatic}, and DistilBERT pre-trained on Spanish (DistilBETO) \citep{tonja2022detection}. These variants extend the applicability of transformer-based models for catering to languages and code-mixed text with unique linguistic traits. 
The landscape of transformer-based models also includes Generative Pretrained Transformer 2 (GPT-2), which explores text generation and provides valuable insights into understanding and detecting aggressive language \citep{akter2022deep}.  Incorporating attention mechanisms into transformer models enhances the critical aspects of aggression detection \citep{Samghabadi2020AggressionAM, ghosh2023transformer}. Additionally, researchers find that maintaining class balance through the inclusion of instances from external datasets consistently improves transformer-based model performance \cite{ramiandrisoa2020irit}. In multi-task learning, \cite{Ramiandrisoa2022MultitaskLF} designed a model to identify English aggression and hate speech using RoBERTa model. Similarly, the study by \cite{ghosh2023transformer} tackles multi-task learning challenges associated with the identification of aggression and offensive content. The shared XLMRoBERTa model captures common task features, while individual multi-head self-attention networks define task-specific characteristics. This showcases the versatility in handling diverse dimensions of aggression and offensive language.
\cite{bansal2022} introduced XLM-RoBERTa with BiGRU, incorporating emoji embedding, for aggressive (abusive) language detection across 13 Indic code-mixed languages. \citet{ta2022gan} used GAN-BERT for Spanish aggression detection. 
In Figure \ref{fig:det_dl_tf_proportion}b, BERT stands out as the most widely used transformer-based model in aggression detection studies, representing 26\% of the total. M-BERT follows closely at 20\% and XLMRoBERTa at 18\%. RoBERTa is employed in 7\% of the studies. These findings highlight the prevalence of transformer-based models, especially BERT for aggression detection.

\subsubsection{Ensemble-based}
The ensemble approach by \citet{tommasel2018textual} utilized three layers of LSTM, DNN, and SVM, effective for aggression identification. Ensemble methods expanded further with \citet{golem2018combining}, proposing a model combining BiLSTM, CNN, LR, and SVM for aggression detection. Similarly, \cite{si2019aggression} introduced a majority voting ensemble model, employing XGBoost, GBM, and SVM for code-mixed Hindi-English aggression. \citet{Shrivastava2021EnhancingAD} adopted a similar approach, incorporating SVM, NB, LR, CNN, and LSTM into their ensemble model for code-mixed Hindi-English aggression detection.
A combination of a three-capsule network with CNN was proposed for aggression detection \cite{patwa2021hater}. 
In the context of code-mixed languages, \citet{Khandelwal2020AUS} addressed the challenge with an average ensemble model that leveraged the Deep Pyramid CNN, Pooled LSTM, and Disconnected RNN, offering a holistic solution for code-mixed aggression detection.
The multifaceted nature of aggression detection extended to the fusion of image and text features in the work of \cite{kumari2022multi}. Their model incorporated VGG-16 for image features and CNN for text features, with Random Forest (RF) used for aggression identification in English text. \citet{aroyehun2018aggression} trained a CNN-LSTM model with the combination of augmented training data and pseudo-labeled Twitter data. 
Further, \citet{ramiandrisoa2020irit} integrated the three uncased-BERT-base variants, each along with a low-dimensional multi-head attention layer known as Projected Attention Layers (PALs) \cite{stickland2019bert}. They fine-tuned PALs using majority voting strategies for aggression detection. To enhance comprehensiveness, \citet{liu2020scmhl5} proposed an ensemble model featuring three uncased-BERT-base variants, each with different feature sets. \citet{risch2018aggression} achieved notable results by combining 20 uncased-BERT models tailored for specific tasks and languages. \citet{sharif2022tackling} proposed a weighted-ensemble model that combines m-BERT, distill-BERT, Bangla-BERT, and XLM-R for Bangla aggression detection. For Spanish aggression detection, \citet{GuzmanSilverio2020TransformersAD} employed a weighted ensemble of 20 pre-trained BERT models, each fine-tuned to the specific. 

\subsection{State-of-the-art Approaches and Emerging Trends}
The state-of-the-art (SOTA) approaches are the ones that have demonstrated outstanding performance on relevant aggression detection datasets. We have examined the state-of-the-art approaches for the three most widely utilized datasets \citep{kumar2018aggression, bhattacharya2020developing, sharif2022tackling}.
\citet{Kumari2021BilingualCD} achieved notable results on the \cite{kumar2018aggression} dataset using an LSTM autoencoder. Their model, comprising a three-layered encoder and decoder LSTM structure to detect aggression for English from Facebook and Twitter, achieved 81\% weighted F1 scores. \citet{rischbaggingBM} established SOTA performance on the \citep{bhattacharya2020developing} dataset through an ensemble of BERT models. Their approach, tailored for English, Hindi, and Bangla, utilized language-specific BERT variants and achieved weighted F1 scores of 80.29-93.85\% across languages. For Bangla-specific detection, \citep{sharif2022tackling} introduced a weighted-ensemble model combining m-BERT, DistilBERT, Bangla-BERT, and XLM-R. This method yielded better results on Facebook and YouTube data, with weighted F1 scores of 93\% for both aggression and misogyny detection. 
While the prior work addressed the multilingual data

\begin{table}[h!]
\centering
\caption{Comparison of algorithm types for aggression detection.}
\label{tab:type-comp}
\resizebox{\textwidth}{!}{%
\begin{tabular}{llll}
\hline
\multicolumn{1}{c}{\textbf{Algorithm Type}} & \multicolumn{1}{c}{\textbf{Performance}} & \multicolumn{1}{c}{\textbf{Advantages}} & \multicolumn{1}{c}{\textbf{Limitations}} \\ \hline
Traditional ML (e.g., SVM, RF) & Moderate (F1: 60-75\%) & \begin{tabular}[c]{@{}l@{}}- Interpretable results\\ - Effective with limited data\end{tabular} & \begin{tabular}[c]{@{}l@{}}- Limited ability to capture context\\ - Requires feature engineering\end{tabular} \\ \hline
Deep Learning (e.g., CNN, LSTM) & Good (F1: 70-85\%) & \begin{tabular}[c]{@{}l@{}}- Captures complex patterns\\ - Effective for long sequences\\ - Automatic feature extraction\end{tabular} & \begin{tabular}[c]{@{}l@{}}- Requires large datasets\\ - Computationally intensive\\ - Less interpretable\end{tabular} \\ \hline
Transformers (e.g., BERT) & Excellent (F1: 80-93\%) & \begin{tabular}[c]{@{}l@{}}- State-of-the-art performance\\ - Captures contextual information\\ - Effective for multiple languages\end{tabular} & \begin{tabular}[c]{@{}l@{}}- Very computationally intensive\\ - Requires extensive fine-tuning\\ - May overfit on small datasets\end{tabular} \\ \hline
Ensemble Methods & Very Good (F1: 75-90\%) & \begin{tabular}[c]{@{}l@{}}- Combines strengths of multiple models\\ - Robust to individual model weaknesses\end{tabular} & \begin{tabular}[c]{@{}l@{}}- Increased complexity\\ - Potentially slower inference\end{tabular} \\ \hline
\end{tabular}%
}
\end{table}

\begin{figure}[h!]
\centering
\includegraphics[scale=0.35]{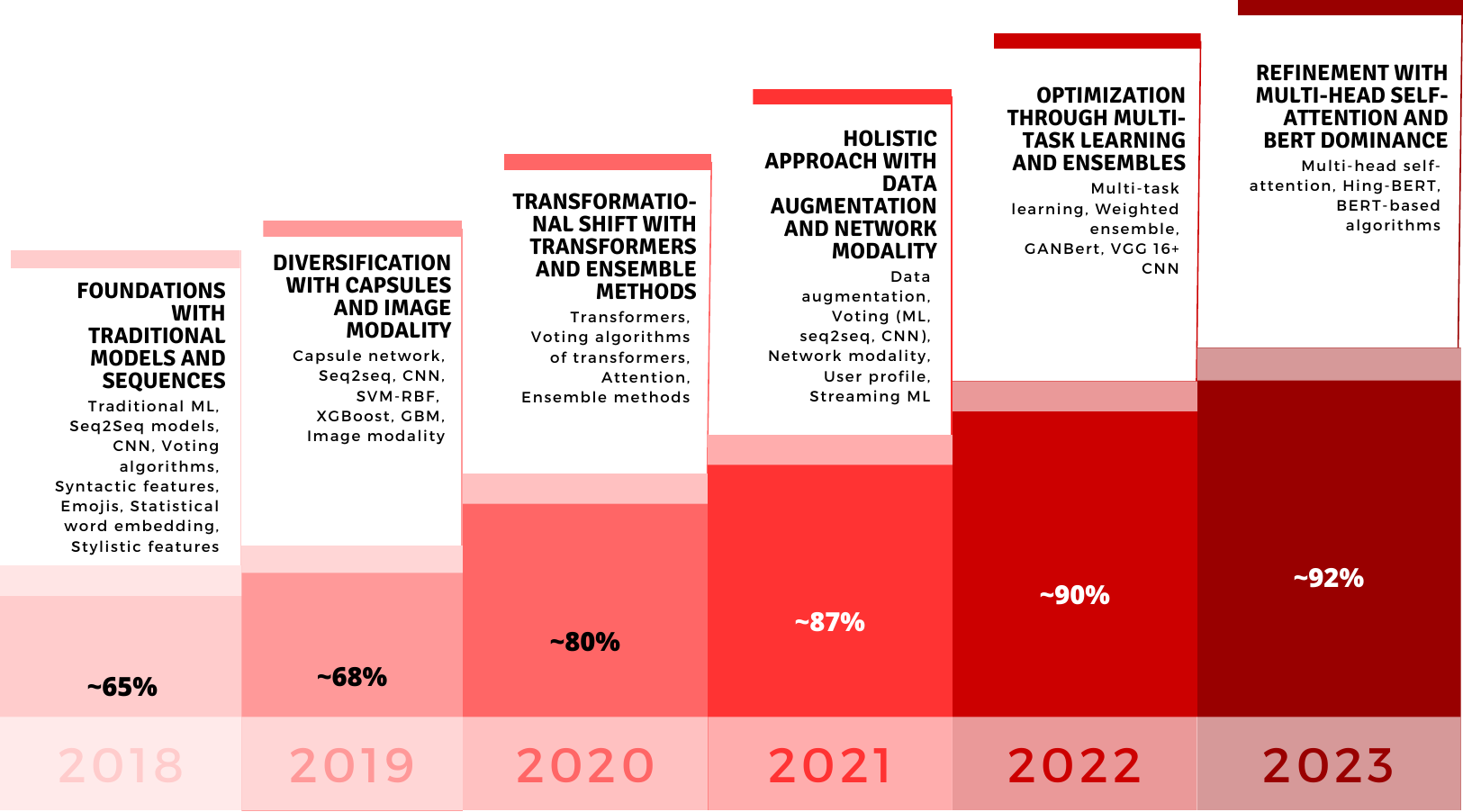}
\caption{Evolution of aggressive content detection over the last six years: Analyzing progress and trends in computational approaches. The approximate ($\sim$) performance is shown in percentage (\%).} 
\label{timeline_progress}
\vspace*{-1\baselineskip}
\end{figure}

Furthermore, Table \ref{tab:type-comp} provides a comparative overview of the major algorithm types used in aggression detection, highlighting their performance, advantages, and limitations.
Additionally, Figure \ref{timeline_progress} presents the evolution of aggression detection from 2018 to 2023 and reveals a progression from traditional models to transformers. The refinement in 2023 with multi-head self-attention and the dominance of BERT-based algorithms marks an evolved state in aggression detection.
Researchers are exploring cross-lingual transfer learning techniques for generalizing aggression detection models across languages with limited language-specific datasets. For instance, \citet{wu-dredze-2020-languages} demonstrated the effectiveness of multilingual BERT in zero-shot cross-lingual transfer for various NLP tasks. Similarly, \citet{pires-etal-2019-multilingual} showed that m-BERT performs well on cross-lingual transfer tasks even for languages unseen during pre-training.
These methods leverage the shared structural similarities between languages captured by multilingual models.

\subsection{Ethical Considerations and Bias Mitigation}
The development and deployment of these algorithms raise important questions about fairness, transparency, and unintended consequences.
One primary concern is algorithmic bias, where models may disproportionately flag content from certain demographic groups or languages as aggressive. This bias can stem from imbalanced training data, cultural misunderstandings, or the encoding of societal biases into the models themselves. For instance, \citet{Mishra2020MultilingualJF} highlighted the challenges of creating fair multilingual models for aggression detection, emphasizing the need for diverse and representative datasets.
Another critical issue is the potential for false positives, where non-aggressive content is incorrectly classified as aggressive. This can lead to unwarranted content removal or account suspensions, potentially infringing on users' freedom of expression. Conversely, false negatives (failing to detect genuinely aggressive content) can allow harmful content to proliferate, undermining the very purpose of these systems.
To address these concerns, researchers are exploring various approaches:
Interpretable AI: Creating models that can provide explanations for their decisions, allowing for greater transparency and easier identification of biases \citep{Karan2023}.
Diverse and representative datasets: Ensuring that training data covers a wide range of languages, cultures, and demographic groups to minimize bias \citep{bhattacharya2020developing}.
Human-in-the-loop systems: Incorporating human oversight and appeals processes to mitigate the impact of algorithmic errors \citep{kumar2018aggression}.
Contextual analysis: Developing more nuanced models that consider broader context and intent, rather than relying solely on keyword-based approaches \citep{Ramiandrisoa2022MultitaskLF}.

{
\tiny
\setlength\LTleft{0pt}
\setlength\LTright{0pt}
\begin{longtable}[c]{lllrlllr}
\caption{Provides a detailed overview of research efforts to detect aggressive content for specific datasets. The table includes information on the targeted languages, the effective features employed, the proposed algorithm and its performance, and the number of citations.}
\label{tab:detection_table}\\
\hline
\multicolumn{1}{c}{\textbf{Ref.}} & \multicolumn{1}{c}{\textbf{Dataset}} & \textbf{Lang.} & \multicolumn{1}{c}{\textbf{Year}} & \multicolumn{1}{c}{\textbf{Features}} & \multicolumn{1}{c}{\textbf{Algorithms}} & \multicolumn{1}{c}{\textbf{Results}} & \multicolumn{1}{l}{\textbf{Cites}} \\ \hline
\endfirsthead
\multicolumn{8}{c}%
{{\bfseries Table \thetable\ continued from previous page}} \\
\hline
\multicolumn{1}{c}{\textbf{Ref.}} & \multicolumn{1}{c}{\textbf{Dataset}} & \textbf{Lang.} & \multicolumn{1}{c}{\textbf{Year}} & \multicolumn{1}{c}{\textbf{Features}} & \multicolumn{1}{c}{\textbf{Algorithms}} & \multicolumn{1}{c}{\textbf{Results}} & \multicolumn{1}{l}{\textbf{Cites}} \\ \hline
\endhead

\cite{samghabadi2018ritual} & \cite{kumar2018aggression} & \begin{tabular}[c]{@{}l@{}}En, \\ Hi\end{tabular} & 2018 & \begin{tabular}[c]{@{}l@{}}En: Lexical, Semantic\\ Hi: Lexical\end{tabular} & LR & \begin{tabular}[c]{@{}l@{}}W-F1: \{Fb: 0.5921, \\ Tw: 0.6451\}\end{tabular} & $\geq41 $\\ \hline
\cite{ramiandrisoa2018irit} & \cite{kumar2018aggression} & En & 2018 & \begin{tabular}[c]{@{}l@{}}Lexical, POS, Negation, \\ Capitalized, Sentiment, \\ Emotions, Swear words, \\ Doc2Vec, Others\end{tabular} & \begin{tabular}[c]{@{}l@{}}RF (features) + \\ LR (Doc2vec)\end{tabular} & \begin{tabular}[c]{@{}l@{}}W-F1: \{Fb: 0.576,\\ Tw: 0.511\}\end{tabular} & $\geq11$ \\ \hline
\cite{Raiyani2018FullyCN} & \cite{kumar2018aggression} & \begin{tabular}[c]{@{}l@{}}En, \\ Hi\end{tabular} & 2018 & \begin{tabular}[c]{@{}l@{}}One-hot \\ encoding\end{tabular} & DNN - 3 layers & \begin{tabular}[c]{@{}l@{}}W-F1: \{\\Fb: \{En: 0.5813, Hi: 0.5951\},\\ Tw: \{En: 0.6009,Hi: 0.4830\}\}\end{tabular} & $\geq38$ \\ \hline
\cite{Nikhil2018LSTMsWA} & \cite{kumar2018aggression} & \begin{tabular}[c]{@{}l@{}}En, \\ Hi\end{tabular} & 2018 & - & LSTM + attention & \begin{tabular}[c]{@{}l@{}}W-F1: \{\\ Fb: \{En: 0.5746, Hi: 0.6032\},\\ Tw: \{En: 0.5548,Hi: 0.4703\}\}\end{tabular} & $\geq29$ \\ \hline
\cite{kumar2018trac} & \cite{kumar2018aggression} & En & 2018 & Deepmoji & LSTM & \begin{tabular}[c]{@{}l@{}}W-F1: \{Fb: 0.3572, \\Tw: 0.1960\}\end{tabular} & $\geq29$ \\ \hline
\cite{roy2018ensemble} & \cite{kumar2018aggression} & \begin{tabular}[c]{@{}l@{}}En, \\ Hi\end{tabular} & 2018 & \begin{tabular}[c]{@{}l@{}}CNN: \{En: Glove, \\ Hi: Pretrained on Wikipedia\}\\ SVM: Unigram and TF-IDF\end{tabular} & CNN + SVM & \begin{tabular}[c]{@{}l@{}}W-F1: \{\\ Fb: \{En: 0.5151, Hi: 0.5599\},\\ Tw: \{En: 0.5099,Hi: 0.3790\}\}\end{tabular} & $\geq17$ \\ \hline
\cite{galery2018aggression} & \cite{kumar2018aggression} & En & 2018 & FastText & Bi-GRU & \begin{tabular}[c]{@{}l@{}}W-F1: \{ Fb: 0.5315, \\ Tw: 0.4389\}\end{tabular} & $\geq17$ \\ \hline
\cite{maitra2018k} & \cite{kumar2018aggression} & \begin{tabular}[c]{@{}l@{}}En, \\ Hi\end{tabular} & 2018 & \begin{tabular}[c]{@{}l@{}}Wordcount, Sentiment, \\ Capitalization, Hashtags\end{tabular} & \begin{tabular}[c]{@{}l@{}}DNN - 3 layers \\ + autoencoder\end{tabular} & \begin{tabular}[c]{@{}l@{}}W-F1: \{\\ Fb: \{En: 0.5694, Hi: 0.4189\},\\ Tw: \{En: 0.3379,Hi: 0.3132\}\}\end{tabular} & $\geq6$ \\ \hline
\cite{aroyehun2018aggression} & \cite{kumar2018aggression} & En & 2018 & \begin{tabular}[c]{@{}l@{}}DA: Translation\\ Pseudo labeling\end{tabular} & LSTM, CNN-LSTM & \begin{tabular}[c]{@{}l@{}}W-F1: \{\\ Fb: 0.6425 (LSTM),\\ Tw: 0.5920 (LSTM-CNN)\}\end{tabular} & $\geq155$ \\ \hline
\cite{srivastava2018identifying} & \cite{kumar2018aggression} & En & 2018 & One-hot encoding & \begin{tabular}[c]{@{}l@{}}CapsuleNet with \\Focal Loss\end{tabular} & \begin{tabular}[c]{@{}l@{}}W-F1: \{Fb: 0.6343, \\ Tw: 0.5941\}\end{tabular} & $\geq92$ \\ \hline
\cite{oravsan2018aggressive} & \cite{kumar2018aggression} & En & 2018 & \begin{tabular}[c]{@{}l@{}}Glove, Emoticons TF-IDF, \\ Sentiment\end{tabular} & SVM, RF & \begin{tabular}[c]{@{}l@{}}W-F1: \{Fb: 0.5830 (RF),\\ Tw: 0.5074 (SVM)\}\end{tabular} & $\geq34$ \\ \hline
\cite{madisettyaggression} & \cite{kumar2018aggression} & En & 2018 & One-hot encoding & \begin{tabular}[c]{@{}l@{}}Majority voting: \{CNN, \\LSTM, and Bi-LSTM\}\end{tabular} & \begin{tabular}[c]{@{}l@{}}W-F1: \{Fb: 0.604,\\ Tw: 0.508\}\end{tabular} & $\geq43$ \\ \hline
\cite{fortuna2018merging} & \cite{kumar2018aggression} & En & 2018 & \begin{tabular}[c]{@{}l@{}}POS, Insults, Punctuation, \\ Sentiment, Capitalization\end{tabular} & RF & \begin{tabular}[c]{@{}l@{}}W-F1: \{Fb: 0.5288, \\ Tw: 0.3633\}\end{tabular} & $\geq33$ \\ \hline
\cite{arroyo2018cyberbullying} & \cite{kumar2018aggression} & En & 2018 & N-gram TF-IDF & \begin{tabular}[c]{@{}l@{}}Passive-aggressive \\ algorithm and SVM\end{tabular} & \begin{tabular}[c]{@{}l@{}}W-F1: \{Fb: 0.6315,\\ Tw: 0.3633\}\end{tabular} & $\geq31$ \\ \hline
\cite{risch2018aggression} & \cite{kumar2018aggression} & \begin{tabular}[c]{@{}l@{}}En, \\ Hi\end{tabular} & 2018 & \begin{tabular}[c]{@{}l@{}}English: FastText\\ Hindi: Wikipedia embeddings\\ Word-ngram, Char-ngram, \\ Hancrafted features\end{tabular} & \begin{tabular}[c]{@{}l@{}}Ensemble of GRU \\ with diffrent features\end{tabular} & \begin{tabular}[c]{@{}l@{}}W-F1: \{\\ Fb: \{En: 0.6011, Hi: 0.6311\},\\ Tw: \{En: 0.5995,Hi: 0.3835\}\}\end{tabular} & $\geq88$ \\ \hline
\cite{orabi2018cyber} & \cite{kumar2018aggression} & En & 2018 & Glove & MultiCNNPooling & \begin{tabular}[c]{@{}l@{}}W-F1: \{Fb: 0.5974,\\ Tw: 0.5690\}\end{tabular} & $\geq7$ \\ \hline
\cite{tommasel2018textual} & \cite{kumar2018aggression} & En & 2018 & \begin{tabular}[c]{@{}l@{}}LSTM: Glove, LSTM: Sentiment, \\ NN: Composed features (SVM: \\ Tf-IDF, Sentiment, Punctuation)\end{tabular} & \begin{tabular}[c]{@{}l@{}} Avg (SVM + \\NN (2LSTM+NN))\end{tabular} & \begin{tabular}[c]{@{}l@{}}W-F1: \{Fb: 0.5948, \\ Tw: 0.5480\}\end{tabular} & $\geq25$ \\ \hline
\cite{golem2018combining} & \cite{kumar2018aggression} & En & 2018 & \begin{tabular}[c]{@{}l@{}}POS, NER, Bad, Capitalization, \\ Numerical tokens, Text length, \\ Sentiment \\ BiLSTM, CNN: GloVe\end{tabular} & \begin{tabular}[c]{@{}l@{}}(BiLSTM+CNN+LR) \\ + SVM\end{tabular} & \begin{tabular}[c]{@{}l@{}}W-F1: \{ Fb: 0.642,\\ Tw: 0.601\}\end{tabular} & $\geq17$ \\ \hline
\cite{modha2018filtering} & \cite{kumar2018aggression} & \begin{tabular}[c]{@{}l@{}}En, \\ Hi\end{tabular} & 2018 & FastText & CNN, LSTM & \begin{tabular}[c]{@{}l@{}}W-F1: \{\\ Fb: \{En: 0.6178 (LSTM), \\ Hi: 0.6081\},\\ Tw: \{En: 0.5520,Hi: 0.4992\}\}\end{tabular} & $\geq54$ \\ \hline
\cite{singh2018aggression} & \cite{kumar2018aggression} & En & 2018 & \begin{tabular}[c]{@{}l@{}}Total, Avg words, Punctuation, \\ Abusive, URLs, Hashtags, Single \\ letters, Uppercase tokens, Ph. no.\end{tabular} & CNN & F1: 0.58 & $\geq43$ \\ \hline
\cite{patwa2021hater} & \cite{kumar2018aggression} & En & 2018 & Glove & CNN-CapsNet & W-F1: 0.6520 & $\geq4$ \\ \hline
\cite{aroyehun2018aggression} & \cite{kumar2018aggression} & En-Hi & 2018 & FastText & \begin{tabular}[c]{@{}l@{}}Fb: LSTM\\ Overall: CNN-LSTM\end{tabular} & \begin{tabular}[c]{@{}l@{}}W-F1: \{Fb: 0.6425 , \\ Overall: 0.5920\}\end{tabular} & $\geq155$ \\ \hline
\cite{gomez2018machine} & \cite{Aragn2020OverviewOM} & Es & 2018 & \begin{tabular}[c]{@{}l@{}}N-grams (Char, words, POS), \\ Aggressive tokens\end{tabular} & SVM & F1: 0.4285 & $\geq21$ \\ \hline
\cite{si2019aggression} & \cite{kumar2018aggression} & En-Hi & 2019 & \begin{tabular}[c]{@{}l@{}}POS, Emojis, Aggressive \\ tokens, Sentiment\end{tabular} & \begin{tabular}[c]{@{}l@{}}Majority voting: \\ (XGBoost, GBM, SVM)\end{tabular} & F1: 0.6813 & $\geq21$ \\ \hline
\cite{modha2022empirical} & \cite{kumar2018aggression} & En & 2019 & FastText & CNN & \begin{tabular}[c]{@{}l@{}}W-F1: \{\\ Fb: \{En: 0.5638, Hi: 0.6081\}, \\ Tw: \{En: 0.5377,Hi: 0.4992\}\}\end{tabular} & $\geq16$ \\ \hline
\cite{Tommasel2019AnES} & \cite{kumar2018aggression, Reynolds2011} & En & 2019 & \begin{tabular}[c]{@{}l@{}}TF-ID, Char tokens, Lemma, \\ NER, POS, Sentiment, Word2Vec, \\ GloVe, \cite{Barbieri2014}, \\ \cite{Farias2016}\end{tabular} & SVM-RBF & \begin{tabular}[c]{@{}l@{}}F1: \{\\ \cite{kumar2018aggression}- Fb: 0.6879,\\ \cite{kumar2018aggression}- Tw: 0.5529,\\ \cite{Reynolds2011}: 0.8416\}\end{tabular} & $\geq6$ \\ \hline
\cite{Srivastava2019DetectingAA} & \begin{tabular}[c]{@{}l@{}}\cite{kumar2018aggression} and \\ Toxic\end{tabular} & En & 2019 & FastText & \begin{tabular}[c]{@{}l@{}}BiLSTM followed \\ Capsule Network\end{tabular} & \begin{tabular}[c]{@{}l@{}}F1: \{Fb: 0.6353, \\ Tw: 0.5795\}\end{tabular} & $\geq12$ \\ \hline
\cite{GutirrezEsparza2019ClassificationOC} & \cite{GutirrezEsparza2019ClassificationOC} & Es & 2019 & VIMs & OneR & Acc.: 0.90 & $\geq37$ \\ \hline
\cite{kumari2019aggressive} & \cite{kumari2019aggressive} & Img & 2019 & - & CNN & \begin{tabular}[c]{@{}l@{}}P: 90, R, F1: 89\end{tabular} & $\geq23$ \\ \hline
\cite{SaneCorpusCA} & \cite{SaneCorpusCA} & En-Ta & 2019 & \begin{tabular}[c]{@{}l@{}}Char, word n-grams, \\ Lexicon, Punctuations, \\ Uppercase, Hashtags, \\ Mentions\end{tabular} & SVM-RBF & F1: 0.67 & 1 \\ \hline
\cite{ramiandrisoa2020aggression} & \cite{kumar2018aggression} & En & 2020 & Emoticon, Hashtags & BERT-base & \begin{tabular}[c]{@{}l@{}}W-F1: \{Fb: 0.627, Tw: 0.595\}\end{tabular} & $\geq5$ \\ \hline
\cite{kumar2020evaluating} & \cite{bhattacharya2020developing} & En & 2020 & Tokenizer & \begin{tabular}[c]{@{}l@{}}BERT-large + \\ Project attention layers\end{tabular} & \begin{tabular}[c]{@{}l@{}}W-F1: \{AG: 0.6367,\\ GEN: 0.8202\}\end{tabular} & $\geq114$ \\ \hline
\cite{liu2020scmhl5} & \cite{bhattacharya2020developing} & En & 2020 & - & \begin{tabular}[c]{@{}l@{}}Majority voting - 3 \\ uncased-BERT-base\end{tabular} & \begin{tabular}[c]{@{}l@{}}W-F1: \{AG: 0.664,\\ GEN: 0.851\}\end{tabular} & $\geq2$ \\ \hline
\cite{pascucci2020role} & \cite{bhattacharya2020developing} & En & 2020 & Stylistic, Linguistic rules & SMO & \begin{tabular}[c]{@{}l@{}}W-F1: \{AG: 0.6291,\\ GEN: 0.6733\}\end{tabular} & $\geq5$ \\ \hline
\cite{altin2020lastus} & \cite{bhattacharya2020developing} & En & 2020 & Word embedding \cite{barbieri2016cosmopolitan} & BiLSTM with attention & \begin{tabular}[c]{@{}l@{}}W-F1: \{AG: 0.7246,\\ GEN: 0.8199\}\end{tabular} & $\geq7$ \\ \hline
\cite{tawalbeh2020saja} & \cite{bhattacharya2020developing} & En & 2020 & USE \cite{cer2018universal} & XGBoost & \begin{tabular}[c]{@{}l@{}}W-F1: \{AG: 0.6075,\\ GEN: 0.8567\}\end{tabular} & $\geq5$ \\ \hline
\cite{rischbaggingBM} & \cite{bhattacharya2020developing} & \begin{tabular}[c]{@{}l@{}}En, Hi, \\ Bn\end{tabular} & 2020 & - & \begin{tabular}[c]{@{}l@{}}BERT, m-BERT (15-25) \\ Ensamble\end{tabular} & \begin{tabular}[c]{@{}l@{}}W-F1: \{\\ En: \{AG: 0.8029, GEN: 0.8514\},\\ Hi: \{AG: 0.8128, GEN: 0.8781\},\\ Bn: \{AG: 0.8219, GEN: 0.9385\}\}\end{tabular} & $\geq62$ \\ \hline
\cite{Baruah2020AggressionII} & \cite{bhattacharya2020developing} & \begin{tabular}[c]{@{}l@{}}En, Hi, \\ Bn\end{tabular} & 2020 & \begin{tabular}[c]{@{}l@{}}TF-IDF of word (1-3), \\ char (1-6) -ngram\end{tabular} & \begin{tabular}[c]{@{}l@{}}SVM (TF-IDF), BERT, \\ XLMRoBERTa\end{tabular} & \begin{tabular}[c]{@{}l@{}}W-F1: \{En: \{AG: 0.73 (BERT), \\GEN: 0.87 (SVM)\} \\ Hi: \{AG: 0.79 (SVM), \\ GEN: 0.87 (XLMRoBERTa)\},\\ Bn: \{AG: 0.81, GEN: 0.93\} (SVM)\}\end{tabular} & $\geq25$ \\ \hline
\cite{Datta2020SpyderAD} & \cite{bhattacharya2020developing} & \begin{tabular}[c]{@{}l@{}}En, Hi, \\ Bn\end{tabular} & 2020 & TF-IDF, Sentiment & \begin{tabular}[c]{@{}l@{}}XGBoost, Gradient \\ Boosting\end{tabular} & \begin{tabular}[c]{@{}l@{}}W-F1: \{En: 0.43 (XGBoost),\\ Hi: 0.59 (GBM),\\ Bn: 0.44 (GBM)\}\end{tabular} & $\geq7$ \\ \hline
\cite{Gordeev2020BERTOA} & \cite{bhattacharya2020developing} & \begin{tabular}[c]{@{}l@{}}En, Hi, \\ Bn\end{tabular} & 2020 & - & m-BERT-base & \begin{tabular}[c]{@{}l@{}}W-F1: \{\\ En: \{AG: 0.7568, GEN: 0.8716\},\\ Hi: \{AG: 0.7761, GEN: 0.7683\},\\ Bn: \{AG: 0.7761, GEN: 0.9297\}\}\end{tabular} & $\geq15$ \\ \hline
\cite{koufakou2020florunito} & \cite{bhattacharya2020developing} & \begin{tabular}[c]{@{}l@{}}En, Hi, \\ Bn\end{tabular} & 2020 & FastText, HurtLex & LSTM & \begin{tabular}[c]{@{}l@{}}W-F1: \{\\ En: \{AG: 0.677, GEN: 0.838\},\\ Hi: \{AG: 0.726, GEN: 0.771\},\\ Bn: \{AG: 0.746, GEN: 0.869\}\}\end{tabular} & $\geq12$ \\ \hline
\cite{Kumari2020ai} & \cite{bhattacharya2020developing} & \begin{tabular}[c]{@{}l@{}}En, Hi, \\ Bn\end{tabular} & 2020 & \begin{tabular}[c]{@{}l@{}}FastText, \\One-hot embedding\end{tabular} & LSTM, CNN & \begin{tabular}[c]{@{}l@{}}W-F1: \{\\ FastText-CNN: \{\\ En: \{AG: 0.6602, GEN: 0.8227\}\},\\ FastText-LSTM: \{ \\ Hi: \{AG: 0.6547, GEN: 0.7363\},\\ Bn: \{AG: 0.7175, GEN: 0.8793\}\}\end{tabular} & $\geq25$ \\ \hline
\cite{Mishra2020MultilingualJF} & \cite{bhattacharya2020developing} & \begin{tabular}[c]{@{}l@{}}En, Hi, \\ Bn\end{tabular} & 2020 & - & \begin{tabular}[c]{@{}l@{}}uncased-m-bert-base,\\ uncased-BERT-base\end{tabular} & \begin{tabular}[c]{@{}l@{}}W-F1: \{BERT: \{\\ En: \{AG: 0.759, GEN: 0.857\},\\ m-BERT: \{\\Hi: \{AG: 0.779, GEN: 0.849\},\\ Bn: \{AG: 0.780, GEN: 0.927\}\}\}\}\end{tabular} & $\geq26$ \\ \hline
\cite{Samghabadi2020AggressionAM} & \cite{bhattacharya2020developing} & \begin{tabular}[c]{@{}l@{}}En, Hi, \\ Bn\end{tabular} & 2020 & - & \begin{tabular}[c]{@{}l@{}}cased-m-bert-base,\\ uncased-BERT-base, \\ along with attention\end{tabular} & \begin{tabular}[c]{@{}l@{}}W-F1: \{BERT: \{\\ En: \{AG: 0.7143, GEN: 0.8579\}\},\\ m-BERT: \{\\Hi: \{AG: 0.7183, GEN: 0.8008\},\\ Bn: \{AG: 0.7369, GEN: 0.9206\}\}\}\end{tabular} & $\geq96$ \\ \hline
\cite{Khandelwal2020AUS} & \cite{kumar2018aggression, sadiq2021aggression} & \begin{tabular}[c]{@{}l@{}}En, \\ En-Hi\end{tabular} & 2020 & Glove and FastText & \begin{tabular}[c]{@{}l@{}}Avg (Deep Pyramid \\CNN + Pooled LSTM \\+ Disconnected RNN)\end{tabular} & \begin{tabular}[c]{@{}l@{}}W-F1: \{\\ \cite{kumar2018aggression}-Facebook: 0.6770, \\ \cite{kumar2018aggression}-Twitter: 0.6480 ,\\ \cite{sadiq2021aggression}: 0.9023\}\end{tabular} & $\geq7$ \\ \hline
\cite{Chen2018VerbalAD} & \cite{chen2017aggressivity} & En & 2020 & TF-IDF & CNN & \begin{tabular}[c]{@{}l@{}}Acc.: 0.92, \\ Micro-AUC: 0.98 \\ Macro-AUC: 0.97\end{tabular} & $\geq87$ \\ \hline
\cite{Tanase2020DetectingAI} & \cite{Aragn2020OverviewOM} & Es & 2020 & N-grams, TF-IDF & XLM-RoBERTa & F1: 0.7969 & $\geq19$ \\ \hline
\cite{GuzmanSilverio2020TransformersAD} & \cite{Aragn2020OverviewOM} & Es & 2020 & - & W-ensamble: 20 BERT & F1: 0.80 & $\geq24$ \\ \hline
\cite{Shrivastava2021EnhancingAD} & \cite{kumar2018aggression} & En-Hi & 2021 & \begin{tabular}[c]{@{}l@{}}TF-IDF followed \\ Data Balancing with \\ GPT-2\end{tabular} & \begin{tabular}[c]{@{}l@{}}Voting-Ensemble: \\ \{SVM, NB, LR, \\CNN, LSTM\}\end{tabular} & F1: 0.65 & $\geq9$ \\ \hline
\cite{Mundra2021EvaluationOT} & \cite{kumar2018aggression} & En-Hi & 2021 & \begin{tabular}[c]{@{}l@{}}Char-One-hot embedding,\\ Aggressive tokens\end{tabular} & CNN & \begin{tabular}[c]{@{}l@{}}P: 0.60, R: 0.59,\\ F1: 0.59, AUC: 0.59\end{tabular} & $\geq2$ \\ \hline
\cite{zhao2021comparative} & \cite{kumar2018aggression} & Fb: En & 2021 & - & XLM-e100 & micro-F1: 0.6196 & $\geq24$ \\ \hline
\cite{Kumari2021BilingualCD} & \cite{kumar2018aggression} & En & 2021 & \begin{tabular}[c]{@{}l@{}}Randomly initialised \\
 embedding\end{tabular} & LSTM-autoencoders & \begin{tabular}[c]{@{}l@{}}W-F1:\{
Fb: \{En: 0.81, Hi: 0.72\},\\
Tw: \{En: 0.71, Hi: 0.58\}\}\end{tabular} & $\geq20$ \\ \hline
\cite{dutta2021efficient} & \cite{bhattacharya2020developing} & En & 2021 & - & \begin{tabular}[c]{@{}l@{}}Data augmentation:\\ BERT, TF-IDF-XgBoost\end{tabular} & \begin{tabular}[c]{@{}l@{}}W-F1: \{AG: 0.735, \\ GEN: 0.852\}\end{tabular} & $\geq2$ \\ \hline
\cite{sadiq2021aggression} & \cite{sadiq2021aggression} & En & 2021 & TF-IDF (Uni+Bi-gram) & DNN - 3 layers & \begin{tabular}[c]{@{}l@{}}Acc: 0.92, P, R, F1: 0.90\end{tabular} & $\geq92$ \\ \hline
\cite{shulginov2021automatic} & \cite{shulginov2021automatic} & Ru & 2021 & - & RuBERT & F1: 0.66 & $\geq4$ \\ \hline
\cite{sharif2022tackling} & \cite{sharif2022tackling} & Bn & 2021 & Word2Vec & CNN-BiLSTM & W-F1: 0.87 & $\geq14$ \\ \hline
\cite{herodotou2021catching} & \cite{founta2018large} & En & 2021 & \begin{tabular}[c]{@{}l@{}}User-Profile, User-post (text), \\ Network Features\end{tabular} & \begin{tabular}[c]{@{}l@{}}Streeming-ML\\ (HT, ARF, SLR)\\ Batch-ML (DT)\end{tabular} & \begin{tabular}[c]{@{}l@{}}F1: \{Streeming: 0.88,\\ Batch: 0.91\}\end{tabular} & $\geq6$ \\ \hline
\cite{akter2022deep} & \cite{bhattacharya2020developing} & \begin{tabular}[c]{@{}l@{}}En, Hi, \\ Bn\end{tabular} & 2022 & DA: Machine Translation & \begin{tabular}[c]{@{}l@{}}BERT: En,\\ m-BERT: Hi, GPT2: Bn\end{tabular} & \begin{tabular}[c]{@{}l@{}}F1: \{En: .77, \\ Hi:0.68, Bn: 0.70\}\end{tabular} & $\geq6$ \\ \hline
\cite{sengupta2022does} & \cite{kumar2018aggression} & En-Hi & 2022 & - & Char-HAL & W-F1: 0.909 & $\geq20$ \\ \hline
\cite{Ramiandrisoa2022MultitaskLF} & \cite{kumar2018aggression, bhattacharya2020developing} & En & 2022 & - & MTL-RoBERTa & \begin{tabular}[c]{@{}l@{}}W-F1: \{ \cite{kumar2018aggression}: 0.64, \cite{bhattacharya2020developing}: 0.73\}\end{tabular} & $\geq1$ \\ \hline
\cite{sharif2022tackling} & \cite{sharif2022tackling} & Bn & 2022 & - & \begin{tabular}[c]{@{}l@{}}W-ensemble: \{m-BERT, \\ distil-BERT, Bangla-BERT, \\ XLM-R\}\end{tabular} & \begin{tabular}[c]{@{}l@{}}W-F1: \{AG: 0.9343, \\ GEN: 0.9311\}\end{tabular} & $\geq28$ \\ \hline
\cite{khan2022aggression} & \cite{sadiq2021aggression} & En & 2022 & 7-Emotions + Word2Vec & DNN & F1: 0.97 & $\geq12$ \\ \hline
\cite{kumari2022multi} & \cite{kumari2022multi} & En & 2022 & Glove, BFFO & \begin{tabular}[c]{@{}l@{}}Text: CNN + \\ Img: VGG-16 +RF\end{tabular} & \begin{tabular}[c]{@{}l@{}}W-P: 0.80, R, F1: 0.79\end{tabular} & $\geq7$ \\ \hline
\cite{tonja2022detection} & \cite{arellano2022overview} & Es & 2022 & - & DistilBETO & \begin{tabular}[c]{@{}l@{}}F1: \{AG: 0.7455, \\ AG-Cat: 0.4903\}\end{tabular} & $\geq8$ \\ \hline
\cite{ta2022multi} & \cite{arellano2022overview} & Es & 2022 & DA: Machine Translation & m-BERT-base & \begin{tabular}[c]{@{}l@{}}P: 0.7552, R: 0.7409, \\ F1: 0.7480\end{tabular} & $\geq2$ \\ \hline
\cite{ta2022gan} & \cite{arellano2022overview} & Es & 2022 & DA: Machine Translation & GAN-BERT \cite{croce2020gan} & \begin{tabular}[c]{@{}l@{}}P: 0.7408, R: 0.7479, \\ F1: 0.7443\end{tabular} & $\geq4$ \\ \hline
\cite{huerta2022verbal} & \cite{Aragn2020OverviewOM} & Es & 2022 & \begin{tabular}[c]{@{}l@{}}Structural, Affective, \\ Dimensional, Emotional, \\ Word2Vec\end{tabular} & SVM & macro-F1: 0.7921 & $\geq1$ \\ \hline
\cite{agbaje2022neural} & \cite{Sentiment140} & En & 2022 & Sentiments & \begin{tabular}[c]{@{}l@{}}Text: LSTM\\ Img: CNN\end{tabular} & \begin{tabular}[c]{@{}l@{}}F1: \{Tw: 0.91, \\ Img: 0.89\}\end{tabular} & $\geq1$ \\ \hline
\cite{gattulli2022cyber} & \cite{gattulli2022cyber} & It & 2022 & Syntactic, Sentiment & RF & Avg-Acc: 0.9125 & $\geq2$ \\ \hline
\cite{Iqbal2019UsingCL} & \cite{kumar2018aggression} & En & 2023 & \begin{tabular}[c]{@{}l@{}}N-grams, LIWC, \\ POS and Sentiments\end{tabular} & RF & F1: 0.67 & $\geq2$ \\ \hline
\cite{mane2023you} & \cite{mane2023you} & En & 2023 & FastText & BiLSTM & \begin{tabular}[c]{@{}l@{}}Acc: 0.81, P:0.77, \\R:0.83, W-F1:0.81\end{tabular}  & $\geq0$ \\ \hline
\cite{ghosh2023transformer} & \cite{bhattacharya2020developing} & \begin{tabular}[c]{@{}l@{}}En, Hi, \\ Bn\end{tabular} & 2023 & - & \begin{tabular}[c]{@{}l@{}}XLM-RoBERTa (XLMR),\\ Multi-head self-attention\end{tabular} & \begin{tabular}[c]{@{}l@{}}W-F1: \{\\ En: \{AG: 0.6453, GEN: 0.8612\},\\ Hi: \{AG: 0.7776, GEN: 0.8657\},\\ Bn: \{AG: 0.7835, GEN: 0.9053\}\}\end{tabular} & $\geq0$ \\ \hline
\cite{rawat2023modelling} & \cite{kumar2018aggression, rawat2023modelling} & En & 2023 & - & \begin{tabular}[c]{@{}l@{}}Political: RoBERTa-base\\ \cite{kumar2018aggression}: Hing-BERT\end{tabular} & \begin{tabular}[c]{@{}l@{}}Macro-F1: \{ Political: 0.66,\\ Political+\cite{kumar2018aggression}: 0.92\}\end{tabular} & $\geq0$ \\ \hline
\end{longtable}
}

\section{Behavioral Analysis of User Aggressiveness}
\label{6_aggressive_behaviour}
\begin{figure}[h!]
\centering
\includegraphics[scale=0.28]{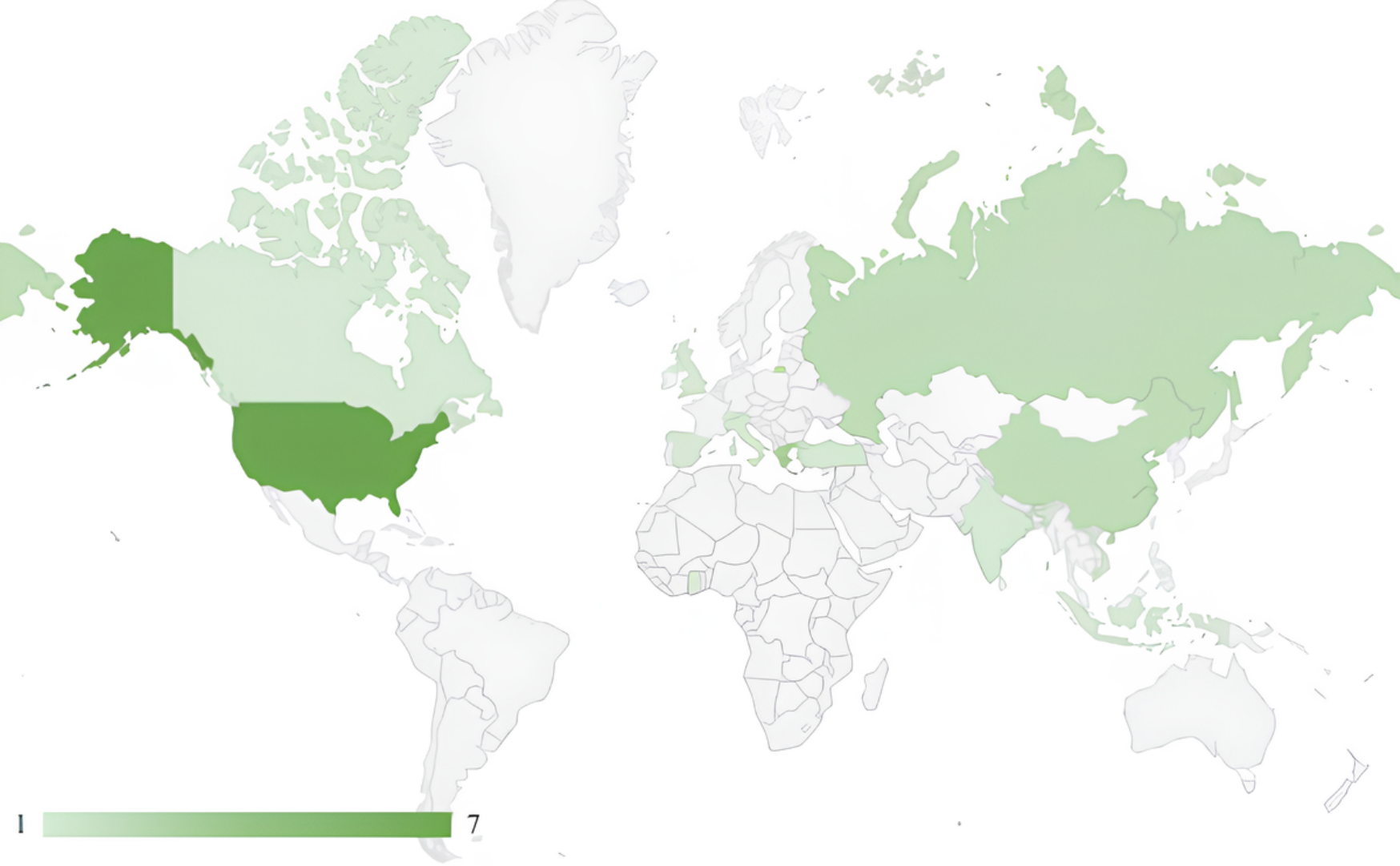}
\caption{Distribution of studies on the behavioral analysis of aggressive users across various countries. The United States leads in the number of studies, followed by Greece and Russia, highlighting a research gap in other regions.} 
\label{world_heatmap}
\vspace*{-1\baselineskip}
\end{figure}

Behavioral analysis of user aggressiveness on social media is addressed through a multifaceted approach, using both qualitative and quantitative methodologies.
Qualitative studies explore the emotional and psychological aspects, offering insights into underlying motivations and contextual factors. Quantitative research quantifies the prevalence and patterns of online aggression, providing empirical evidence to monitor trends.
Figure \ref{world_heatmap} presents the distribution of the number of behavioral studies conducted in different countries.

\subsection{Qualitative Studies: Unraveling the `Why' of Online Aggression} 
Qualitative studies rely on data collected from surveys and interviews involving diverse participants.
\subsubsection{Influencing factors}
Various studies have been conducted to examine the factors that influence aggressive behavior on social media. 
Notable factors such as counter-comments \citep{eraslan2019social}, specific topics or events \citep{pascual2021toxicity}, peer or friend influence \citep{sobkin2021adolescents, henneberger2017effect}, anonymity \citep{zimmerman2016online}, religious influence \cite{boadi2021social}, the impact of fake media \cite{galyashinafake}, moral emotions \cite{adinugroho2022understanding}, social media usage patterns \citep{wong2022association, borraccino2022problematic, ferguson2021does, adinugroho2022understanding}, gender stereotype traits \citep{wright2020role}, and person and situational factors \cite{allen2018general, weingartner2019online}. 
A study by \citet{eraslan2019social} revealed that counter-comments about a user's values substantially affect the user's exhibition of aggressive tendencies.  
Religion can also have an impact on online aggression behaviors, according to \citet{boadi2021social}. The researchers used \textit{Contemporary Deterrence} theory to examine how religion influences an individual's daily life and captures behavioral outcomes or its direct influences on online aggression behaviors. 
Additionally, \citet{borraccino2022problematic} found that cyber aggression was more prevalent among girls, and problematic social media use was more prevalent among adolescents.
In \citet{wright2020role} work, it has been shown that girls with more feminine characteristics are more likely to engage in cyber-aggression. In contrast, those endorsing more masculine traits reported cyber relational aggression and cyber verbal aggression more frequently through online gaming. In contrast, those with more masculine traits reported experiencing or using cyber-aggression more often.
\citet{sobkin2021adolescents} examined the behavior of teenagers and revealed that those who have experienced real-life bullying are more likely to be involved in aggressive situations on both sides of social media.

A study conducted by \citet{ferguson2021does} found that social media may contribute to adverse outcomes, but the relationship (media \& aggression) is complex and dependent on individual motivation and situational factors. 
\citet{pascual2021toxicity} investigated aggression in online discourse, particularly related to wearing face masks during the COVID-19 pandemic. The study found that tweets with anti-mask hashtags used more aggressive language, while tweets with pro-mask hashtags displayed a relatively less aggressive tone.
Moreover, \citet{galyashinafake} examined the potential aggressive impact of fake media on social media users, raising questions about the safety of social network communication. 
In \cite{chaturvedi2022} study, qualitatively analyzed transitive and lattice graphs representing the co-occurrence of aggressive (hate) terms in contextual data related to women and regions. 
A study by \citet{adinugroho2022understanding} identified two predictors of cyber aggression: moral emotions and the frequency of social media usage. Their study revealed that guilt inhibits cyber-aggression, while shame encourages it.
\citet{weingartner2019online} developed a comprehensive theoretical framework considering various factors that contribute to cyber-aggressive behavior. This model includes socio-structural determinants, individual determinants, and situational determinants.
\citet{runions2015online} disclosed that cyber-aggressive behavior can be influenced by various factors such as the influence of aggressive peers, the influence of the initial aggressor on their network, and the cascading diffusion trend from the source of the behavior to its connections. Furthermore, \citet{zimmerman2016online} found that anonymous users were more aggressive than non-anonymous users. \citet{zeng2023effects} explored the effect of an interpretation bias modification program on cyber-aggression among adolescents, specifically focusing on modifying hostile interpretation bias. Results indicated that the program effectively reduced reactive cyber-aggression, but this effect was more pronounced in females than males.

\subsubsection{Consequences and Impact}
The consequences of aggressive behavior on social media are of paramount concern due to its potential to disrupt public stability and, in some cases, escalate into real-world violence and aggression \cite{quang2021some}. 
A study \citet{torregrosa2022mixed} demonstrated that the tone of electoral campaign discourse may not be as negative as perceived by citizens. The study confirms that the most ideologically extreme parties tend to use more aggressive language than normal or moderate parties. 
The experience of online aggression among university students was investigated by \citet{mishna2018social}, and it was found that a minority of students experienced online aggression. 
The study conducted by \citet{wong2022association} found a relationship between the motivation to seek rewards through aggressive behavior and the emergence of cyberbullying, which can lead to social media addiction. Their findings indicated that inclusive norms may reduce cyberbullying but increase social media addiction. Cyberbullying has been found to be indirectly associated with social media addiction to increased aggression. 
Furthermore, \citet{vladimirou2021aggressive} explored how complaining is expressed on social media and how it can become more aggressive due to complex participation and multimodality. A case study centered on an online protest against a waste management policy. Initially, users expressed their dissatisfaction with the policy on social media. However, as the protest continued, the tone of the posts became increasingly aggressive, with participants resorting to profanity, insults, and threats. This aggressive behavior on social media eventually spilled over into the real world, disrupting public stability.

\subsection{Quantitative Studies: Mapping the `What' and `How' of Online Aggression}
Quantitative studies employ statistical analyses, machine learning techniques, and network science approaches to quantify and model aggressive behavior on social media platforms.

\subsubsection{Comprehensive Frameworks and Models}
The General Aggression Model (GAM) \cite{anderson2002human} is a framework that explores the person and situational factors that influence cognitions, feelings, and arousal, leading to aggressive behavior. \citet{allen2018general} applied the GAM to understand aggressive behavior on social media and found that factors such as anonymity, disinhibition, and social comparison contribute to aggression.
Similarly, \citet{mathew2019spread} utilized the DeGroot model to identify aggressive and non-aggressive users on social media. The DeGroot model is a mathematical framework that analyzes the spread of information within a network. \citet{mathew2019spread} found that the posts of aggressive users tend to spread faster and wider, indicating the potential for hate speech to impact social media significantly.
The Social Information Processing Theory (SIPT) \cite{walther2008social} suggests that aggressive behavior is a result of a lack of social cues on social media, leading to misinterpretation and misunderstanding. The Frustration-Aggression Theory suggests that frustration leads to aggression, which can be exacerbated by social media platforms providing a platform for anonymous venting.
\citet{chaturvedi2022} identified severe aggressive terms quantitatively by applying a proposed threshold of an offensive score derived from inter-agreement measures.

\subsubsection{Network and User-based Approaches.}
The networked nature of social media platforms necessitates approaches that consider both individual user characteristics and broader network dynamics in understanding and addressing online aggression. \citet{terizi2021modeling} proposed a method that uses opinion dynamics based on network dynamics to study the propagation of aggression on social media. They analyzed how users' connections to non-aggressive users affected the spread of aggression on the platform.
Similarly, \citet{gallo2020predicting} used a network knowledge-based model to identify online users' reactions on Twitter using a machine learning approach. It explored user beliefs after exposure to their news feed and identified several social cues as features to predict whether a user would react in a given time period.
Another study by \citet{chatzakou2017mean} proposed a methodology to detect aggressive behavior users based on text, user, and network-based features. They analyzed user behavior and found that aggressors are relatively popular and tend to include more negative posts.
\citet{ali2023social} conducted a study to identify influential users in the aggressive community using the Girvan Newman algorithm. In their method, they utilized an LSTM-GRU model to effectively detect aggressive tweets. 
\citet{basak2019online} identified the characteristics of online shaming, such as the use of negative words and retweeting, and proposed a machine learning algorithm to detect it.
Similarly, \citet{Karan2023} identified patterns related to social media features and the percentage of aggressive posts among aggressor and target users using LSTM.

A study by \citet{balci2015automatic}, analyzed verbal aggression and abusive behaviors of users on online social games. They analyzed user behavior and characteristics of offending users to identify patterns and potential solutions.
Further, \citet{aragon2019overview} provided an overview of MEX-A3T, a shared task for identifying user profiles and aggressive tweets in Mexican Spanish. The researchers developed the user profile and aggression detection dataset with baseline results from \cite{Ortega} and \cite{Graff2018}, respectively.
\citet{poiitis2021aggression} focused on developing a model for predicting the diffusion of aggressive tweets on Twitter. Their proposed solution aimed to minimize the spread of such content by identifying influential users and selectively blocking their actions.
\citet{mane2023you} introduced a novel metric, ``user aggression intensity,'' to measure overall aggressive activity.  Their findings indicate that the content in social media feeds can influence aggressive behavior, especially with event-specific feeds leading to increased aggression. Additionally, the study found that users support and encourage aggressive content on social media.

\section{Socio-Computational Aggressive Behavior Analysis}
\label{socio-comptational}
The integration of sociological insights with computational techniques offers a promising approach to understanding and mitigating online aggression. This interdisciplinary synergy leverages the knowledge of human behavior provided by sociology and the scalable, data-driven methods of computer science. As illustrated in Table \ref{tab:socio-comput}, key sociological findings can inform various computational methods, enhancing their effectiveness in addressing online aggression. For instance, \citet{zimmerman2016online} work on anonymity's role in fostering aggressive behavior suggests incorporating user anonymity levels as features in detection algorithms. Similarly, \citet{sobkin2021adolescents} findings on peer influence in online aggression highlight the potential of including social network metrics in predictive models. The contextual nature of online aggression, as demonstrated by \citet{pascual2021toxicity} in their study of topic-specific aggression underscores the need for context-aware detection models. By integrating these sociological insights, computational techniques can evolve beyond simple keyword-based approaches to more sophisticated methods that consider broader conversational and social contexts. Network-based interventions represent another promising area where sociological insights enhance computational techniques. Studies like \citet{terizi2021modeling} work on opinion dynamics models provide valuable information on how aggressive behavior propagates online, informing targeted intervention strategies that focus on influential users or key network nodes.

\begin{table}[h!]
\centering
\caption{Integration of content detection and behavioral analysis findings.}
\label{tab:socio-comput}
\resizebox{\textwidth}{!}{%
\begin{tabular}{lll}
\hline
\multicolumn{1}{c}{\textbf{Content Detection Insights}} & \multicolumn{1}{c}{\textbf{Behavioral Analysis Findings}} & \multicolumn{1}{c}{\textbf{Integrated Approach}} \\ \hline
Linguistic markers of aggression. & Psychological factors influencing aggressive behavior. & Using linguistic patterns to infer psychological states. \\
Temporal patterns of aggressive content. & Social contagion effects in online aggression. & Predictive models for the spread of aggressive behavior. \\
User network characteristics. & Influence of anonymity and social norms. & Network-based intervention strategies. \\
Multimodal aggression detection. & Impact of platform design on user behavior. & Platform-specific aggression mitigation techniques. \\ \hline
\end{tabular}%
}
\end{table}

\citet{wright2020role} research on gender stereotype traits and cyber-aggression can inform user models that predict susceptibility to online aggression, enabling tailored interventions. Studies by \citet{wong2022association} and \citet{borraccino2022problematic} on the relationship between platform features, usage patterns, and aggressive behavior offer valuable guidance for creating online environments that discourage aggression. 
Moreover, computational methods provide quantitative insights into behavioral patterns from large-scale data, which are often missed by traditional sociological methods. These algorithms analyze social media data to track trends in online aggression and the spread of aggressive behavior. They also help validate sociological theories like peer influence and social contagion. By offering real-time data analysis, computational models enable a deeper understanding of aggression's evolution in online communities.
Table \ref{tab:mapping} presents a mapping of key sociological insights to effective computational methods. As this interdisciplinary field evolves, it must address critical challenges such as ethical considerations in data use and cultural sensitivity in global contexts. Researchers can develop more effective, nuanced, and ethically sound strategies for creating safer and more positive online environments by combining sociological insights with computational techniques.

\begin{table}[h!]
\centering
\caption{Mapping key sociological insights to effective computational methods for addressing online aggression and their potential applications in social media contexts.}
\label{tab:mapping}
\resizebox{\textwidth}{!}{%
\begin{tabular}{lll}
\hline
\multicolumn{1}{c}{\textbf{Sociological Insight}} & \multicolumn{1}{c}{\textbf{Computational Method}} & \multicolumn{1}{c}{\textbf{Potential Application}} \\ \hline
Anonymity fosters aggressive behavior \cite{zimmerman2016online}. & Feature engineering, User profiling. & Incorporate anonymity levels in detection algorithms. \\  
Peer influence impacts online aggression \cite{sobkin2021adolescents}. & Social network analysis, Graph neural networks. & Develop influence-aware intervention strategies. \\ 
Topic-specific triggers for aggression \cite{pascual2021toxicity}. & Topic modeling, Contextual embeddings. & Create context-sensitive detection models. \\
Gender stereotype traits affect cyber-aggression \cite{wright2020role}. & Personalized machine learning models. & Tailor intervention approaches to individual characteristics. \\ 
Platform design influences aggressive behavior \cite{wong2022association}. & Simulation models, A/B testing. & Optimize platform features to minimize aggression. \\ 
Temporal patterns in aggressive behavior \cite{terizi2021modeling}. & Time series analysis, Recurrent neural networks. & Predict and prevent escalation of aggressive behavior. \\
Cross-platform aggression dynamics. & Transfer learning, Multi-task learning. & Develop cross-platform detection and intervention strategies. \\ 
Cultural differences in communication norms. & Domain adaptation, Multilingual models. & Create culturally sensitive detection algorithms. \\ \hline
\end{tabular}%
}
\end{table}

\section{Future Scope and Research Directions}
\label{7_research_gap}
Despite the considerable research conducted on aggression content detection and behavioral analysis of user aggression on social media, several critical challenges and opportunities remain. 
This section describes critical areas for future investigation, presenting specific research questions to advance the field of 
\\
\textbf{RQ1:} \textit{How can we develop comprehensive datasets that capture the full spectrum of online aggression?}
The existing datasets have provided valuable insights, they often fall short in capturing the full spectrum of aggression types and contexts across languages and diverse social media platforms (see Table \ref{tab:dataset}). To address this, future research should focus on:
Creating large-scale, diverse language datasets encompassing a wide range of aggression types, including subtle forms of covert aggression, across multiple social media platforms.
Generating platform-specific datasets that account for unique characteristics (e.g., writing style, post length) to enhance the generalizability of detection algorithms.
\\
\textbf{RQ2:} \textit{How can we effectively integrate multimodal and multilingual data to improve aggression detection performance?}
Current research predominantly focuses on English text-based aggression detection (as evidenced in Table \ref{tab:detection_table}). To address this limitation, we propose:
Developing robust multilingual datasets, with an emphasis on code-mixed languages prevalent in social media communications.
Expanding research to incorporate visual, audio, and network data alongside textual content for a more comprehensive understanding of online aggression.
\\
\textbf{RQ3:} \textit{How can advanced Large Language Models (LLMs) be leveraged to enhance aggression detection while maintaining interpretability and fairness?}
Large-scale language models (LLMs) have shown promising results in related areas like hate speech, offensive language, toxic language and anomaly detection \cite{li2023hot, roy-etal-2023-probing, elhafsi2023}. However, their potential for addressing aggression detection has not been widely explored yet.
To address this, we suggest:
Investigate how LLMs such as Llama, Mistral, Gemma, and Gemini-Pro can be fine-tuned or adapted to improve the performance and interpretability of aggression detection algorithms.
Conduct rigorous comparisons between traditional transformer models and newer LLMs to quantify improvements in aggression detection accuracy and nuance.
\\
\textbf{RQ4:} \textit{How can sociological insights be effectively integrated into computational models of online aggression?}
The field of sociology offers valuable theoretical frameworks and empirical findings that can significantly enhance our understanding of online aggression. These sociological factors can be validated through quantitative research and leveraged in studies analyzing behavior propagation and predicting future aggressive behavior based on network dynamics and qualitative findings. To address this, we propose:
Forming interdisciplinary collaborations between computer scientists and sociologists to develop more nuanced and contextually aware aggression detection models.
Incorporating sociological theories of aggression and social interaction into the design of computational models and feature selection processes.
\\
\textbf{RQ5:} \textit{What are the cross-cultural differences in online aggression manifestation, and how do they evolve over time?}
To address the limitations of current research, which often focuses on specific cultural contexts (see Figure \ref{world_heatmap}) and short-term observations, we recommend:
Conducting cross-cultural comparative analyses to examine how cultural factors (e.g., region, religious beliefs, social norms) influence the manifestation and perception of online aggression across different societies.
Implementing longitudinal studies to analyze the evolution of aggressive behavior patterns over extended periods, considering factors such as major events, platform policy changes, and societal shifts.
\\
\textbf{RQ6:} \textit{How can we develop aggression detection models that are robust to evolving language patterns and emerging forms of online aggression?}
Current aggression detection techniques predominantly target overt forms of aggression, while covert expressions remain challenging to identify effectively \citep{kumar2018aggression}. To move beyond surface-level text analysis, future research should focus on:
Developing algorithms that incorporate user relationships, interaction history, and platform-specific contextual cues to improve aggression detection performance, particularly for subtle yet pervasive forms of aggression.
Analyzing the propagation patterns of aggressive content within social networks to identify influential nodes and understand the mechanisms of aggression spread.
\\
\textbf{RQ7:} \textit{How can we ensure transparency and fairness in AI-driven aggression detection systems?}
The development of transparent and interpretable models is crucial for building trust and identifying potential biases. We propose:
Designing aggression detection algorithms that clearly explain their decisions facilitates better understanding and potential intervention strategies.
Implementing techniques to identify and mitigate biases in aggression detection models, ensuring fair and equitable application across diverse user groups.
\\
\textbf{RQ8:} \textit{What standardized metrics and evaluation frameworks can be developed to assess aggression detection algorithms comprehensively?}
To address the lack of standardized metrics in the field, we recommend:
Developing a set of standardized metrics that assess not only the accuracy of aggression detection algorithms but also their fairness, robustness, and generalizability across different contexts.
Creating benchmark datasets and organizing community challenges to facilitate direct comparisons between different approaches and drive innovation in the field.
\\
\textbf{RQ9:} \textit{What are the most effective intervention strategies for reducing aggressive behavior on social media, and how do their impacts vary across different user demographics and cultural contexts?}
Moving beyond detection, future research should focus on developing effective intervention strategies:
Creating and evaluating real-time intervention techniques that can mitigate aggressive behavior as it occurs on social media platforms.
Developing user-facing tools that provide individuals with resources to manage online conflict, report abuse, and foster positive interactions within their online communities.

By addressing these research directions, the field can move towards effective approaches to understanding and mitigating online aggression, ultimately contributing to safer and more positive online environments.

\section{Conclusion}
\label{8_conclusion}
In this systematic literature review, we have effectively bridged the gap between two disparate yet inherently interconnected studies, namely aggression content detection and aggressive behavior analysis. In this literature, we proposed a unified cyber-aggression definition. We thoroughly examined the components of aggression content detection, including datasets, features, and detection algorithms. Despite the progress made, our review underscores the scarcity of multimodal, multilingual, and social media-specific datasets. The incorporation of multimodal features has shown promise in enhancing aggression detection. An important observation is the dominance of research in English-language aggression detection, leaving a notable gap in the study of low-resource language aggression detection. We observed that transformer-based BERT variants outperform traditional ML and DL algorithms. Overall, ensemble-based approaches including the transformer algorithm consistently performed better in detecting aggression.  
Further, we discussed the ethical considerations and potential biases in the field. Additionally, we systematically present meta-information on aggression detection studies in Table \ref{tab:detection_table}. In our exploration of aggressive user behavior analysis, we extensively investigate both qualitative and quantitative studies. 
This literature review reveals significant sociological insights that can enhance computational techniques, potentially improving the effectiveness of cyber-aggression prevention methods. 
Finally, we have identified specific challenges and provided future research directions, highlighting a shift toward socio-computational approaches. This work encourages researchers to adopt a socio-computational perspective in addressing the widespread issue of cyber-aggressive behavior. 

\medskip

\begin{acks}
This research was partially supported by the Prime Minister Research Fellowship funded by the Ministry of Education (MoE), India. Suman Kundu would like to acknowledge grant no. 4(2)/2024-ITEA of MeitY, GoI; and Srijan: Center for Generative AI (grant no. ET/23/2024-ET) of MeitY under the IndiaAI mission with the support of Meta for partial support. Rajesh Sharma is supported by the EU H2020 program under the SoBigData++ project (grant agreement No. 871042) and partially funded by the CHIST-ERA project HAMISON.
\end{acks}


\bibliographystyle{ACM-Reference-Format.bst}
\bibliography{references}


\begin{thebibliography}{182}


\ifx \showCODEN    \undefined \def \showCODEN     #1{\unskip}     \fi
\ifx \showDOI      \undefined \def \showDOI       #1{#1}\fi
\ifx \showISBNx    \undefined \def \showISBNx     #1{\unskip}     \fi
\ifx \showISBNxiii \undefined \def \showISBNxiii  #1{\unskip}     \fi
\ifx \showISSN     \undefined \def \showISSN      #1{\unskip}     \fi
\ifx \showLCCN     \undefined \def \showLCCN      #1{\unskip}     \fi
\ifx \shownote     \undefined \def \shownote      #1{#1}          \fi
\ifx \showarticletitle \undefined \def \showarticletitle #1{#1}   \fi
\ifx \showURL      \undefined \def \showURL       {\relax}        \fi
\providecommand\bibfield[2]{#2}
\providecommand\bibinfo[2]{#2}
\providecommand\natexlab[1]{#1}
\providecommand\showeprint[2][]{arXiv:#2}

\bibitem[Adinugroho et~al\mbox{.}(2022)]%
        {adinugroho2022understanding}
\bibfield{author}{\bibinfo{person}{Indro Adinugroho}, \bibinfo{person}{Priska
  Kristiani}, {and} \bibinfo{person}{Nani Nurrachman}.}
  \bibinfo{year}{2022}\natexlab{}.
\newblock \showarticletitle{Understanding Aggression in Digital Environment:
  Relationship between Shame and Guilt and Cyber Aggression in Online Social
  Network}.
\newblock \bibinfo{journal}{\emph{Makara Human Behavior Studies in Asia}}
  \bibinfo{volume}{26}, \bibinfo{number}{2} (\bibinfo{year}{2022}),
  \bibinfo{pages}{105--113}.
\newblock


\bibitem[AGBAJE and Afolabi(2022)]%
        {agbaje2022neural}
\bibfield{author}{\bibinfo{person}{MICHAEL AGBAJE} {and}
  \bibinfo{person}{Oreoluwa Afolabi}.} \bibinfo{year}{2022}\natexlab{}.
\newblock \showarticletitle{Neural Network-Based Cyber-Bullying and
  Cyber-Aggression Detection Using Twitter Text}.
\newblock \bibinfo{journal}{\emph{Research Square}}  \bibinfo{volume}{1}
  (\bibinfo{year}{2022}).
\newblock
\urldef\tempurl%
\url{https://doi.org/10.21203/rs.3.rs-1878604/v1}
\showDOI{\tempurl}


\bibitem[Akter et~al\mbox{.}(2022)]%
        {akter2022deep}
\bibfield{author}{\bibinfo{person}{M. Akter}, \bibinfo{person}{H. Shahriar},
  \bibinfo{person}{N. Ahmed}, {and} \bibinfo{person}{A. Cuzzocrea}.}
  \bibinfo{year}{2022}\natexlab{}.
\newblock \showarticletitle{Deep Learning Approach for Classifying the
  Aggressive Comments on Social Media: Machine Translated Data Vs Real Life
  Data}. In \bibinfo{booktitle}{\emph{2022 IEEE International Conference on Big
  Data (Big Data)}}. \bibinfo{publisher}{IEEE Computer Society},
  \bibinfo{address}{Los Alamitos, CA, USA}, \bibinfo{pages}{5646--5655}.
\newblock
\urldef\tempurl%
\url{https://doi.org/10.1109/BigData55660.2022.10020249}
\showDOI{\tempurl}


\bibitem[Al-Alosi(2017)]%
        {al2017cyber}
\bibfield{author}{\bibinfo{person}{Hadeel Al-Alosi}.}
  \bibinfo{year}{2017}\natexlab{}.
\newblock \showarticletitle{Cyber-violence: digital abuse in the context of
  domestic violence.}
\newblock \bibinfo{journal}{\emph{TheUNIVERSITY OF NEW SOUTH WALES LAW
  JOURNAL}} \bibinfo{volume}{40}, \bibinfo{number}{4} (\bibinfo{year}{2017}),
  \bibinfo{pages}{1573--1603}.
\newblock


\bibitem[Ali et~al\mbox{.}(2023)]%
        {ali2023social}
\bibfield{author}{\bibinfo{person}{Mohsan Ali}, \bibinfo{person}{Mehdi Hassan},
  \bibinfo{person}{Kashif Kifayat}, \bibinfo{person}{Jin~Young Kim},
  \bibinfo{person}{Saqib Hakak}, {and} \bibinfo{person}{Muhammad~Khurram
  Khan}.} \bibinfo{year}{2023}\natexlab{}.
\newblock \showarticletitle{Social media content classification and community
  detection using deep learning and graph analytics}.
\newblock \bibinfo{journal}{\emph{Technological Forecasting and Social Change}}
   \bibinfo{volume}{188} (\bibinfo{year}{2023}), \bibinfo{pages}{122252}.
\newblock


\bibitem[Allen et~al\mbox{.}(2018)]%
        {allen2018general}
\bibfield{author}{\bibinfo{person}{Johnie~J Allen}, \bibinfo{person}{Craig~A
  Anderson}, {and} \bibinfo{person}{Brad~J Bushman}.}
  \bibinfo{year}{2018}\natexlab{}.
\newblock \showarticletitle{The general aggression model}.
\newblock \bibinfo{journal}{\emph{Current opinion in psychology}}
  \bibinfo{volume}{19} (\bibinfo{year}{2018}), \bibinfo{pages}{75--80}.
\newblock


\bibitem[Anderson and Bushman(2001)]%
        {anderson2001effects}
\bibfield{author}{\bibinfo{person}{Craig~A Anderson} {and}
  \bibinfo{person}{Brad~J Bushman}.} \bibinfo{year}{2001}\natexlab{}.
\newblock \showarticletitle{Effects of violent video games on aggressive
  behavior, aggressive cognition, aggressive affect, physiological arousal, and
  prosocial behavior: A meta-analytic review of the scientific literature}.
\newblock \bibinfo{journal}{\emph{Psychological science}} \bibinfo{volume}{12},
  \bibinfo{number}{5} (\bibinfo{year}{2001}), \bibinfo{pages}{353--359}.
\newblock


\bibitem[Anderson and Bushman(2002)]%
        {anderson2002human}
\bibfield{author}{\bibinfo{person}{Craig~A Anderson} {and}
  \bibinfo{person}{Brad~J Bushman}.} \bibinfo{year}{2002}\natexlab{}.
\newblock \showarticletitle{Human aggression}.
\newblock \bibinfo{journal}{\emph{Annual review of psychology}}
  \bibinfo{volume}{53}, \bibinfo{number}{1} (\bibinfo{year}{2002}),
  \bibinfo{pages}{27--51}.
\newblock


\bibitem[Arag{\'{o}}n et~al\mbox{.}(2019)]%
        {aragon2019overview}
\bibfield{author}{\bibinfo{person}{Mario~Ezra Arag{\'{o}}n},
  \bibinfo{person}{Miguel {\'{A}}ngel~{\'{A}}lvarez Carmona},
  \bibinfo{person}{Manuel Montes{-}y{-}G{\'{o}}mez}, \bibinfo{person}{Hugo~Jair
  Escalante}, \bibinfo{person}{Luis~Villase{\~{n}}or Pineda}, {and}
  \bibinfo{person}{Daniela Moctezuma}.} \bibinfo{year}{2019}\natexlab{}.
\newblock \showarticletitle{Overview of {MEX-A3T} at IberLEF 2019: Authorship
  and Aggressiveness Analysis in Mexican Spanish Tweets}. In
  \bibinfo{booktitle}{\emph{Proceedings of the Iberian Languages Evaluation
  Forum co-located with 35th Conference of the Spanish Society for Natural
  Language Processing, IberLEF@SEPLN 2019, September 24th, 2019}}
  \emph{(\bibinfo{series}{{CEUR} Workshop Proceedings},
  Vol.~\bibinfo{volume}{2421})}. \bibinfo{publisher}{CEUR-WS.org},
  \bibinfo{address}{Bilbao, Spain}, \bibinfo{pages}{478--494}.
\newblock
\urldef\tempurl%
\url{https://ceur-ws.org/Vol-2421/MEX-A3T\_overview.pdf}
\showURL{%
\tempurl}


\bibitem[Arag{\'{o}}n et~al\mbox{.}(2020)]%
        {Aragn2020OverviewOM}
\bibfield{author}{\bibinfo{person}{Mario~Ezra Arag{\'{o}}n},
  \bibinfo{person}{Horacio~Jes{\'{u}}s Jarqu{\'{\i}}n{-}V{\'{a}}squez},
  \bibinfo{person}{Manuel Montes{-}y{-}G{\'{o}}mez}, \bibinfo{person}{Hugo~Jair
  Escalante}, \bibinfo{person}{Luis~Villase{\~{n}}or Pineda},
  \bibinfo{person}{Helena G{\'{o}}mez{-}Adorno}, \bibinfo{person}{Juan~Pablo
  Posadas{-}Dur{\'{a}}n}, {and} \bibinfo{person}{Gemma Bel{-}Enguix}.}
  \bibinfo{year}{2020}\natexlab{}.
\newblock \showarticletitle{Overview of {MEX-A3T} at IberLEF 2020: Fake News
  and Aggressiveness Analysis in Mexican Spanish}. In
  \bibinfo{booktitle}{\emph{Proceedings of the Iberian Languages Evaluation
  Forum (IberLEF 2020) co-located with 36th Conference of the Spanish Society
  for Natural Language Processing {(SEPLN} 2020), M{\'{a}}laga, Spain,
  September 23th, 2020}} \emph{(\bibinfo{series}{{CEUR} Workshop Proceedings},
  Vol.~\bibinfo{volume}{2664})}. \bibinfo{publisher}{CEUR-WS.org},
  \bibinfo{address}{Spain}, \bibinfo{pages}{222--235}.
\newblock
\urldef\tempurl%
\url{https://ceur-ws.org/Vol-2664/mex-a3t\_overview.pdf}
\showURL{%
\tempurl}


\bibitem[Arellano et~al\mbox{.}(2022)]%
        {arellano2022overview}
\bibfield{author}{\bibinfo{person}{Luis~Joaqu{\'{\i}}n Arellano},
  \bibinfo{person}{Hugo~Jair Escalante}, \bibinfo{person}{Luis~Villase{\~{n}}or
  Pineda}, \bibinfo{person}{Manuel Montes{-}y{-}G{\'{o}}mez}, {and}
  \bibinfo{person}{Fernando S{\'{a}}nchez{-}Vega}.}
  \bibinfo{year}{2022}\natexlab{}.
\newblock \showarticletitle{Overview of {DA-VINCIS} at IberLEF 2022: Detection
  of Aggressive and Violent Incidents from Social Media in Spanish}.
\newblock \bibinfo{journal}{\emph{Proces. del Leng. Natural}}
  \bibinfo{volume}{69} (\bibinfo{year}{2022}), \bibinfo{pages}{207--215}.
\newblock
\urldef\tempurl%
\url{http://journal.sepln.org/sepln/ojs/ojs/index.php/pln/article/view/6441}
\showURL{%
\tempurl}


\bibitem[Aroyehun and Gelbukh(2018a)]%
        {aroyehun2018aggression}
\bibfield{author}{\bibinfo{person}{Segun~Taofeek Aroyehun} {and}
  \bibinfo{person}{Alexander Gelbukh}.} \bibinfo{year}{2018}\natexlab{a}.
\newblock \showarticletitle{Aggression Detection in Social Media: Using Deep
  Neural Networks, Data Augmentation, and Pseudo Labeling}. In
  \bibinfo{booktitle}{\emph{Proceedings of the First Workshop on Trolling,
  Aggression and Cyberbullying ({TRAC}-2018)}},
  \bibfield{editor}{\bibinfo{person}{Ritesh Kumar}, \bibinfo{person}{Atul~Kr.
  Ojha}, \bibinfo{person}{Marcos Zampieri}, {and} \bibinfo{person}{Shervin
  Malmasi}} (Eds.). \bibinfo{publisher}{Association for Computational
  Linguistics}, \bibinfo{address}{Santa Fe, New Mexico, USA},
  \bibinfo{pages}{90--97}.
\newblock
\urldef\tempurl%
\url{https://aclanthology.org/W18-4411}
\showURL{%
\tempurl}


\bibitem[Aroyehun and Gelbukh(2018b)]%
        {madisettyaggression}
\bibfield{author}{\bibinfo{person}{Segun~Taofeek Aroyehun} {and}
  \bibinfo{person}{Alexander~F. Gelbukh}.} \bibinfo{year}{2018}\natexlab{b}.
\newblock \showarticletitle{Aggression Detection in Social Media: Using Deep
  Neural Networks, Data Augmentation, and Pseudo Labeling}. In
  \bibinfo{booktitle}{\emph{Proceedings of the First Workshop on Trolling,
  Aggression and Cyberbullying, TRAC@COLING 2018, August 25, 2018}}.
  \bibinfo{publisher}{Association for Computational Linguistics},
  \bibinfo{address}{Santa Fe, New Mexico, USA}, \bibinfo{pages}{90--97}.
\newblock
\urldef\tempurl%
\url{https://aclanthology.org/W18-4411/}
\showURL{%
\tempurl}


\bibitem[Arroyo{-}Fern{\'{a}}ndez et~al\mbox{.}(2018)]%
        {arroyo2018cyberbullying}
\bibfield{author}{\bibinfo{person}{Ignacio Arroyo{-}Fern{\'{a}}ndez},
  \bibinfo{person}{Dominic Forest}, \bibinfo{person}{Juan{-}Manuel
  Torres{-}Moreno}, \bibinfo{person}{Mauricio Carrasco{-}Ruiz},
  \bibinfo{person}{Thomas Legeleux}, {and} \bibinfo{person}{Karen Joannette}.}
  \bibinfo{year}{2018}\natexlab{}.
\newblock \showarticletitle{Cyberbullying Detection Task: the {EBSI-LIA-UNAM}
  System {(ELU)} at COLING'18 {TRAC-1}}. In
  \bibinfo{booktitle}{\emph{Proceedings of the First Workshop on Trolling,
  Aggression and Cyberbullying, TRAC@COLING 2018, Santa Fe, New Mexico, USA,
  August 25, 2018}}, \bibfield{editor}{\bibinfo{person}{Ritesh Kumar},
  \bibinfo{person}{Atul~Kr. Ojha}, \bibinfo{person}{Marcos Zampieri}, {and}
  \bibinfo{person}{Shervin Malmasi}} (Eds.). \bibinfo{publisher}{Association
  for Computational Linguistics}, \bibinfo{pages}{140--149}.
\newblock
\urldef\tempurl%
\url{https://aclanthology.org/W18-4417/}
\showURL{%
\tempurl}


\bibitem[Balci and Salah(2015)]%
        {balci2015automatic}
\bibfield{author}{\bibinfo{person}{Koray Balci} {and}
  \bibinfo{person}{Albert~Ali Salah}.} \bibinfo{year}{2015}\natexlab{}.
\newblock \showarticletitle{Automatic analysis and identification of verbal
  aggression and abusive behaviors for online social games}.
\newblock \bibinfo{journal}{\emph{Computers in Human Behavior}}
  \bibinfo{volume}{53} (\bibinfo{year}{2015}), \bibinfo{pages}{517--526}.
\newblock


\bibitem[Bansal et~al\mbox{.}(2022)]%
        {bansal2022}
\bibfield{author}{\bibinfo{person}{Vibhuti Bansal}, \bibinfo{person}{Mrinal
  Tyagi}, \bibinfo{person}{Rajesh Sharma}, \bibinfo{person}{Vedika Gupta},
  {and} \bibinfo{person}{Qin Xin}.} \bibinfo{year}{2022}\natexlab{}.
\newblock \showarticletitle{A Transformer Based Approach for Abuse Detection in
  Code Mixed Indic Languages.}
\newblock \bibinfo{journal}{\emph{ACM Trans. Asian Low-Resour. Lang. Inf.
  Process.}} (\bibinfo{date}{nov} \bibinfo{year}{2022}).
\newblock
\showISSN{2375-4699}
\urldef\tempurl%
\url{https://doi.org/10.1145/3571818}
\showDOI{\tempurl}
\newblock
\shownote{Just Accepted}.


\bibitem[Barbieri et~al\mbox{.}(2016)]%
        {barbieri2016cosmopolitan}
\bibfield{author}{\bibinfo{person}{Francesco Barbieri}, \bibinfo{person}{German
  Kruszewski}, \bibinfo{person}{Francesco Ronzano}, {and}
  \bibinfo{person}{Horacio Saggion}.} \bibinfo{year}{2016}\natexlab{}.
\newblock \showarticletitle{How Cosmopolitan Are Emojis? Exploring Emojis Usage
  and Meaning over Different Languages with Distributional Semantics}. In
  \bibinfo{booktitle}{\emph{Proceedings of the 24th ACM International
  Conference on Multimedia}} (Amsterdam, The Netherlands)
  \emph{(\bibinfo{series}{MM '16})}. \bibinfo{publisher}{Association for
  Computing Machinery}, \bibinfo{address}{New York, NY, USA},
  \bibinfo{pages}{531–535}.
\newblock
\showISBNx{9781450336031}
\urldef\tempurl%
\url{https://doi.org/10.1145/2964284.2967278}
\showDOI{\tempurl}


\bibitem[Barbieri and Saggion(2014)]%
        {Barbieri2014}
\bibfield{author}{\bibinfo{person}{Francesco Barbieri} {and}
  \bibinfo{person}{Horacio Saggion}.} \bibinfo{year}{2014}\natexlab{}.
\newblock \showarticletitle{Modelling Irony in Twitter}.
\newblock \bibinfo{journal}{\emph{Proceedings of the Student Research Workshop
  at the 14th Conference of the European Chapter of the Association for
  Computational Linguistics}}, \bibinfo{pages}{56--64}.
\newblock
\urldef\tempurl%
\url{https://doi.org/10.3115/v1/E14-3007}
\showDOI{\tempurl}


\bibitem[Baron and Richardson(1994)]%
        {baron1994human}
\bibfield{author}{\bibinfo{person}{Robert~A Baron} {and}
  \bibinfo{person}{Deborah~R Richardson}.} \bibinfo{year}{1994}\natexlab{}.
\newblock \bibinfo{booktitle}{\emph{Human aggression}}.
\newblock \bibinfo{publisher}{Springer Science \& Business Media}.
\newblock


\bibitem[Baruah et~al\mbox{.}(2020)]%
        {Baruah2020AggressionII}
\bibfield{author}{\bibinfo{person}{Arup Baruah}, \bibinfo{person}{Kaushik Das},
  \bibinfo{person}{Ferdous Barbhuiya}, {and} \bibinfo{person}{Kuntal Dey}.}
  \bibinfo{year}{2020}\natexlab{}.
\newblock \showarticletitle{Aggression Identification in {E}nglish, {H}indi and
  {B}angla Text using {BERT}, {R}o{BERT}a and {SVM}}. In
  \bibinfo{booktitle}{\emph{Proceedings of the Second Workshop on Trolling,
  Aggression and Cyberbullying}}. \bibinfo{publisher}{European Language
  Resources Association (ELRA)}, \bibinfo{address}{Marseille, France},
  \bibinfo{pages}{76--82}.
\newblock
\showISBNx{979-10-95546-56-6}
\urldef\tempurl%
\url{https://aclanthology.org/2020.trac-1.12}
\showURL{%
\tempurl}


\bibitem[Basak et~al\mbox{.}(2019)]%
        {basak2019online}
\bibfield{author}{\bibinfo{person}{Rajesh Basak}, \bibinfo{person}{Shamik
  Sural}, \bibinfo{person}{Niloy Ganguly}, {and} \bibinfo{person}{Soumya~K
  Ghosh}.} \bibinfo{year}{2019}\natexlab{}.
\newblock \showarticletitle{Online public shaming on Twitter: Detection,
  analysis, and mitigation}.
\newblock \bibinfo{journal}{\emph{IEEE Transactions on computational social
  systems}} \bibinfo{volume}{6}, \bibinfo{number}{2} (\bibinfo{year}{2019}),
  \bibinfo{pages}{208--220}.
\newblock


\bibitem[Bazarova et~al\mbox{.}(2013)]%
        {bazarova2013managing}
\bibfield{author}{\bibinfo{person}{Natalya~N Bazarova},
  \bibinfo{person}{Jessie~G Taft}, \bibinfo{person}{Yoon~Hyung Choi}, {and}
  \bibinfo{person}{Dan Cosley}.} \bibinfo{year}{2013}\natexlab{}.
\newblock \showarticletitle{Managing impressions and relationships on Facebook:
  Self-presentational and relational concerns revealed through the analysis of
  language style}.
\newblock \bibinfo{journal}{\emph{Journal of Language and Social Psychology}}
  \bibinfo{volume}{32}, \bibinfo{number}{2} (\bibinfo{year}{2013}),
  \bibinfo{pages}{121--141}.
\newblock


\bibitem[Bender and Friedman(2018)]%
        {benderfriedman2018data}
\bibfield{author}{\bibinfo{person}{Emily~M. Bender} {and}
  \bibinfo{person}{Batya Friedman}.} \bibinfo{year}{2018}\natexlab{}.
\newblock \showarticletitle{Data Statements for Natural Language Processing:
  Toward Mitigating System Bias and Enabling Better Science}.
\newblock \bibinfo{journal}{\emph{Transactions of the Association for
  Computational Linguistics}}  \bibinfo{volume}{6} (\bibinfo{year}{2018}),
  \bibinfo{pages}{587--604}.
\newblock
\urldef\tempurl%
\url{https://doi.org/10.1162/tacl_a_00041}
\showDOI{\tempurl}


\bibitem[Bhattacharya et~al\mbox{.}(2020)]%
        {bhattacharya2020developing}
\bibfield{author}{\bibinfo{person}{Shiladitya Bhattacharya},
  \bibinfo{person}{Siddharth Singh}, \bibinfo{person}{Ritesh Kumar},
  \bibinfo{person}{Akanksha Bansal}, \bibinfo{person}{Akash Bhagat},
  \bibinfo{person}{Yogesh Dawer}, \bibinfo{person}{Bornini Lahiri}, {and}
  \bibinfo{person}{Atul~Kr Ojha}.} \bibinfo{year}{2020}\natexlab{}.
\newblock \showarticletitle{Developing a multilingual annotated corpus of
  misogyny and aggression}.
\newblock \bibinfo{journal}{\emph{arXiv preprint arXiv:2003.07428}}
  (\bibinfo{year}{2020}).
\newblock


\bibitem[Binns et~al\mbox{.}(2017)]%
        {Binns2017LikeTL}
\bibfield{author}{\bibinfo{person}{Reuben Binns}, \bibinfo{person}{Michael
  Veale}, \bibinfo{person}{Max~Van Kleek}, {and} \bibinfo{person}{Nigel
  Shadbolt}.} \bibinfo{year}{2017}\natexlab{}.
\newblock \showarticletitle{Like Trainer, Like Bot? Inheritance of Bias in
  Algorithmic Content Moderation}.
\newblock \bibinfo{journal}{\emph{ArXiv}}  \bibinfo{volume}{abs/1707.01477}
  (\bibinfo{year}{2017}).
\newblock
\urldef\tempurl%
\url{https://api.semanticscholar.org/CorpusID:2814848}
\showURL{%
\tempurl}


\bibitem[B{\l}achnio and Przepi{\'o}rka(2018)]%
        {blachnio2018facebook}
\bibfield{author}{\bibinfo{person}{Agata B{\l}achnio} {and}
  \bibinfo{person}{Aneta Przepi{\'o}rka}.} \bibinfo{year}{2018}\natexlab{}.
\newblock \showarticletitle{Facebook intrusion, fear of missing out,
  narcissism, and life satisfaction: A cross-sectional study}.
\newblock \bibinfo{journal}{\emph{Psychiatry research}}  \bibinfo{volume}{259}
  (\bibinfo{year}{2018}), \bibinfo{pages}{514--519}.
\newblock


\bibitem[Boadi and Kolog(2021)]%
        {boadi2021social}
\bibfield{author}{\bibinfo{person}{Caleb Boadi} {and}
  \bibinfo{person}{Emmanuel~Awuni Kolog}.} \bibinfo{year}{2021}\natexlab{}.
\newblock \showarticletitle{Social Media Aggression: An Assessment Based on the
  Contemporary Deterrence Theory.}. In \bibinfo{booktitle}{\emph{AMCIS}}.
\newblock


\bibitem[Bojanowski et~al\mbox{.}(2017)]%
        {bojanowski2017enriching}
\bibfield{author}{\bibinfo{person}{Piotr Bojanowski}, \bibinfo{person}{Edouard
  Grave}, \bibinfo{person}{Armand Joulin}, {and} \bibinfo{person}{Tomas
  Mikolov}.} \bibinfo{year}{2017}\natexlab{}.
\newblock \showarticletitle{Enriching word vectors with subword information}.
\newblock \bibinfo{journal}{\emph{Transactions of the association for
  computational linguistics}}  \bibinfo{volume}{5} (\bibinfo{year}{2017}),
  \bibinfo{pages}{135--146}.
\newblock


\bibitem[Borraccino et~al\mbox{.}(2022)]%
        {borraccino2022problematic}
\bibfield{author}{\bibinfo{person}{Alberto Borraccino}, \bibinfo{person}{Noemi
  Marengo}, \bibinfo{person}{Paola Dalmasso}, \bibinfo{person}{Claudia Marino},
  \bibinfo{person}{Silvia Ciardullo}, \bibinfo{person}{Paola Nardone},
  \bibinfo{person}{Patrizia Lemma}, {and} \bibinfo{person}{2018 HBSC-Italia
  Group}.} \bibinfo{year}{2022}\natexlab{}.
\newblock \showarticletitle{Problematic social media use and cyber aggression
  in Italian adolescents: the remarkable role of social support}.
\newblock \bibinfo{journal}{\emph{International journal of environmental
  research and public health}} \bibinfo{volume}{19}, \bibinfo{number}{15}
  (\bibinfo{year}{2022}), \bibinfo{pages}{9763}.
\newblock


\bibitem[Bouhnik and Deshen(2014)]%
        {bouhnik2014whatsapp}
\bibfield{author}{\bibinfo{person}{Dan Bouhnik} {and} \bibinfo{person}{Mor
  Deshen}.} \bibinfo{year}{2014}\natexlab{}.
\newblock \showarticletitle{WhatsApp goes to school: Mobile instant messaging
  between teachers and students}.
\newblock \bibinfo{journal}{\emph{Journal of Information Technology Education.
  Research}}  \bibinfo{volume}{13} (\bibinfo{year}{2014}),
  \bibinfo{pages}{217}.
\newblock


\bibitem[Burnap and Williams(2015)]%
        {burnap2015cyber}
\bibfield{author}{\bibinfo{person}{Pete Burnap} {and}
  \bibinfo{person}{Matthew~L Williams}.} \bibinfo{year}{2015}\natexlab{}.
\newblock \showarticletitle{Cyber hate speech on twitter: An application of
  machine classification and statistical modeling for policy and decision
  making}.
\newblock \bibinfo{journal}{\emph{Policy \& internet}} \bibinfo{volume}{7},
  \bibinfo{number}{2} (\bibinfo{year}{2015}), \bibinfo{pages}{223--242}.
\newblock


\bibitem[Buss(1962)]%
        {buss1962psychology}
\bibfield{author}{\bibinfo{person}{Arnold~H Buss}.}
  \bibinfo{year}{1962}\natexlab{}.
\newblock \bibinfo{title}{The psychology of aggression}.
\newblock


\bibitem[Cer et~al\mbox{.}(2018)]%
        {cer2018universal}
\bibfield{author}{\bibinfo{person}{Daniel Cer}, \bibinfo{person}{Yinfei Yang},
  \bibinfo{person}{Sheng-yi Kong}, \bibinfo{person}{Nan Hua},
  \bibinfo{person}{Nicole Limtiaco}, \bibinfo{person}{Rhomni~St John},
  \bibinfo{person}{Noah Constant}, \bibinfo{person}{Mario Guajardo-Cespedes},
  \bibinfo{person}{Steve Yuan}, \bibinfo{person}{Chris Tar}, {et~al\mbox{.}}}
  \bibinfo{year}{2018}\natexlab{}.
\newblock \showarticletitle{Universal sentence encoder for English}. In
  \bibinfo{booktitle}{\emph{Proceedings of the 2018 conference on empirical
  methods in natural language processing: system demonstrations}}.
  \bibinfo{pages}{169--174}.
\newblock


\bibitem[Chaturvedi and Sharma(2022)]%
        {chaturvedi2022}
\bibfield{author}{\bibinfo{person}{Animesh Chaturvedi} {and}
  \bibinfo{person}{Rajesh Sharma}.} \bibinfo{year}{2022}\natexlab{}.
\newblock \showarticletitle{minOffense: Inter-Agreement Hate Terms for Stable
  Rules, Concepts, Transitivities, and Lattices}. In
  \bibinfo{booktitle}{\emph{2022 IEEE 9th International Conference on Data
  Science and Advanced Analytics (DSAA)}}. \bibinfo{pages}{1--10}.
\newblock
\urldef\tempurl%
\url{https://doi.org/10.1109/DSAA54385.2022.10032389}
\showDOI{\tempurl}


\bibitem[Chatzakou et~al\mbox{.}(2017)]%
        {chatzakou2017mean}
\bibfield{author}{\bibinfo{person}{Despoina Chatzakou},
  \bibinfo{person}{Nicolas Kourtellis}, \bibinfo{person}{Jeremy Blackburn},
  \bibinfo{person}{Emiliano De~Cristofaro}, \bibinfo{person}{Gianluca
  Stringhini}, {and} \bibinfo{person}{Athena Vakali}.}
  \bibinfo{year}{2017}\natexlab{}.
\newblock \showarticletitle{Mean birds: Detecting aggression and bullying on
  twitter}. In \bibinfo{booktitle}{\emph{Proceedings of the 2017 ACM on web
  science conference}}. \bibinfo{pages}{13--22}.
\newblock


\bibitem[Chatzakou et~al\mbox{.}(2019)]%
        {chatzakou2019detecting}
\bibfield{author}{\bibinfo{person}{Despoina Chatzakou}, \bibinfo{person}{Ilias
  Leontiadis}, \bibinfo{person}{Jeremy Blackburn}, \bibinfo{person}{Emiliano~De
  Cristofaro}, \bibinfo{person}{Gianluca Stringhini}, \bibinfo{person}{Athena
  Vakali}, {and} \bibinfo{person}{Nicolas Kourtellis}.}
  \bibinfo{year}{2019}\natexlab{}.
\newblock \showarticletitle{Detecting cyberbullying and cyberaggression in
  social media}.
\newblock \bibinfo{journal}{\emph{ACM Transactions on the Web (TWEB)}}
  \bibinfo{volume}{13}, \bibinfo{number}{3} (\bibinfo{year}{2019}),
  \bibinfo{pages}{1--51}.
\newblock


\bibitem[Chen et~al\mbox{.}(2018)]%
        {Chen2018VerbalAD}
\bibfield{author}{\bibinfo{person}{Junyi Chen}, \bibinfo{person}{Shankai Yan},
  {and} \bibinfo{person}{Ka chun Wong}.} \bibinfo{year}{2018}\natexlab{}.
\newblock \showarticletitle{Verbal aggression detection on Twitter comments:
  convolutional neural network for short-text sentiment analysis}.
\newblock \bibinfo{journal}{\emph{Neural Computing and Applications}}
  (\bibinfo{year}{2018}), \bibinfo{pages}{1--10}.
\newblock


\bibitem[Chen et~al\mbox{.}(2017)]%
        {chen2017aggressivity}
\bibfield{author}{\bibinfo{person}{Junyi Chen}, \bibinfo{person}{Shankai Yan},
  {and} \bibinfo{person}{Ka-Chun Wong}.} \bibinfo{year}{2017}\natexlab{}.
\newblock \showarticletitle{Aggressivity Detection on Social Network Comments}.
  In \bibinfo{booktitle}{\emph{Proceedings of the 2017 International Conference
  on Intelligent Systems, Metaheuristics \& Swarm Intelligence}} (Hong Kong,
  Hong Kong) \emph{(\bibinfo{series}{ISMSI '17})}.
  \bibinfo{publisher}{Association for Computing Machinery},
  \bibinfo{address}{New York, NY, USA}, \bibinfo{pages}{103–107}.
\newblock
\showISBNx{9781450347983}
\urldef\tempurl%
\url{https://doi.org/10.1145/3059336.3059348}
\showDOI{\tempurl}


\bibitem[Corcoran et~al\mbox{.}(2015)]%
        {corcoran2015cyberbullying}
\bibfield{author}{\bibinfo{person}{Lucie Corcoran}, \bibinfo{person}{Conor
  Mc~Guckin}, {and} \bibinfo{person}{Garry Prentice}.}
  \bibinfo{year}{2015}\natexlab{}.
\newblock \showarticletitle{Cyberbullying or cyber aggression?: A review of
  existing definitions of cyber-based peer-to-peer aggression}.
\newblock \bibinfo{journal}{\emph{Societies}} \bibinfo{volume}{5},
  \bibinfo{number}{2} (\bibinfo{year}{2015}), \bibinfo{pages}{245--255}.
\newblock


\bibitem[Croce et~al\mbox{.}(2020)]%
        {croce2020gan}
\bibfield{author}{\bibinfo{person}{Danilo Croce}, \bibinfo{person}{Giuseppe
  Castellucci}, {and} \bibinfo{person}{Roberto Basili}.}
  \bibinfo{year}{2020}\natexlab{}.
\newblock \showarticletitle{GAN-BERT: Generative adversarial learning for
  robust text classification with a bunch of labeled examples}.
\newblock  (\bibinfo{year}{2020}).
\newblock


\bibitem[Datta et~al\mbox{.}(2020)]%
        {Datta2020SpyderAD}
\bibfield{author}{\bibinfo{person}{Anisha Datta}, \bibinfo{person}{Shukrity
  Si}, \bibinfo{person}{Urbi Chakraborty}, {and} \bibinfo{person}{Sudip~Kumar
  Naskar}.} \bibinfo{year}{2020}\natexlab{}.
\newblock \showarticletitle{Spyder: Aggression Detection on Multilingual
  Tweets}. In \bibinfo{booktitle}{\emph{Workshop on Trolling, Aggression and
  Cyberbullying}}.
\newblock


\bibitem[De~Souza and Da~Costa-Abreu(2020)]%
        {souza2020}
\bibfield{author}{\bibinfo{person}{Gabriel~Araújo De~Souza} {and}
  \bibinfo{person}{Márjory Da~Costa-Abreu}.} \bibinfo{year}{2020}\natexlab{}.
\newblock \showarticletitle{Automatic offensive language detection from Twitter
  data using machine learning and feature selection of metadata}. In
  \bibinfo{booktitle}{\emph{2020 International Joint Conference on Neural
  Networks (IJCNN)}}. \bibinfo{pages}{1--6}.
\newblock
\urldef\tempurl%
\url{https://doi.org/10.1109/IJCNN48605.2020.9207652}
\showDOI{\tempurl}


\bibitem[D{\'\i}az-Torres et~al\mbox{.}(2020)]%
        {diaz2020automatic}
\bibfield{author}{\bibinfo{person}{Mar{\'\i}a~Jos{\'e} D{\'\i}az-Torres},
  \bibinfo{person}{Paulina~Alejandra Mor{\'a}n-M{\'e}ndez},
  \bibinfo{person}{Luis Villasenor-Pineda}, \bibinfo{person}{Manuel Montes},
  \bibinfo{person}{Juan Aguilera}, {and} \bibinfo{person}{Luis
  Meneses-Ler{\'\i}n}.} \bibinfo{year}{2020}\natexlab{}.
\newblock \showarticletitle{Automatic detection of offensive language in social
  media: Defining linguistic criteria to build a Mexican Spanish dataset}. In
  \bibinfo{booktitle}{\emph{Proceedings of the Second Workshop on Trolling,
  Aggression and Cyberbullying}}. \bibinfo{pages}{132--136}.
\newblock


\bibitem[Dutta et~al\mbox{.}(2021)]%
        {dutta2021efficient}
\bibfield{author}{\bibinfo{person}{Sandip Dutta}, \bibinfo{person}{Utso
  Majumder}, {and} \bibinfo{person}{Sudip~Kumar Naskar}.}
  \bibinfo{year}{2021}\natexlab{}.
\newblock \showarticletitle{An efficient bert based approach to detect
  aggression and misogyny}. In \bibinfo{booktitle}{\emph{Proceedings of the
  18th International Conference on Natural Language Processing (ICON)}}.
  \bibinfo{pages}{493--498}.
\newblock


\bibitem[Elhafsi et~al\mbox{.}(2023)]%
        {elhafsi2023}
\bibfield{author}{\bibinfo{person}{Amine Elhafsi}, \bibinfo{person}{Rohan
  Sinha}, \bibinfo{person}{Christopher Agia}, \bibinfo{person}{Edward
  Schmerling}, \bibinfo{person}{Issa A.~D. Nesnas}, {and}
  \bibinfo{person}{Marco Pavone}.} \bibinfo{year}{2023}\natexlab{}.
\newblock \showarticletitle{Semantic anomaly detection with large language
  models}.
\newblock \bibinfo{journal}{\emph{Auton. Robots}} \bibinfo{volume}{47},
  \bibinfo{number}{8} (\bibinfo{date}{oct} \bibinfo{year}{2023}),
  \bibinfo{pages}{1035–1055}.
\newblock
\showISSN{0929-5593}
\urldef\tempurl%
\url{https://doi.org/10.1007/s10514-023-10132-6}
\showDOI{\tempurl}


\bibitem[Eraslan and Kukuoglu(2019)]%
        {eraslan2019social}
\bibfield{author}{\bibinfo{person}{Levent Eraslan} {and} \bibinfo{person}{Ahmet
  Kukuoglu}.} \bibinfo{year}{2019}\natexlab{}.
\newblock \showarticletitle{Social Relations in Virtual World and Social Media
  Aggression.}
\newblock \bibinfo{journal}{\emph{World Journal on Educational Technology:
  Current Issues}} \bibinfo{volume}{11}, \bibinfo{number}{2}
  (\bibinfo{year}{2019}), \bibinfo{pages}{1--11}.
\newblock


\bibitem[Facebook({[n.\,d.]})]%
        {facebook}
\bibfield{author}{\bibinfo{person}{Facebook}.}
  \bibinfo{year}{[n.\,d.]}\natexlab{}.
\newblock


\bibitem[Farías et~al\mbox{.}(2016)]%
        {Farias2016}
\bibfield{author}{\bibinfo{person}{Delia Irazú~Hernańdez Farías},
  \bibinfo{person}{Viviana Patti}, {and} \bibinfo{person}{Paolo Rosso}.}
  \bibinfo{year}{2016}\natexlab{}.
\newblock \showarticletitle{Irony Detection in Twitter}.
\newblock \bibinfo{journal}{\emph{ACM Transactions on Internet Technology}}
  \bibinfo{volume}{16} (\bibinfo{date}{8} \bibinfo{year}{2016}),
  \bibinfo{pages}{1--24}.
\newblock
Issue 3.
\showISSN{1533-5399}
\urldef\tempurl%
\url{https://doi.org/10.1145/2930663}
\showDOI{\tempurl}


\bibitem[Ferguson(2021)]%
        {ferguson2021does}
\bibfield{author}{\bibinfo{person}{Christopher~J Ferguson}.}
  \bibinfo{year}{2021}\natexlab{}.
\newblock \showarticletitle{Does the Internet Make the World Worse? Depression,
  Aggression and Polarization in the Social Media Age}.
\newblock \bibinfo{journal}{\emph{Bulletin of Science, Technology \& Society}}
  \bibinfo{volume}{41}, \bibinfo{number}{4} (\bibinfo{year}{2021}),
  \bibinfo{pages}{116--135}.
\newblock


\bibitem[Fortuna et~al\mbox{.}(2018)]%
        {fortuna2018merging}
\bibfield{author}{\bibinfo{person}{Paula Fortuna}, \bibinfo{person}{Jos{\'e}
  Ferreira}, \bibinfo{person}{Luiz Pires}, \bibinfo{person}{Guilherme Routar},
  {and} \bibinfo{person}{S{\'e}rgio Nunes}.} \bibinfo{year}{2018}\natexlab{}.
\newblock \showarticletitle{Merging Datasets for Aggressive Text
  Identification}. In \bibinfo{booktitle}{\emph{Proceedings of the First
  Workshop on Trolling, Aggression and Cyberbullying ({TRAC}-2018)}}.
  \bibinfo{publisher}{Association for Computational Linguistics},
  \bibinfo{address}{Santa Fe, New Mexico, USA}, \bibinfo{pages}{128--139}.
\newblock
\urldef\tempurl%
\url{https://aclanthology.org/W18-4416}
\showURL{%
\tempurl}


\bibitem[Fortuna and Nunes(2018)]%
        {fortuna2018survey}
\bibfield{author}{\bibinfo{person}{Paula Fortuna} {and}
  \bibinfo{person}{S{\'e}rgio Nunes}.} \bibinfo{year}{2018}\natexlab{}.
\newblock \showarticletitle{A survey on automatic detection of hate speech in
  text}.
\newblock \bibinfo{journal}{\emph{ACM Computing Surveys (CSUR)}}
  \bibinfo{volume}{51}, \bibinfo{number}{4} (\bibinfo{year}{2018}),
  \bibinfo{pages}{1--30}.
\newblock


\bibitem[Founta et~al\mbox{.}(2018)]%
        {founta2018large}
\bibfield{author}{\bibinfo{person}{Antigoni{-}Maria Founta},
  \bibinfo{person}{Constantinos Djouvas}, \bibinfo{person}{Despoina Chatzakou},
  \bibinfo{person}{Ilias Leontiadis}, \bibinfo{person}{Jeremy Blackburn},
  \bibinfo{person}{Gianluca Stringhini}, \bibinfo{person}{Athena Vakali},
  \bibinfo{person}{Michael Sirivianos}, {and} \bibinfo{person}{Nicolas
  Kourtellis}.} \bibinfo{year}{2018}\natexlab{}.
\newblock \showarticletitle{Large Scale Crowdsourcing and Characterization of
  Twitter Abusive Behavior}.
\newblock \bibinfo{journal}{\emph{CoRR}}  \bibinfo{volume}{abs/1802.00393}
  (\bibinfo{year}{2018}).
\newblock
\showeprint[arXiv]{1802.00393}
\urldef\tempurl%
\url{http://arxiv.org/abs/1802.00393}
\showURL{%
\tempurl}


\bibitem[Galery et~al\mbox{.}(2018)]%
        {galery2018aggression}
\bibfield{author}{\bibinfo{person}{Thiago Galery}, \bibinfo{person}{Efstathios
  Charitos}, {and} \bibinfo{person}{Ye Tian}.} \bibinfo{year}{2018}\natexlab{}.
\newblock \showarticletitle{Aggression Identification and Multi Lingual Word
  Embeddings}. In \bibinfo{booktitle}{\emph{Proceedings of the First Workshop
  on Trolling, Aggression and Cyberbullying ({TRAC}-2018)}}.
  \bibinfo{publisher}{Association for Computational Linguistics},
  \bibinfo{address}{Santa Fe, New Mexico, USA}, \bibinfo{pages}{74--79}.
\newblock
\urldef\tempurl%
\url{https://aclanthology.org/W18-4409}
\showURL{%
\tempurl}


\bibitem[Gallo et~al\mbox{.}(2020)]%
        {gallo2020predicting}
\bibfield{author}{\bibinfo{person}{Fabio~R Gallo}, \bibinfo{person}{Gerardo~I
  Simari}, \bibinfo{person}{Maria~Vanina Martinez}, {and}
  \bibinfo{person}{Marcelo~A Falappa}.} \bibinfo{year}{2020}\natexlab{}.
\newblock \showarticletitle{Predicting user reactions to Twitter feed content
  based on personality type and social cues}.
\newblock \bibinfo{journal}{\emph{Future Generation Computer Systems}}
  \bibinfo{volume}{110} (\bibinfo{year}{2020}), \bibinfo{pages}{918--930}.
\newblock


\bibitem[Galyashina et~al\mbox{.}({[n.\,d.]})]%
        {galyashinafake}
\bibfield{author}{\bibinfo{person}{Elena~I Galyashina},
  \bibinfo{person}{Vladimir~D Nikishin}, {et~al\mbox{.}}}
  \bibinfo{year}{[n.\,d.]}\natexlab{}.
\newblock \showarticletitle{Fake Media Products As Speech Aggression Provokers
  In Network Communication}.
\newblock \bibinfo{journal}{\emph{European Proceedings of Social and
  Behavioural Sciences}} (\bibinfo{year}{[n.\,d.]}).
\newblock


\bibitem[Gattulli et~al\mbox{.}(2022)]%
        {gattulli2022cyber}
\bibfield{author}{\bibinfo{person}{Vincenzo Gattulli}, \bibinfo{person}{Donato
  Impedovo}, \bibinfo{person}{Giuseppe Pirlo}, {and} \bibinfo{person}{Lucia
  Sarcinella}.} \bibinfo{year}{2022}\natexlab{}.
\newblock \showarticletitle{Cyber Aggression and Cyberbullying Identification
  on Social Networks}. In \bibinfo{booktitle}{\emph{International Conference on
  Pattern Recognition Applications and Methods}}.
\newblock
\urldef\tempurl%
\url{https://api.semanticscholar.org/CorpusID:246960574}
\showURL{%
\tempurl}


\bibitem[Ghosh et~al\mbox{.}(2023)]%
        {ghosh2023transformer}
\bibfield{author}{\bibinfo{person}{Soumitra Ghosh}, \bibinfo{person}{Amit
  Priyankar}, \bibinfo{person}{Asif Ekbal}, {and} \bibinfo{person}{Pushpak
  Bhattacharyya}.} \bibinfo{year}{2023}\natexlab{}.
\newblock \showarticletitle{A transformer-based multi-task framework for joint
  detection of aggression and hate on social media data}.
\newblock \bibinfo{journal}{\emph{Natural Language Engineering}}
  (\bibinfo{year}{2023}), \bibinfo{pages}{1--21}.
\newblock


\bibitem[Go et~al\mbox{.}(2009)]%
        {Sentiment140}
\bibfield{author}{\bibinfo{person}{Alec Go}, \bibinfo{person}{Richa Bhayani},
  {and} \bibinfo{person}{Lei Huang}.} \bibinfo{year}{2009}\natexlab{}.
\newblock \bibinfo{booktitle}{\emph{Twitter Sentiment Classification using
  Distant Supervision}}.
\newblock
\urldef\tempurl%
\url{http://help.sentiment140.com/home}
\showURL{%
\tempurl}


\bibitem[Golem et~al\mbox{.}(2018)]%
        {golem2018combining}
\bibfield{author}{\bibinfo{person}{Viktor Golem}, \bibinfo{person}{Mladen
  Karan}, {and} \bibinfo{person}{Jan {\v{S}}najder}.}
  \bibinfo{year}{2018}\natexlab{}.
\newblock \showarticletitle{Combining shallow and deep learning for aggressive
  text detection}. In \bibinfo{booktitle}{\emph{Proceedings of the First
  Workshop on Trolling, Aggression and Cyberbullying (TRAC-2018)}}.
  \bibinfo{pages}{188--198}.
\newblock


\bibitem[G{\'o}mez-Adorno et~al\mbox{.}(2018)]%
        {gomez2018machine}
\bibfield{author}{\bibinfo{person}{Helena G{\'o}mez-Adorno},
  \bibinfo{person}{Gemma~Bel Enguix}, \bibinfo{person}{Gerardo Sierra},
  \bibinfo{person}{Octavio S{\'a}nchez}, {and} \bibinfo{person}{Daniela
  Quezada}.} \bibinfo{year}{2018}\natexlab{}.
\newblock \showarticletitle{A Machine Learning Approach for Detecting
  Aggressive Tweets in Spanish.}. In \bibinfo{booktitle}{\emph{IberEval@
  SEPLN}}. \bibinfo{pages}{102--107}.
\newblock


\bibitem[Gordeev and Lykova(2020)]%
        {Gordeev2020BERTOA}
\bibfield{author}{\bibinfo{person}{Denis Gordeev} {and} \bibinfo{person}{Olga
  Lykova}.} \bibinfo{year}{2020}\natexlab{}.
\newblock \showarticletitle{{BERT} of all trades, master of some}. In
  \bibinfo{booktitle}{\emph{Proceedings of the Second Workshop on Trolling,
  Aggression and Cyberbullying}}. \bibinfo{publisher}{European Language
  Resources Association (ELRA)}, \bibinfo{address}{Marseille, France},
  \bibinfo{pages}{93--98}.
\newblock
\showISBNx{979-10-95546-56-6}
\urldef\tempurl%
\url{https://aclanthology.org/2020.trac-1.15}
\showURL{%
\tempurl}


\bibitem[Graff et~al\mbox{.}(2018)]%
        {Graff2018}
\bibfield{author}{\bibinfo{person}{Mario Graff}, \bibinfo{person}{Sabino
  Miranda-Jiménez}, \bibinfo{person}{Eric~S Tellez}, \bibinfo{person}{Daniela
  Moctezuma}, \bibinfo{person}{Vladimir Salgado}, \bibinfo{person}{José
  Ortiz-Bejar}, {and} \bibinfo{person}{Claudia~N Sánchez}.}
  \bibinfo{year}{2018}\natexlab{}.
\newblock \showarticletitle{INGEOTEC at MEX-A3T: Author profiling and
  aggressiveness analysis in Twitter using µTC and EvoMSA}.
\newblock \bibinfo{journal}{\emph{Proceedings of the Third Workshop on
  Evaluation of Human Language Technologies for Iberian Languages (IberEval
  2018)}}.
\newblock
\urldef\tempurl%
\url{https://mexa3t.wixsite.com/home}
\showURL{%
\tempurl}


\bibitem[Grigg(2010)]%
        {grigg2010cyber}
\bibfield{author}{\bibinfo{person}{Dorothy~Wunmi Grigg}.}
  \bibinfo{year}{2010}\natexlab{}.
\newblock \showarticletitle{Cyber-aggression: Definition and concept of
  cyberbullying}.
\newblock \bibinfo{journal}{\emph{Journal of Psychologists and Counsellors in
  Schools}} \bibinfo{volume}{20}, \bibinfo{number}{2} (\bibinfo{year}{2010}),
  \bibinfo{pages}{143--156}.
\newblock


\bibitem[Guti{\'e}rrez-Esparza et~al\mbox{.}(2019)]%
        {GutirrezEsparza2019ClassificationOC}
\bibfield{author}{\bibinfo{person}{Guadalupe~O. Guti{\'e}rrez-Esparza},
  \bibinfo{person}{Mait{\'e} Vallejo-Allende}, {and} \bibinfo{person}{Jos{\'e}
  Hern{\'a}ndez-Torruco}.} \bibinfo{year}{2019}\natexlab{}.
\newblock \showarticletitle{Classification of Cyber-Aggression Cases Applying
  Machine Learning}.
\newblock \bibinfo{journal}{\emph{Applied Sciences}} (\bibinfo{year}{2019}).
\newblock


\bibitem[Guzman{-}Silverio et~al\mbox{.}(2020)]%
        {GuzmanSilverio2020TransformersAD}
\bibfield{author}{\bibinfo{person}{Mario Guzman{-}Silverio},
  \bibinfo{person}{{\'{A}}ngel Balderas{-}Paredes}, {and}
  \bibinfo{person}{Adri{\'{a}}n~Pastor L{\'{o}}pez{-}Monroy}.}
  \bibinfo{year}{2020}\natexlab{}.
\newblock \showarticletitle{Transformers and Data Augmentation for
  Aggressiveness Detection in Mexican Spanish}. In
  \bibinfo{booktitle}{\emph{Proceedings of the Iberian Languages Evaluation
  Forum (IberLEF 2020) co-located with 36th Conference of the Spanish Society
  for Natural Language Processing {(SEPLN} 2020), M{\'{a}}laga, Spain,
  September 23th, 2020}} \emph{(\bibinfo{series}{{CEUR} Workshop Proceedings},
  Vol.~\bibinfo{volume}{2664})}. \bibinfo{publisher}{CEUR-WS.org},
  \bibinfo{pages}{293--302}.
\newblock
\urldef\tempurl%
\url{https://ceur-ws.org/Vol-2664/mexa3t\_paper9.pdf}
\showURL{%
\tempurl}


\bibitem[Henneberger et~al\mbox{.}(2017)]%
        {henneberger2017effect}
\bibfield{author}{\bibinfo{person}{Angela~K Henneberger},
  \bibinfo{person}{Donna~L Coffman}, {and} \bibinfo{person}{Scott~D Gest}.}
  \bibinfo{year}{2017}\natexlab{}.
\newblock \showarticletitle{The effect of having aggressive friends on
  aggressive behavior in childhood: Using propensity scores to strengthen
  causal inference}.
\newblock \bibinfo{journal}{\emph{Social development}} \bibinfo{volume}{26},
  \bibinfo{number}{2} (\bibinfo{year}{2017}), \bibinfo{pages}{295--309}.
\newblock


\bibitem[Herodotou et~al\mbox{.}(2021)]%
        {herodotou2021catching}
\bibfield{author}{\bibinfo{person}{Herodotos Herodotou},
  \bibinfo{person}{Despoina Chatzakou}, {and} \bibinfo{person}{Nicolas
  Kourtellis}.} \bibinfo{year}{2021}\natexlab{}.
\newblock \showarticletitle{Catching them red-handed: Real-time aggression
  detection on social media}. In \bibinfo{booktitle}{\emph{2021 IEEE 37th
  International Conference on Data Engineering (ICDE)}}. IEEE,
  \bibinfo{pages}{2123--2128}.
\newblock


\bibitem[Hinduja and Patchin(2018)]%
        {Hinduja2018}
\bibfield{author}{\bibinfo{person}{Sameer Hinduja} {and}
  \bibinfo{person}{Justin~W Patchin}.} \bibinfo{year}{2018}\natexlab{}.
\newblock \showarticletitle{Cyberbullying Identification, Prevention, and
  Response}.
\newblock \bibinfo{journal}{\emph{Cyberbullying Research Center}}
  (\bibinfo{year}{2018}).
\newblock


\bibitem[Huerta-Velasco and Calvo(2022)]%
        {huerta2022verbal}
\bibfield{author}{\bibinfo{person}{Daniel~Abraham Huerta-Velasco} {and}
  \bibinfo{person}{Hiram Calvo}.} \bibinfo{year}{2022}\natexlab{}.
\newblock \showarticletitle{Verbal Aggressions Detection in Mexican Tweets}.
\newblock \bibinfo{journal}{\emph{Computaci{\'o}n y Sistemas}}
  \bibinfo{volume}{26}, \bibinfo{number}{1} (\bibinfo{year}{2022}),
  \bibinfo{pages}{261--269}.
\newblock


\bibitem[Iqbal and Keshtkar(2019)]%
        {Iqbal2019UsingCL}
\bibfield{author}{\bibinfo{person}{Sayef Iqbal} {and} \bibinfo{person}{Fazel
  Keshtkar}.} \bibinfo{year}{2019}\natexlab{}.
\newblock \showarticletitle{Using Cognitive Learning Method to Analyze
  Aggression in Social Media Text}. In \bibinfo{booktitle}{\emph{Conference on
  Intelligent Text Processing and Computational Linguistics}}.
\newblock


\bibitem[Karan and Kundu(2023)]%
        {Karan2023}
\bibfield{author}{\bibinfo{person}{Suman Karan} {and} \bibinfo{person}{Suman
  Kundu}.} \bibinfo{year}{2023}\natexlab{}.
\newblock \showarticletitle{Cyberbully: Aggressive Tweets, Bully and Bully
  Target Profiling from Multilingual Indian Tweets}. In
  \bibinfo{booktitle}{\emph{Pattern Recognition and Machine Intelligence}},
  \bibfield{editor}{\bibinfo{person}{Pradipta Maji}, \bibinfo{person}{Tingwen
  Huang}, \bibinfo{person}{Nikhil~R. Pal}, \bibinfo{person}{Santanu Chaudhury},
  {and} \bibinfo{person}{Rajat~K. De}} (Eds.). \bibinfo{publisher}{Springer
  Nature Switzerland}, \bibinfo{address}{Cham}, \bibinfo{pages}{638--645}.
\newblock
\showISBNx{978-3-031-45170-6}


\bibitem[Kelly and Arnold(2016)]%
        {igiDef}
\bibfield{author}{\bibinfo{person}{Deirdre~M. Kelly} {and}
  \bibinfo{person}{Chrissie Arnold}.} \bibinfo{year}{2016}\natexlab{}.
\newblock \bibinfo{title}{World's Largest Database of Information Science and
  Technology}.
\newblock
\urldef\tempurl%
\url{https://www.igi-global.com/dictionary/cyber-aggression/6573}
\showURL{%
\tempurl}


\bibitem[Khan et~al\mbox{.}(2022)]%
        {khan2022aggression}
\bibfield{author}{\bibinfo{person}{Umair Khan}, \bibinfo{person}{Salabat Khan},
  \bibinfo{person}{Atif Rizwan}, \bibinfo{person}{Ghada Atteia},
  \bibinfo{person}{Mona~M Jamjoom}, {and} \bibinfo{person}{Nagwan~Abdel
  Samee}.} \bibinfo{year}{2022}\natexlab{}.
\newblock \showarticletitle{Aggression Detection in Social Media from Textual
  Data Using Deep Learning Models}.
\newblock \bibinfo{journal}{\emph{Applied Sciences}} \bibinfo{volume}{12},
  \bibinfo{number}{10} (\bibinfo{year}{2022}), \bibinfo{pages}{5083}.
\newblock


\bibitem[Khandelwal and Kumar(2020)]%
        {Khandelwal2020AUS}
\bibfield{author}{\bibinfo{person}{Anant Khandelwal} {and}
  \bibinfo{person}{Niraj Kumar}.} \bibinfo{year}{2020}\natexlab{}.
\newblock \showarticletitle{A Unified System for Aggression Identification in
  English Code-Mixed and Uni-Lingual Texts}.
\newblock \bibinfo{journal}{\emph{Proceedings of the 7th ACM IKDD CoDS and 25th
  COMAD}} (\bibinfo{year}{2020}).
\newblock


\bibitem[Kitchenham and Charters(2007)]%
        {Kitchenham2007}
\bibfield{author}{\bibinfo{person}{Barbara~Ann Kitchenham} {and}
  \bibinfo{person}{Stuart Charters}.} \bibinfo{year}{2007}\natexlab{}.
\newblock \bibinfo{booktitle}{\emph{Guidelines for performing Systematic
  Literature Reviews in Software Engineering}}.
\newblock \bibinfo{type}{{T}echnical {R}eport} EBSE 2007-001.
  \bibinfo{institution}{Keele University and Durham University Joint Report}.
\newblock
\urldef\tempurl%
\url{https://www.elsevier.com/__data/promis_misc/525444systematicreviewsguide.pdf}
\showURL{%
\tempurl}


\bibitem[Koufakou et~al\mbox{.}(2020)]%
        {koufakou2020florunito}
\bibfield{author}{\bibinfo{person}{Anna Koufakou}, \bibinfo{person}{Valerio
  Basile}, {and} \bibinfo{person}{Viviana Patti}.}
  \bibinfo{year}{2020}\natexlab{}.
\newblock \showarticletitle{{F}lor{U}ni{T}o@{TRAC}-2: Retrofitting Word
  Embeddings on an Abusive Lexicon for Aggressive Language Detection}. In
  \bibinfo{booktitle}{\emph{Proceedings of the Second Workshop on Trolling,
  Aggression and Cyberbullying}}. \bibinfo{publisher}{European Language
  Resources Association (ELRA)}, \bibinfo{address}{Marseille, France},
  \bibinfo{pages}{106--112}.
\newblock
\showISBNx{979-10-95546-56-6}
\urldef\tempurl%
\url{https://aclanthology.org/2020.trac-1.17}
\showURL{%
\tempurl}


\bibitem[Kowalski et~al\mbox{.}(2014)]%
        {kowalski2014bullying}
\bibfield{author}{\bibinfo{person}{Robin~M Kowalski}, \bibinfo{person}{Gary~W
  Giumetti}, \bibinfo{person}{Amber~N Schroeder}, {and}
  \bibinfo{person}{Micah~R Lattanner}.} \bibinfo{year}{2014}\natexlab{}.
\newblock \showarticletitle{Bullying in the digital age: a critical review and
  meta-analysis of cyberbullying research among youth.}
\newblock \bibinfo{journal}{\emph{Psychological bulletin}}
  \bibinfo{volume}{140}, \bibinfo{number}{4} (\bibinfo{year}{2014}),
  \bibinfo{pages}{1073}.
\newblock


\bibitem[Kowalski and Limber(2013)]%
        {kowalski2013psychological}
\bibfield{author}{\bibinfo{person}{Robin~M Kowalski} {and}
  \bibinfo{person}{Susan~P Limber}.} \bibinfo{year}{2013}\natexlab{}.
\newblock \showarticletitle{Psychological, physical, and academic correlates of
  cyberbullying and traditional bullying}.
\newblock \bibinfo{journal}{\emph{Journal of adolescent health}}
  \bibinfo{volume}{53}, \bibinfo{number}{1} (\bibinfo{year}{2013}),
  \bibinfo{pages}{S13--S20}.
\newblock


\bibitem[Kumar et~al\mbox{.}(2018a)]%
        {kumar2018trac}
\bibfield{author}{\bibinfo{person}{Ritesh Kumar}, \bibinfo{person}{Guggilla
  Bhanodai}, \bibinfo{person}{Rajendra Pamula}, {and}
  \bibinfo{person}{Maheshwar~Reddy Chennuru}.}
  \bibinfo{year}{2018}\natexlab{a}.
\newblock \showarticletitle{{TRAC}-1 Shared Task on Aggression Identification:
  {IIT}({ISM})@{COLING}{'}18}. In \bibinfo{booktitle}{\emph{Proceedings of the
  First Workshop on Trolling, Aggression and Cyberbullying ({TRAC}-2018)}}.
  \bibinfo{publisher}{Association for Computational Linguistics},
  \bibinfo{address}{Santa Fe, New Mexico, USA}, \bibinfo{pages}{58--65}.
\newblock
\urldef\tempurl%
\url{https://aclanthology.org/W18-4407}
\showURL{%
\tempurl}


\bibitem[Kumar et~al\mbox{.}(2020)]%
        {kumar2020evaluating}
\bibfield{author}{\bibinfo{person}{Ritesh Kumar}, \bibinfo{person}{Atul~Kr.
  Ojha}, \bibinfo{person}{Shervin Malmasi}, {and} \bibinfo{person}{Marcos
  Zampieri}.} \bibinfo{year}{2020}\natexlab{}.
\newblock \showarticletitle{Evaluating Aggression Identification in Social
  Media}. In \bibinfo{booktitle}{\emph{Proceedings of the Second Workshop on
  Trolling, Aggression and Cyberbullying}}. \bibinfo{publisher}{European
  Language Resources Association (ELRA)}, \bibinfo{address}{Marseille, France},
  \bibinfo{pages}{1--5}.
\newblock
\showISBNx{979-10-95546-56-6}
\urldef\tempurl%
\url{https://aclanthology.org/2020.trac-1.1}
\showURL{%
\tempurl}


\bibitem[Kumar et~al\mbox{.}(2022)]%
        {kumaretal2022comma}
\bibfield{author}{\bibinfo{person}{Ritesh Kumar}, \bibinfo{person}{Shyam
  Ratan}, \bibinfo{person}{Siddharth Singh}, \bibinfo{person}{Enakshi Nandi},
  \bibinfo{person}{Laishram~Niranjana Devi}, \bibinfo{person}{Akash Bhagat},
  \bibinfo{person}{Yogesh Dawer}, \bibinfo{person}{Bornini Lahiri},
  \bibinfo{person}{Akanksha Bansal}, {and} \bibinfo{person}{Atul~Kr. Ojha}.}
  \bibinfo{year}{2022}\natexlab{}.
\newblock \showarticletitle{The {C}om{MA} Dataset V0.2: Annotating Aggression
  and Bias in Multilingual Social Media Discourse}. In
  \bibinfo{booktitle}{\emph{Proceedings of the Thirteenth Language Resources
  and Evaluation Conference}}. \bibinfo{publisher}{European Language Resources
  Association}, \bibinfo{address}{Marseille, France},
  \bibinfo{pages}{4149--4161}.
\newblock
\urldef\tempurl%
\url{https://aclanthology.org/2022.lrec-1.441}
\showURL{%
\tempurl}


\bibitem[Kumar et~al\mbox{.}(2018b)]%
        {kumar2018aggression}
\bibfield{author}{\bibinfo{person}{Ritesh Kumar}, \bibinfo{person}{Aishwarya~N
  Reganti}, \bibinfo{person}{Akshit Bhatia}, {and} \bibinfo{person}{Tushar
  Maheshwari}.} \bibinfo{year}{2018}\natexlab{b}.
\newblock \showarticletitle{Aggression-annotated corpus of hindi-english
  code-mixed data}.
\newblock \bibinfo{journal}{\emph{arXiv preprint arXiv:1803.09402}}
  (\bibinfo{year}{2018}).
\newblock


\bibitem[Kumari and Singh(2020)]%
        {kumari2020ai}
\bibfield{author}{\bibinfo{person}{Kirti Kumari} {and}
  \bibinfo{person}{Jyoti~Prakash Singh}.} \bibinfo{year}{2020}\natexlab{}.
\newblock \showarticletitle{AI\_ML\_NIT\_Patna@ TRAC-2: Deep learning approach
  for multi-lingual aggression identification}. In
  \bibinfo{booktitle}{\emph{Proceedings of the second workshop on trolling,
  aggression and cyberbullying}}. \bibinfo{publisher}{European Language
  Resources Association (ELRA)}, \bibinfo{pages}{113--119}.
\newblock
\showISBNx{979-10-95546-56-6}
\urldef\tempurl%
\url{https://aclanthology.org/2020.trac-1.18}
\showURL{%
\tempurl}


\bibitem[Kumari and Singh(2021)]%
        {Kumari2021MultimodalCD}
\bibfield{author}{\bibinfo{person}{Kirti Kumari} {and}
  \bibinfo{person}{Jyoti~Prakash Singh}.} \bibinfo{year}{2021}\natexlab{}.
\newblock \showarticletitle{Multi-modal cyber-aggression detection with feature
  optimization by firefly algorithm}.
\newblock \bibinfo{journal}{\emph{Multimedia Systems}}  \bibinfo{volume}{28}
  (\bibinfo{year}{2021}), \bibinfo{pages}{1951 -- 1962}.
\newblock
\urldef\tempurl%
\url{https://api.semanticscholar.org/CorpusID:234811604}
\showURL{%
\tempurl}


\bibitem[Kumari and Singh(2022)]%
        {kumari2022multi}
\bibfield{author}{\bibinfo{person}{Kirti Kumari} {and}
  \bibinfo{person}{Jyoti~Prakash Singh}.} \bibinfo{year}{2022}\natexlab{}.
\newblock \showarticletitle{Multi-modal cyber-aggression detection with feature
  optimization by firefly algorithm}.
\newblock \bibinfo{journal}{\emph{Multimedia systems}} \bibinfo{volume}{28},
  \bibinfo{number}{6} (\bibinfo{year}{2022}), \bibinfo{pages}{1951--1962}.
\newblock


\bibitem[Kumari et~al\mbox{.}(2019)]%
        {kumari2019aggressive}
\bibfield{author}{\bibinfo{person}{Kirti Kumari},
  \bibinfo{person}{Jyoti~Prakash Singh}, \bibinfo{person}{Yogesh~K. Dwivedi},
  {and} \bibinfo{person}{Nripendra~P. Rana}.} \bibinfo{year}{2019}\natexlab{}.
\newblock \showarticletitle{Aggressive Social Media Post Detection System
  Containing Symbolic Images}. In \bibinfo{booktitle}{\emph{Digital
  Transformation for a Sustainable Society in the 21st Century}},
  \bibfield{editor}{\bibinfo{person}{Ilias~O. Pappas}, \bibinfo{person}{Patrick
  Mikalef}, \bibinfo{person}{Yogesh~K. Dwivedi}, \bibinfo{person}{Letizia
  Jaccheri}, \bibinfo{person}{John Krogstie}, {and} \bibinfo{person}{Matti
  M{\"a}ntym{\"a}ki}} (Eds.). \bibinfo{publisher}{Springer International
  Publishing}, \bibinfo{address}{Cham}, \bibinfo{pages}{415--424}.
\newblock
\showISBNx{978-3-030-29374-1}


\bibitem[Kumari et~al\mbox{.}(2021)]%
        {Kumari2021BilingualCD}
\bibfield{author}{\bibinfo{person}{Kirti Kumari},
  \bibinfo{person}{Jyoti~Prakash Singh}, \bibinfo{person}{Yogesh~K. Dwivedi},
  {and} \bibinfo{person}{Nripendra~P. Rana}.} \bibinfo{year}{2021}\natexlab{}.
\newblock \showarticletitle{Bilingual Cyber-aggression detection on social
  media using LSTM autoencoder}.
\newblock \bibinfo{journal}{\emph{Soft Computing}}  \bibinfo{volume}{25}
  (\bibinfo{year}{2021}), \bibinfo{pages}{8999 -- 9012}.
\newblock


\bibitem[Langham and Gosha(2018)]%
        {langham2018classification}
\bibfield{author}{\bibinfo{person}{Jaida Langham} {and} \bibinfo{person}{Kinnis
  Gosha}.} \bibinfo{year}{2018}\natexlab{}.
\newblock \showarticletitle{The classification of aggressive dialogue in social
  media platforms}. In \bibinfo{booktitle}{\emph{Proceedings of the 2018 ACM
  SIGMIS Conference on Computers and People Research}}.
  \bibinfo{pages}{60--63}.
\newblock


\bibitem[Le and Mikolov(2014)]%
        {le2014distributed}
\bibfield{author}{\bibinfo{person}{Quoc Le} {and} \bibinfo{person}{Tomas
  Mikolov}.} \bibinfo{year}{2014}\natexlab{}.
\newblock \showarticletitle{Distributed representations of sentences and
  documents}. In \bibinfo{booktitle}{\emph{International conference on machine
  learning}}. PMLR, \bibinfo{pages}{1188--1196}.
\newblock


\bibitem[Lee(2022)]%
        {lee2022germany}
\bibfield{author}{\bibinfo{person}{Diana Lee}.}
  \bibinfo{year}{2022}\natexlab{}.
\newblock \showarticletitle{Germany’s NetzDG and the threat to online free
  speech}.
\newblock \bibinfo{journal}{\emph{artigo eletr{\^o}nico]. In: Yale Law School:
  media, freedom \& infomation access clinic}}  \bibinfo{volume}{10}
  (\bibinfo{year}{2022}).
\newblock


\bibitem[Li et~al\mbox{.}(2023)]%
        {li2023hot}
\bibfield{author}{\bibinfo{person}{Lingyao Li}, \bibinfo{person}{Lizhou Fan},
  \bibinfo{person}{Shubham Atreja}, {and} \bibinfo{person}{Libby Hemphill}.}
  \bibinfo{year}{2023}\natexlab{}.
\newblock \showarticletitle{"HOT" ChatGPT: The promise of ChatGPT in detecting
  and discriminating hateful, offensive, and toxic comments on social media}.
\newblock \bibinfo{journal}{\emph{arXiv preprint arXiv:2304.10619}}
  (\bibinfo{year}{2023}).
\newblock


\bibitem[Liu et~al\mbox{.}(2020)]%
        {liu2020scmhl5}
\bibfield{author}{\bibinfo{person}{Han Liu}, \bibinfo{person}{Pete Burnap},
  \bibinfo{person}{Wafa Alorainy}, {and} \bibinfo{person}{Matthew~L.
  Williams}.} \bibinfo{year}{2020}\natexlab{}.
\newblock \showarticletitle{Scmhl5 at {TRAC-2} Shared Task on Aggression
  Identification: Bert Based Ensemble Learning Approach}. In
  \bibinfo{booktitle}{\emph{Proceedings of the Second Workshop on Trolling,
  Aggression and Cyberbullying, TRAC@LREC 2020Marseille, France, May 2020}}.
  \bibinfo{publisher}{European Language Resources Association {(ELRA)}},
  \bibinfo{address}{Marseille, France}, \bibinfo{pages}{62--68}.
\newblock
\urldef\tempurl%
\url{https://aclanthology.org/2020.trac-1.10/}
\showURL{%
\tempurl}


\bibitem[Livingstone et~al\mbox{.}(2011)]%
        {livingstone2011risks}
\bibfield{author}{\bibinfo{person}{Sonia Livingstone}, \bibinfo{person}{Leslie
  Haddon}, \bibinfo{person}{Anke G{\"o}rzig}, {and} \bibinfo{person}{Kjartan
  {\'O}lafsson}.} \bibinfo{year}{2011}\natexlab{}.
\newblock \showarticletitle{Risks and safety on the internet: the perspective
  of European children: full findings and policy implications from the EU Kids
  Online survey of 9-16 year olds and their parents in 25 countries}.
\newblock  (\bibinfo{year}{2011}).
\newblock


\bibitem[Maitra and Sarkhel(2018)]%
        {maitra2018k}
\bibfield{author}{\bibinfo{person}{Promita Maitra} {and}
  \bibinfo{person}{Ritesh Sarkhel}.} \bibinfo{year}{2018}\natexlab{}.
\newblock \showarticletitle{A K-Competitive Autoencoder for Aggression
  Detection in Social Media Text}. In \bibinfo{booktitle}{\emph{Proceedings of
  the First Workshop on Trolling, Aggression and Cyberbullying, TRAC@COLING
  2018, Santa Fe, New Mexico, USA, August 25, 2018}}.
  \bibinfo{publisher}{Association for Computational Linguistics},
  \bibinfo{pages}{80--89}.
\newblock
\urldef\tempurl%
\url{https://aclanthology.org/W18-4410/}
\showURL{%
\tempurl}


\bibitem[Mandl et~al\mbox{.}(2019)]%
        {mandl2019overview}
\bibfield{author}{\bibinfo{person}{Thomas Mandl}, \bibinfo{person}{Sandip
  Modha}, \bibinfo{person}{Prasenjit Majumder}, \bibinfo{person}{Daksh Patel},
  \bibinfo{person}{Mohana Dave}, \bibinfo{person}{Chintak Mandlia}, {and}
  \bibinfo{person}{Aditya Patel}.} \bibinfo{year}{2019}\natexlab{}.
\newblock \showarticletitle{Overview of the HASOC track at FIRE 2019: Hate
  Speech and Offensive Content Identification in Indo-European Languages}. In
  \bibinfo{booktitle}{\emph{Proceedings of the 11th Annual Meeting of the Forum
  for Information Retrieval Evaluation}} (Kolkata, India)
  \emph{(\bibinfo{series}{FIRE '19})}. \bibinfo{publisher}{Association for
  Computing Machinery}, \bibinfo{address}{New York, NY, USA},
  \bibinfo{pages}{14–17}.
\newblock
\showISBNx{9781450377508}
\urldef\tempurl%
\url{https://doi.org/10.1145/3368567.3368584}
\showDOI{\tempurl}


\bibitem[Mane et~al\mbox{.}(2023)]%
        {mane2023you}
\bibfield{author}{\bibinfo{person}{Swapnil Mane}, \bibinfo{person}{Suman
  Kundu}, {and} \bibinfo{person}{Rajesh Sharma}.}
  \bibinfo{year}{2023}\natexlab{}.
\newblock \showarticletitle{You are what your feeds makes you: A study of user
  aggressive behaviour on Twitter}.
\newblock  (\bibinfo{year}{2023}).
\newblock


\bibitem[Martins(2020)]%
        {martins2020effects}
\bibfield{author}{\bibinfo{person}{Nicole Martins}.}
  \bibinfo{year}{2020}\natexlab{}.
\newblock \showarticletitle{Effects of media use on social aggression in
  childhood and adolescence}.
\newblock \bibinfo{journal}{\emph{The international encyclopedia of media
  psychology}} (\bibinfo{year}{2020}), \bibinfo{pages}{1--5}.
\newblock


\bibitem[Mathew et~al\mbox{.}(2019)]%
        {mathew2019spread}
\bibfield{author}{\bibinfo{person}{Binny Mathew}, \bibinfo{person}{Ritam Dutt},
  \bibinfo{person}{Pawan Goyal}, {and} \bibinfo{person}{Animesh Mukherjee}.}
  \bibinfo{year}{2019}\natexlab{}.
\newblock \showarticletitle{Spread of Hate Speech in Online Social Media}. In
  \bibinfo{booktitle}{\emph{Proceedings of the 11th {ACM} Conference on Web
  Science, WebSci 2019, Boston, MA, USA, June 30 - July 03, 2019}}.
  \bibinfo{publisher}{{ACM}}, \bibinfo{pages}{173--182}.
\newblock
\urldef\tempurl%
\url{https://doi.org/10.1145/3292522.3326034}
\showDOI{\tempurl}


\bibitem[Mikolov et~al\mbox{.}(2013)]%
        {mikolov2013efficient}
\bibfield{author}{\bibinfo{person}{Tomas Mikolov}, \bibinfo{person}{Kai Chen},
  \bibinfo{person}{Greg Corrado}, {and} \bibinfo{person}{Jeffrey Dean}.}
  \bibinfo{year}{2013}\natexlab{}.
\newblock \showarticletitle{Efficient estimation of word representations in
  vector space}.
\newblock \bibinfo{journal}{\emph{arXiv preprint arXiv:1301.3781}}
  (\bibinfo{year}{2013}).
\newblock


\bibitem[Mishna et~al\mbox{.}(2023)]%
        {UT_ref}
\bibfield{author}{\bibinfo{person}{Faye Mishna}, \bibinfo{person}{Joanne
  Daciuk}, \bibinfo{person}{Ashley Lacombe-Duncan}, \bibinfo{person}{Gwendolyn
  Fearing}, {and} \bibinfo{person}{Melissa Van~Wert}.}
  \bibinfo{year}{2023}\natexlab{}.
\newblock \bibinfo{title}{University of Toronto (SLC2552)}.
\newblock
\urldef\tempurl%
\url{https://www.viceprovoststudents.utoronto.ca/wp-content/uploads/SLC2552_Cyber-Bullying_AODA.pdf}
\showURL{%
\tempurl}


\bibitem[Mishna et~al\mbox{.}(2018)]%
        {mishna2018social}
\bibfield{author}{\bibinfo{person}{Faye Mishna}, \bibinfo{person}{Cheryl
  Regehr}, \bibinfo{person}{Ashley Lacombe-Duncan}, \bibinfo{person}{Joanne
  Daciuk}, \bibinfo{person}{Gwendolyn Fearing}, {and} \bibinfo{person}{Melissa
  Van~Wert}.} \bibinfo{year}{2018}\natexlab{}.
\newblock \showarticletitle{Social media, cyber-aggression and student mental
  health on a university campus}.
\newblock \bibinfo{journal}{\emph{Journal of mental health}}
  \bibinfo{volume}{27}, \bibinfo{number}{3} (\bibinfo{year}{2018}),
  \bibinfo{pages}{222--229}.
\newblock


\bibitem[Mishra et~al\mbox{.}(2020)]%
        {Mishra2020MultilingualJF}
\bibfield{author}{\bibinfo{person}{Sudhanshu Mishra}, \bibinfo{person}{Shivangi
  Prasad}, {and} \bibinfo{person}{Shubhanshu Mishra}.}
  \bibinfo{year}{2020}\natexlab{}.
\newblock \showarticletitle{Multilingual Joint Fine-tuning of Transformer
  models for identifying Trolling, Aggression and Cyberbullying at {TRAC}
  2020}. In \bibinfo{booktitle}{\emph{Proceedings of the Second Workshop on
  Trolling, Aggression and Cyberbullying, TRAC@LREC 2020, Marseille, France,
  May 2020}}. \bibinfo{publisher}{European Language Resources Association
  {(ELRA)}}, \bibinfo{address}{Marseille, France}, \bibinfo{pages}{120--125}.
\newblock
\urldef\tempurl%
\url{https://aclanthology.org/2020.trac-1.19/}
\showURL{%
\tempurl}


\bibitem[Mladenovi{\'c} et~al\mbox{.}(2021)]%
        {mladenovic2021cyber}
\bibfield{author}{\bibinfo{person}{Miljana Mladenovi{\'c}},
  \bibinfo{person}{Vera O{\v{s}}mjanski}, {and}
  \bibinfo{person}{Sta{\v{s}}a~Vuji{\v{c}}i{\'c} Stankovi{\'c}}.}
  \bibinfo{year}{2021}\natexlab{}.
\newblock \showarticletitle{Cyber-aggression, cyberbullying, and
  cyber-grooming: a survey and research challenges}.
\newblock \bibinfo{journal}{\emph{ACM Computing Surveys (CSUR)}}
  \bibinfo{volume}{54}, \bibinfo{number}{1} (\bibinfo{year}{2021}),
  \bibinfo{pages}{1--42}.
\newblock


\bibitem[Modha et~al\mbox{.}(2018)]%
        {modha2018filtering}
\bibfield{author}{\bibinfo{person}{Sandip Modha}, \bibinfo{person}{Prasenjit
  Majumder}, {and} \bibinfo{person}{Thomas Mandl}.}
  \bibinfo{year}{2018}\natexlab{}.
\newblock \showarticletitle{Filtering aggression from the multilingual social
  media feed}. In \bibinfo{booktitle}{\emph{Proceedings of the first workshop
  on trolling, aggression and cyberbullying (TRAC-2018)}}.
  \bibinfo{pages}{199--207}.
\newblock


\bibitem[Modha et~al\mbox{.}(2022)]%
        {modha2022empirical}
\bibfield{author}{\bibinfo{person}{Sandip Modha}, \bibinfo{person}{Prasenjit
  Majumder}, {and} \bibinfo{person}{Thomas Mandl}.}
  \bibinfo{year}{2022}\natexlab{}.
\newblock \showarticletitle{An empirical evaluation of text representation
  schemes to filter the social media stream}.
\newblock \bibinfo{journal}{\emph{Journal of Experimental \& Theoretical
  Artificial Intelligence}} \bibinfo{volume}{34}, \bibinfo{number}{3}
  (\bibinfo{year}{2022}), \bibinfo{pages}{499--525}.
\newblock


\bibitem[Moher et~al\mbox{.}(2009)]%
        {Moher2009}
\bibfield{author}{\bibinfo{person}{David Moher}, \bibinfo{person}{Alessandro
  Liberati}, \bibinfo{person}{Jennifer Tetzlaff}, \bibinfo{person}{Douglas~G
  Altman}, {and} \bibinfo{person}{t PRISMA~Group*}.}
  \bibinfo{year}{2009}\natexlab{}.
\newblock \showarticletitle{Preferred reporting items for systematic reviews
  and meta-analyses: the PRISMA statement}.
\newblock \bibinfo{journal}{\emph{Annals of internal medicine}}
  \bibinfo{volume}{151}, \bibinfo{number}{4} (\bibinfo{year}{2009}),
  \bibinfo{pages}{264--269}.
\newblock


\bibitem[Mundra and Mittal(2021)]%
        {Mundra2021EvaluationOT}
\bibfield{author}{\bibinfo{person}{Shikha Mundra} {and} \bibinfo{person}{Namita
  Mittal}.} \bibinfo{year}{2021}\natexlab{}.
\newblock \showarticletitle{Evaluation of Text Representation Method to Detect
  Cyber Aggression in Hindi English Code Mixed Social Media Text}.
\newblock \bibinfo{journal}{\emph{2021 Thirteenth International Conference on
  Contemporary Computing (IC3-2021)}} (\bibinfo{year}{2021}).
\newblock


\bibitem[Mut~Alt{\i}n et~al\mbox{.}(2020)]%
        {altin2020lastus}
\bibfield{author}{\bibinfo{person}{L{\"u}tfiye~Seda Mut~Alt{\i}n},
  \bibinfo{person}{Alex Bravo}, {and} \bibinfo{person}{Horacio Saggion}.}
  \bibinfo{year}{2020}\natexlab{}.
\newblock \showarticletitle{{L}a{STUS}/{TALN} at {TRAC} - 2020 Trolling,
  Aggression and Cyberbullying}. In \bibinfo{booktitle}{\emph{Proceedings of
  the Second Workshop on Trolling, Aggression and Cyberbullying}}.
  \bibinfo{publisher}{European Language Resources Association (ELRA)},
  \bibinfo{address}{Marseille, France}, \bibinfo{pages}{83--86}.
\newblock
\showISBNx{979-10-95546-56-6}
\urldef\tempurl%
\url{https://aclanthology.org/2020.trac-1.13}
\showURL{%
\tempurl}


\bibitem[Nikhil et~al\mbox{.}(2018)]%
        {Nikhil2018LSTMsWA}
\bibfield{author}{\bibinfo{person}{Nishant Nikhil}, \bibinfo{person}{Ramit
  Pahwa}, \bibinfo{person}{Mehul~Kumar Nirala}, {and} \bibinfo{person}{Rohan
  Khilnani}.} \bibinfo{year}{2018}\natexlab{}.
\newblock \showarticletitle{LSTMs with Attention for Aggression Detection}. In
  \bibinfo{booktitle}{\emph{TRAC@COLING 2018}}.
\newblock


\bibitem[Nobata et~al\mbox{.}(2016)]%
        {nobata2016abusive}
\bibfield{author}{\bibinfo{person}{Chikashi Nobata}, \bibinfo{person}{Joel
  Tetreault}, \bibinfo{person}{Achint Thomas}, \bibinfo{person}{Yashar Mehdad},
  {and} \bibinfo{person}{Yi Chang}.} \bibinfo{year}{2016}\natexlab{}.
\newblock \showarticletitle{Abusive language detection in online user content}.
  In \bibinfo{booktitle}{\emph{Proceedings of the 25th international conference
  on world wide web}}. \bibinfo{pages}{145--153}.
\newblock


\bibitem[Orabi et~al\mbox{.}(2018)]%
        {orabi2018cyber}
\bibfield{author}{\bibinfo{person}{Ahmed~Husseini Orabi},
  \bibinfo{person}{Mahmoud~Husseini Orabi}, \bibinfo{person}{Qianjia Huang},
  \bibinfo{person}{Diana Inkpen}, {and} \bibinfo{person}{David Van~Bruwaene}.}
  \bibinfo{year}{2018}\natexlab{}.
\newblock \showarticletitle{Cyber-aggression detection using cross
  segment-and-concatenate multi-task learning from text}. In
  \bibinfo{booktitle}{\emph{Proceedings of the first workshop on trolling,
  aggression and cyberbullying (TRAC-2018)}}. \bibinfo{pages}{159--165}.
\newblock


\bibitem[Ora{\v{s}}an(2018)]%
        {oravsan2018aggressive}
\bibfield{author}{\bibinfo{person}{Constantin Ora{\v{s}}an}.}
  \bibinfo{year}{2018}\natexlab{}.
\newblock \showarticletitle{Aggressive language identification using word
  embeddings and sentiment features}. In \bibinfo{booktitle}{\emph{Proceedings
  of the first workshop on trolling, aggression and cyberbullying
  (TRAC-2018)}}. \bibinfo{pages}{113--119}.
\newblock


\bibitem[Ortega-Mendoza and López-Monroy(2018)]%
        {Ortega}
\bibfield{author}{\bibinfo{person}{Rosa~María Ortega-Mendoza} {and}
  \bibinfo{person}{A~Pastor López-Monroy}.} \bibinfo{year}{2018}\natexlab{}.
\newblock \showarticletitle{The Winning Approach for Author Profiling of
  Mexican Users in Twitter at MEX.A3T@IBEREVAL-2018}.
\newblock \bibinfo{journal}{\emph{Proceedings of the Third Workshop on
  Evaluation of Human Language Technologies for Iberian Languages (IberEval
  2018)}}, \bibinfo{pages}{140--148}.
\newblock


\bibitem[Ousidhoum et~al\mbox{.}(2020)]%
        {ousidhoum-etal-2020-comparative}
\bibfield{author}{\bibinfo{person}{Nedjma Ousidhoum}, \bibinfo{person}{Yangqiu
  Song}, {and} \bibinfo{person}{Dit-Yan Yeung}.}
  \bibinfo{year}{2020}\natexlab{}.
\newblock \showarticletitle{Comparative Evaluation of Label-Agnostic Selection
  Bias in Multilingual Hate Speech Datasets}. In
  \bibinfo{booktitle}{\emph{Proceedings of the 2020 Conference on Empirical
  Methods in Natural Language Processing (EMNLP)}},
  \bibfield{editor}{\bibinfo{person}{Bonnie Webber}, \bibinfo{person}{Trevor
  Cohn}, \bibinfo{person}{Yulan He}, {and} \bibinfo{person}{Yang Liu}} (Eds.).
  \bibinfo{publisher}{Association for Computational Linguistics},
  \bibinfo{address}{Online}, \bibinfo{pages}{2532--2542}.
\newblock
\urldef\tempurl%
\url{https://doi.org/10.18653/v1/2020.emnlp-main.199}
\showDOI{\tempurl}


\bibitem[Pantic et~al\mbox{.}(2012)]%
        {pantic2012association}
\bibfield{author}{\bibinfo{person}{Igor Pantic}, \bibinfo{person}{Aleksandar
  Damjanovic}, \bibinfo{person}{Jovana Todorovic}, \bibinfo{person}{Dubravka
  Topalovic}, \bibinfo{person}{Dragana Bojovic-Jovic}, \bibinfo{person}{Sinisa
  Ristic}, {and} \bibinfo{person}{Senka Pantic}.}
  \bibinfo{year}{2012}\natexlab{}.
\newblock \showarticletitle{Association between online social networking and
  depression in high school students: behavioral physiology viewpoint}.
\newblock \bibinfo{journal}{\emph{Psychiatria Danubina}} \bibinfo{volume}{24},
  \bibinfo{number}{1.} (\bibinfo{year}{2012}), \bibinfo{pages}{90--93}.
\newblock


\bibitem[Pascual-Ferr{\'a} et~al\mbox{.}(2021)]%
        {pascual2021toxicity}
\bibfield{author}{\bibinfo{person}{Paola Pascual-Ferr{\'a}},
  \bibinfo{person}{Neil Alperstein}, \bibinfo{person}{Daniel~J Barnett}, {and}
  \bibinfo{person}{Rajiv~N Rimal}.} \bibinfo{year}{2021}\natexlab{}.
\newblock \showarticletitle{Toxicity and verbal aggression on social media:
  Polarized discourse on wearing face masks during the COVID-19 pandemic}.
\newblock \bibinfo{journal}{\emph{Big Data \& Society}} \bibinfo{volume}{8},
  \bibinfo{number}{1} (\bibinfo{year}{2021}),
  \bibinfo{pages}{20539517211023533}.
\newblock


\bibitem[Pascucci et~al\mbox{.}(2020)]%
        {pascucci2020role}
\bibfield{author}{\bibinfo{person}{Antonio Pascucci}, \bibinfo{person}{Raffaele
  Manna}, \bibinfo{person}{Vincenzo Masucci}, {and} \bibinfo{person}{Johanna
  Monti}.} \bibinfo{year}{2020}\natexlab{}.
\newblock \showarticletitle{The role of computational stylometry in identifying
  (misogynistic) aggression in english social media texts}. In
  \bibinfo{booktitle}{\emph{Proceedings of the Second Workshop on Trolling,
  Aggression and Cyberbullying}}. \bibinfo{pages}{69--75}.
\newblock


\bibitem[Patchin and Hinduja(2015)]%
        {patchin2015measuring}
\bibfield{author}{\bibinfo{person}{Justin~W Patchin} {and}
  \bibinfo{person}{Sameer Hinduja}.} \bibinfo{year}{2015}\natexlab{}.
\newblock \showarticletitle{Measuring cyberbullying: Implications for
  research}.
\newblock \bibinfo{journal}{\emph{Aggression and Violent Behavior}}
  \bibinfo{volume}{23} (\bibinfo{year}{2015}), \bibinfo{pages}{69--74}.
\newblock


\bibitem[Patwa et~al\mbox{.}(2021)]%
        {patwa2021hater}
\bibfield{author}{\bibinfo{person}{Parth Patwa}, \bibinfo{person}{Srinivas
  Pykl}, \bibinfo{person}{Amitava Das}, \bibinfo{person}{Prerana Mukherjee},
  {and} \bibinfo{person}{Viswanath Pulabaigari}.}
  \bibinfo{year}{2021}\natexlab{}.
\newblock \showarticletitle{Hater-o-genius aggression classification using
  capsule networks}.
\newblock \bibinfo{journal}{\emph{arXiv preprint arXiv:2105.11219}}
  (\bibinfo{year}{2021}).
\newblock


\bibitem[Peng et~al\mbox{.}(2003)]%
        {peng2003language}
\bibfield{author}{\bibinfo{person}{Fuchun Peng}, \bibinfo{person}{Dale
  Schuurmans}, \bibinfo{person}{Vlado Keselj}, {and} \bibinfo{person}{Shaojun
  Wang}.} \bibinfo{year}{2003}\natexlab{}.
\newblock \showarticletitle{Language independent authorship attribution with
  character level n-grams}. In \bibinfo{booktitle}{\emph{10th Conference of the
  European Chapter of the Association for Computational Linguistics}}.
\newblock


\bibitem[Pennington et~al\mbox{.}(2014)]%
        {pennington2014glove}
\bibfield{author}{\bibinfo{person}{Jeffrey Pennington},
  \bibinfo{person}{Richard Socher}, {and} \bibinfo{person}{Christopher~D
  Manning}.} \bibinfo{year}{2014}\natexlab{}.
\newblock \showarticletitle{Glove: Global vectors for word representation}. In
  \bibinfo{booktitle}{\emph{Proceedings of the 2014 conference on empirical
  methods in natural language processing (EMNLP)}}.
  \bibinfo{pages}{1532--1543}.
\newblock


\bibitem[Pires et~al\mbox{.}(2019)]%
        {pires-etal-2019-multilingual}
\bibfield{author}{\bibinfo{person}{Telmo Pires}, \bibinfo{person}{Eva
  Schlinger}, {and} \bibinfo{person}{Dan Garrette}.}
  \bibinfo{year}{2019}\natexlab{}.
\newblock \showarticletitle{How Multilingual is Multilingual {BERT}?}. In
  \bibinfo{booktitle}{\emph{Proceedings of the 57th Annual Meeting of the
  Association for Computational Linguistics}},
  \bibfield{editor}{\bibinfo{person}{Anna Korhonen}, \bibinfo{person}{David
  Traum}, {and} \bibinfo{person}{Llu{\'\i}s M{\`a}rquez}} (Eds.).
  \bibinfo{publisher}{Association for Computational Linguistics},
  \bibinfo{address}{Florence, Italy}, \bibinfo{pages}{4996--5001}.
\newblock
\urldef\tempurl%
\url{https://doi.org/10.18653/v1/P19-1493}
\showDOI{\tempurl}


\bibitem[Poiitis et~al\mbox{.}(2021)]%
        {poiitis2021aggression}
\bibfield{author}{\bibinfo{person}{Marinos Poiitis}, \bibinfo{person}{Athena
  Vakali}, {and} \bibinfo{person}{Nicolas Kourtellis}.}
  \bibinfo{year}{2021}\natexlab{}.
\newblock \showarticletitle{On the aggression diffusion modeling and
  minimization in Twitter}.
\newblock \bibinfo{journal}{\emph{ACM Transactions on the Web (TWEB)}}
  \bibinfo{volume}{16}, \bibinfo{number}{1} (\bibinfo{year}{2021}),
  \bibinfo{pages}{1--24}.
\newblock


\bibitem[Quang-Loc(2021)]%
        {quang2021some}
\bibfield{author}{\bibinfo{person}{Nguyen Quang-Loc}.}
  \bibinfo{year}{2021}\natexlab{}.
\newblock \showarticletitle{Some thoughts on Vietnamese aggression and violence
  due to social media effects}.
\newblock  (\bibinfo{year}{2021}).
\newblock


\bibitem[Raiyani et~al\mbox{.}(2018)]%
        {Raiyani2018FullyCN}
\bibfield{author}{\bibinfo{person}{Kashyap Raiyani}, \bibinfo{person}{Teresa
  Gonçalves}, \bibinfo{person}{Paulo Quaresma}, {and}
  \bibinfo{person}{V{\'i}tor~Beires Nogueira}.}
  \bibinfo{year}{2018}\natexlab{}.
\newblock \showarticletitle{Fully Connected Neural Network with Advance
  Preprocessor to Identify Aggression over Facebook and Twitter}. In
  \bibinfo{booktitle}{\emph{TRAC@COLING 2018}}.
\newblock


\bibitem[Raman et~al\mbox{.}(2022)]%
        {raman2022hate}
\bibfield{author}{\bibinfo{person}{Shatakshi Raman}, \bibinfo{person}{Vedika
  Gupta}, \bibinfo{person}{Preeti Nagrath}, {and} \bibinfo{person}{KC
  Santosh}.} \bibinfo{year}{2022}\natexlab{}.
\newblock \showarticletitle{Hate and aggression analysis in NLP with
  explainable AI}.
\newblock \bibinfo{journal}{\emph{International Journal of Pattern Recognition
  and Artificial Intelligence}} \bibinfo{volume}{36}, \bibinfo{number}{15}
  (\bibinfo{year}{2022}), \bibinfo{pages}{2259036}.
\newblock


\bibitem[Ramiandrisoa(2022)]%
        {Ramiandrisoa2022MultitaskLF}
\bibfield{author}{\bibinfo{person}{Faneva Ramiandrisoa}.}
  \bibinfo{year}{2022}\natexlab{}.
\newblock \showarticletitle{Multi-task Learning for Hate Speech and Aggression
  Detection}. In \bibinfo{booktitle}{\emph{Joint Conference of the Information
  Retrieval Communities in Europe}}.
\newblock


\bibitem[Ramiandrisoa and Mothe(2018)]%
        {ramiandrisoa2018irit}
\bibfield{author}{\bibinfo{person}{Faneva Ramiandrisoa} {and}
  \bibinfo{person}{Josiane Mothe}.} \bibinfo{year}{2018}\natexlab{}.
\newblock \showarticletitle{Irit at trac 2018}. Association for Computational
  Linguistics (ACL).
\newblock


\bibitem[Ramiandrisoa and Mothe(2020a)]%
        {ramiandrisoa2020aggression}
\bibfield{author}{\bibinfo{person}{Faneva Ramiandrisoa} {and}
  \bibinfo{person}{Josiane Mothe}.} \bibinfo{year}{2020}\natexlab{a}.
\newblock \showarticletitle{Aggression identification in social media: a
  transfer learning based approach}. In \bibinfo{booktitle}{\emph{Second
  Workshop on Trolling, Aggression and Cyberbullying}}.
  \bibinfo{pages}{26--31}.
\newblock


\bibitem[Ramiandrisoa and Mothe(2020b)]%
        {ramiandrisoa2020irit}
\bibfield{author}{\bibinfo{person}{Faneva Ramiandrisoa} {and}
  \bibinfo{person}{Josiane Mothe}.} \bibinfo{year}{2020}\natexlab{b}.
\newblock \showarticletitle{Irit at trac 2020}. In
  \bibinfo{booktitle}{\emph{Second Workshop on Trolling, Aggression and
  Cyberbullying}}. \bibinfo{pages}{49--54}.
\newblock


\bibitem[Ranasinghe and Zampieri(2021)]%
        {ranasinghe2021multilingual}
\bibfield{author}{\bibinfo{person}{Tharindu Ranasinghe} {and}
  \bibinfo{person}{Marcos Zampieri}.} \bibinfo{year}{2021}\natexlab{}.
\newblock \showarticletitle{Multilingual offensive language identification for
  low-resource languages}.
\newblock \bibinfo{journal}{\emph{Transactions on Asian and Low-Resource
  Language Information Processing}} \bibinfo{volume}{21}, \bibinfo{number}{1}
  (\bibinfo{year}{2021}), \bibinfo{pages}{1--13}.
\newblock


\bibitem[Rawat et~al\mbox{.}(2023)]%
        {rawat2023modelling}
\bibfield{author}{\bibinfo{person}{Akash Rawat}, \bibinfo{person}{Nazia Nafis},
  \bibinfo{person}{Dnyaneshwar Bhadane}, \bibinfo{person}{Diptesh Kanojia},
  {and} \bibinfo{person}{Rudra Murthy}.} \bibinfo{year}{2023}\natexlab{}.
\newblock \showarticletitle{Modelling Political Aggression on Social Media
  Platforms}. In \bibinfo{booktitle}{\emph{Proceedings of the 13th Workshop on
  Computational Approaches to Subjectivity, Sentiment, \& Social Media
  Analysis}}. \bibinfo{pages}{497--510}.
\newblock


\bibitem[Reynolds et~al\mbox{.}(2011)]%
        {Reynolds2011}
\bibfield{author}{\bibinfo{person}{Kelly Reynolds}, \bibinfo{person}{April
  Kontostathis}, {and} \bibinfo{person}{Lynne Edwards}.}
  \bibinfo{year}{2011}\natexlab{}.
\newblock \showarticletitle{Using machine learning to detect cyberbullying}.
\newblock \bibinfo{journal}{\emph{Proceedings - 10th International Conference
  on Machine Learning and Applications, ICMLA 2011}}  \bibinfo{volume}{2}
  (\bibinfo{year}{2011}), \bibinfo{pages}{241--244}.
\newblock
\showISBNx{9780769546070}
\urldef\tempurl%
\url{https://doi.org/10.1109/ICMLA.2011.152}
\showDOI{\tempurl}


\bibitem[Risch and Krestel(2018)]%
        {risch2018aggression}
\bibfield{author}{\bibinfo{person}{Julian Risch} {and} \bibinfo{person}{Ralf
  Krestel}.} \bibinfo{year}{2018}\natexlab{}.
\newblock \showarticletitle{Aggression identification using deep learning and
  data augmentation}. In \bibinfo{booktitle}{\emph{Proceedings of the first
  workshop on trolling, aggression and cyberbullying (TRAC-2018)}}.
  \bibinfo{pages}{150--158}.
\newblock


\bibitem[Risch and Krestel(2020)]%
        {rischbaggingBM}
\bibfield{author}{\bibinfo{person}{Julian Risch} {and} \bibinfo{person}{Ralf
  Krestel}.} \bibinfo{year}{2020}\natexlab{}.
\newblock \showarticletitle{Bagging BERT models for robust aggression
  identification}. In \bibinfo{booktitle}{\emph{Proceedings of the Second
  Workshop on Trolling, Aggression and Cyberbullying}}.
  \bibinfo{pages}{55--61}.
\newblock


\bibitem[Roy et~al\mbox{.}(2018)]%
        {roy2018ensemble}
\bibfield{author}{\bibinfo{person}{Arjun Roy}, \bibinfo{person}{Prashant
  Kapil}, \bibinfo{person}{Kingshuk Basak}, {and} \bibinfo{person}{Asif
  Ekbal}.} \bibinfo{year}{2018}\natexlab{}.
\newblock \showarticletitle{An ensemble approach for aggression identification
  in English and Hindi text}. In \bibinfo{booktitle}{\emph{Proceedings of the
  first workshop on trolling, aggression and cyberbullying (TRAC-2018)}}.
  \bibinfo{pages}{66--73}.
\newblock


\bibitem[Roy et~al\mbox{.}(2023)]%
        {roy-etal-2023-probing}
\bibfield{author}{\bibinfo{person}{Sarthak Roy}, \bibinfo{person}{Ashish
  Harshvardhan}, \bibinfo{person}{Animesh Mukherjee}, {and}
  \bibinfo{person}{Punyajoy Saha}.} \bibinfo{year}{2023}\natexlab{}.
\newblock \showarticletitle{Probing {LLM}s for hate speech detection: strengths
  and vulnerabilities}. In \bibinfo{booktitle}{\emph{Findings of the
  Association for Computational Linguistics: EMNLP 2023}},
  \bibfield{editor}{\bibinfo{person}{Houda Bouamor}, \bibinfo{person}{Juan
  Pino}, {and} \bibinfo{person}{Kalika Bali}} (Eds.).
  \bibinfo{publisher}{Association for Computational Linguistics},
  \bibinfo{address}{Singapore}, \bibinfo{pages}{6116--6128}.
\newblock
\urldef\tempurl%
\url{https://doi.org/10.18653/v1/2023.findings-emnlp.407}
\showDOI{\tempurl}


\bibitem[Runions and Bak(2015)]%
        {runions2015online}
\bibfield{author}{\bibinfo{person}{Kevin~C Runions} {and}
  \bibinfo{person}{Michal Bak}.} \bibinfo{year}{2015}\natexlab{}.
\newblock \showarticletitle{Online moral disengagement, cyberbullying, and
  cyber-aggression}.
\newblock \bibinfo{journal}{\emph{Cyberpsychology, Behavior, and Social
  Networking}} \bibinfo{volume}{18}, \bibinfo{number}{7}
  (\bibinfo{year}{2015}), \bibinfo{pages}{400--405}.
\newblock


\bibitem[Sadiq et~al\mbox{.}(2021)]%
        {sadiq2021aggression}
\bibfield{author}{\bibinfo{person}{Saima Sadiq}, \bibinfo{person}{Arif
  Mehmood}, \bibinfo{person}{Saleem Ullah}, \bibinfo{person}{Maqsood Ahmad},
  \bibinfo{person}{Gyu~Sang Choi}, {and} \bibinfo{person}{Byung-Won On}.}
  \bibinfo{year}{2021}\natexlab{}.
\newblock \showarticletitle{Aggression detection through deep neural model on
  twitter}.
\newblock \bibinfo{journal}{\emph{Future Generation Computer Systems}}
  \bibinfo{volume}{114} (\bibinfo{year}{2021}), \bibinfo{pages}{120--129}.
\newblock


\bibitem[Saeed et~al\mbox{.}(2023)]%
        {saeed2023detection}
\bibfield{author}{\bibinfo{person}{Ramsha Saeed}, \bibinfo{person}{Hammad
  Afzal}, \bibinfo{person}{Sadaf~Abdul Rauf}, {and} \bibinfo{person}{Naima
  Iltaf}.} \bibinfo{year}{2023}\natexlab{}.
\newblock \showarticletitle{Detection of Offensive Language and its Severity
  for Low Resource Language}.
\newblock \bibinfo{journal}{\emph{ACM Transactions on Asian and Low-Resource
  Language Information Processing}} (\bibinfo{year}{2023}).
\newblock


\bibitem[Saha et~al\mbox{.}(2023)]%
        {saha2023}
\bibfield{author}{\bibinfo{person}{Punyajoy Saha}, \bibinfo{person}{Mithun
  Das}, \bibinfo{person}{Binny Mathew}, {and} \bibinfo{person}{Animesh
  Mukherjee}.} \bibinfo{year}{2023}\natexlab{}.
\newblock \showarticletitle{Hate Speech: Detection, Mitigation and Beyond}. In
  \bibinfo{booktitle}{\emph{Proceedings of the Sixteenth ACM International
  Conference on Web Search and Data Mining}} (Singapore, Singapore)
  \emph{(\bibinfo{series}{WSDM '23})}. \bibinfo{publisher}{Association for
  Computing Machinery}, \bibinfo{address}{New York, NY, USA},
  \bibinfo{pages}{1232–1235}.
\newblock
\showISBNx{9781450394079}
\urldef\tempurl%
\url{https://doi.org/10.1145/3539597.3572721}
\showDOI{\tempurl}


\bibitem[Samghabadi et~al\mbox{.}(2018)]%
        {samghabadi2018ritual}
\bibfield{author}{\bibinfo{person}{Niloofar~Safi Samghabadi},
  \bibinfo{person}{Deepthi Mave}, \bibinfo{person}{Sudipta Kar}, {and}
  \bibinfo{person}{Thamar Solorio}.} \bibinfo{year}{2018}\natexlab{}.
\newblock \showarticletitle{Ritual-uh at trac 2018 shared task: aggression
  identification}.
\newblock \bibinfo{journal}{\emph{arXiv preprint arXiv:1807.11712}}
  (\bibinfo{year}{2018}).
\newblock


\bibitem[Samghabadi et~al\mbox{.}(2020)]%
        {Samghabadi2020AggressionAM}
\bibfield{author}{\bibinfo{person}{Niloofar~Safi Samghabadi},
  \bibinfo{person}{Parth Patwa}, \bibinfo{person}{Pykl Srinivas},
  \bibinfo{person}{Prerana Mukherjee}, \bibinfo{person}{Amitava Das}, {and}
  \bibinfo{person}{Thamar Solorio}.} \bibinfo{year}{2020}\natexlab{}.
\newblock \showarticletitle{Aggression and Misogyny Detection using BERT: A
  Multi-Task Approach}. In \bibinfo{booktitle}{\emph{Workshop on Trolling,
  Aggression and Cyberbullying}}.
\newblock


\bibitem[Sane et~al\mbox{.}({[n.\,d.]})]%
        {SaneCorpusCA}
\bibfield{author}{\bibinfo{person}{Koushik~Reddy Sane},
  \bibinfo{person}{Sai~Teja Kolla}, \bibinfo{person}{Sushmitha~Reddy Sane},
  {and} \bibinfo{person}{R. Mamidi}.} \bibinfo{year}{[n.\,d.]}\natexlab{}.
\newblock \showarticletitle{Corpus Creation and Baseline System for Aggression
  Detection in Telugu-English Code-Mixed Social Media Data}.
\newblock
\urldef\tempurl%
\url{https://api.semanticscholar.org/CorpusID:247080842}
\showURL{%
\tempurl}


\bibitem[Sap et~al\mbox{.}(2014)]%
        {sap2014developing}
\bibfield{author}{\bibinfo{person}{Maarten Sap}, \bibinfo{person}{Gregory
  Park}, \bibinfo{person}{Johannes Eichstaedt}, \bibinfo{person}{Margaret
  Kern}, \bibinfo{person}{David Stillwell}, \bibinfo{person}{Michal Kosinski},
  \bibinfo{person}{Lyle Ungar}, {and} \bibinfo{person}{H~Andrew Schwartz}.}
  \bibinfo{year}{2014}\natexlab{}.
\newblock \showarticletitle{Developing age and gender predictive lexica over
  social media}. In \bibinfo{booktitle}{\emph{Proceedings of the 2014
  conference on empirical methods in natural language processing (EMNLP)}}.
  \bibinfo{pages}{1146--1151}.
\newblock


\bibitem[Sengupta et~al\mbox{.}(2022)]%
        {sengupta2022does}
\bibfield{author}{\bibinfo{person}{Ayan Sengupta},
  \bibinfo{person}{Sourabh~Kumar Bhattacharjee}, \bibinfo{person}{Md~Shad
  Akhtar}, {and} \bibinfo{person}{Tanmoy Chakraborty}.}
  \bibinfo{year}{2022}\natexlab{}.
\newblock \showarticletitle{Does aggression lead to hate? Detecting and
  reasoning offensive traits in hinglish code-mixed texts}.
\newblock \bibinfo{journal}{\emph{Neurocomputing}}  \bibinfo{volume}{488}
  (\bibinfo{year}{2022}), \bibinfo{pages}{598--617}.
\newblock


\bibitem[Sharif and Hoque(2022)]%
        {sharif2022tackling}
\bibfield{author}{\bibinfo{person}{Omar Sharif} {and}
  \bibinfo{person}{Mohammed~Moshiul Hoque}.} \bibinfo{year}{2022}\natexlab{}.
\newblock \showarticletitle{Tackling cyber-aggression: Identification and
  fine-grained categorization of aggressive texts on social media using
  weighted ensemble of transformers}.
\newblock \bibinfo{journal}{\emph{Neurocomputing}}  \bibinfo{volume}{490}
  (\bibinfo{year}{2022}), \bibinfo{pages}{462--481}.
\newblock


\bibitem[Shrivastava et~al\mbox{.}(2021)]%
        {Shrivastava2021EnhancingAD}
\bibfield{author}{\bibinfo{person}{Adarsh Shrivastava},
  \bibinfo{person}{Rushikesh Pupale}, {and} \bibinfo{person}{Pradeep Singh}.}
  \bibinfo{year}{2021}\natexlab{}.
\newblock \showarticletitle{Enhancing Aggression Detection using GPT-2 based
  Data Balancing Technique}.
\newblock \bibinfo{journal}{\emph{2021 5th International Conference on
  Intelligent Computing and Control Systems (ICICCS)}} (\bibinfo{year}{2021}),
  \bibinfo{pages}{1345--1350}.
\newblock


\bibitem[Shulginov et~al\mbox{.}(2021)]%
        {shulginov2021automatic}
\bibfield{author}{\bibinfo{person}{V.~A. Shulginov}, \bibinfo{person}{Ramis~Z
  Mustafin}, {and} \bibinfo{person}{A.~A. Tillabaeva}.}
  \bibinfo{year}{2021}\natexlab{}.
\newblock \showarticletitle{Automatic Detection of Implicit Aggression in
  Russian Social Media Comments}.
\newblock
\urldef\tempurl%
\url{https://api.semanticscholar.org/CorpusID:240077467}
\showURL{%
\tempurl}


\bibitem[Si et~al\mbox{.}(2019)]%
        {si2019aggression}
\bibfield{author}{\bibinfo{person}{Shukrity Si}, \bibinfo{person}{Anisha
  Datta}, \bibinfo{person}{Somnath Banerjee}, {and}
  \bibinfo{person}{Sudip~Kumar Naskar}.} \bibinfo{year}{2019}\natexlab{}.
\newblock \showarticletitle{Aggression detection on multilingual social media
  text}. In \bibinfo{booktitle}{\emph{2019 10th International Conference on
  Computing, Communication and Networking Technologies (ICCCNT)}}. IEEE,
  \bibinfo{pages}{1--5}.
\newblock


\bibitem[Singh et~al\mbox{.}(2018)]%
        {singh2018aggression}
\bibfield{author}{\bibinfo{person}{Vinay Singh}, \bibinfo{person}{Aman
  Varshney}, \bibinfo{person}{Syed~Sarfaraz Akhtar}, \bibinfo{person}{Deepanshu
  Vijay}, {and} \bibinfo{person}{Manish Shrivastava}.}
  \bibinfo{year}{2018}\natexlab{}.
\newblock \showarticletitle{Aggression detection on social media text using
  deep neural networks}. In \bibinfo{booktitle}{\emph{Proceedings of the 2nd
  workshop on abusive language online (ALW2)}}. \bibinfo{pages}{43--50}.
\newblock


\bibitem[Smith et~al\mbox{.}(2008)]%
        {smith2008cyberbullying}
\bibfield{author}{\bibinfo{person}{Peter~K Smith}, \bibinfo{person}{Jess
  Mahdavi}, \bibinfo{person}{Manuel Carvalho}, \bibinfo{person}{Sonja Fisher},
  \bibinfo{person}{Shanette Russell}, {and} \bibinfo{person}{Neil Tippett}.}
  \bibinfo{year}{2008}\natexlab{}.
\newblock \showarticletitle{Cyberbullying: Its nature and impact in secondary
  school pupils}.
\newblock \bibinfo{journal}{\emph{Journal of child psychology and psychiatry}}
  \bibinfo{volume}{49}, \bibinfo{number}{4} (\bibinfo{year}{2008}),
  \bibinfo{pages}{376--385}.
\newblock


\bibitem[Sobkin and Fedotova(2021)]%
        {sobkin2021adolescents}
\bibfield{author}{\bibinfo{person}{Vladimir~S Sobkin} {and}
  \bibinfo{person}{Aleksandra~V Fedotova}.} \bibinfo{year}{2021}\natexlab{}.
\newblock \showarticletitle{Adolescents on social media: Aggression and
  cyberbullying}.
\newblock \bibinfo{journal}{\emph{Psychology in Russia: State of the Art}}
  \bibinfo{volume}{14}, \bibinfo{number}{4} (\bibinfo{year}{2021}),
  \bibinfo{pages}{55--70}.
\newblock


\bibitem[Srivastava et~al\mbox{.}(2020)]%
        {Srivastava2020AMV}
\bibfield{author}{\bibinfo{person}{Arjit Srivastava}, \bibinfo{person}{Avijit
  Vajpayee}, \bibinfo{person}{Syed~Sarfaraz Akhtar}, \bibinfo{person}{Naman
  Jain}, \bibinfo{person}{Vinay Singh}, {and} \bibinfo{person}{Manish
  Shrivastava}.} \bibinfo{year}{2020}\natexlab{}.
\newblock \showarticletitle{A Multi-Dimensional View of Aggression when voicing
  Opinion}. In \bibinfo{booktitle}{\emph{Workshop on Trolling, Aggression and
  Cyberbullying}}.
\newblock
\urldef\tempurl%
\url{https://api.semanticscholar.org/CorpusID:218974314}
\showURL{%
\tempurl}


\bibitem[Srivastava and Khurana(2019)]%
        {Srivastava2019DetectingAA}
\bibfield{author}{\bibinfo{person}{Saurabh Srivastava} {and}
  \bibinfo{person}{Prerna Khurana}.} \bibinfo{year}{2019}\natexlab{}.
\newblock \showarticletitle{Detecting Aggression and Toxicity using a Multi
  Dimension Capsule Network}.
\newblock \bibinfo{journal}{\emph{Proceedings of the Third Workshop on Abusive
  Language Online}} (\bibinfo{year}{2019}).
\newblock


\bibitem[Srivastava et~al\mbox{.}(2018)]%
        {srivastava2018identifying}
\bibfield{author}{\bibinfo{person}{Saurabh Srivastava}, \bibinfo{person}{Prerna
  Khurana}, {and} \bibinfo{person}{Vartika Tewari}.}
  \bibinfo{year}{2018}\natexlab{}.
\newblock \showarticletitle{Identifying aggression and toxicity in comments
  using capsule network}. In \bibinfo{booktitle}{\emph{Proceedings of the first
  workshop on trolling, aggression and cyberbullying (TRAC-2018)}}.
  \bibinfo{pages}{98--105}.
\newblock


\bibitem[Stickland and Murray(2019)]%
        {stickland2019bert}
\bibfield{author}{\bibinfo{person}{Asa~Cooper Stickland} {and}
  \bibinfo{person}{Iain Murray}.} \bibinfo{year}{2019}\natexlab{}.
\newblock \showarticletitle{Bert and pals: Projected attention layers for
  efficient adaptation in multi-task learning}. In
  \bibinfo{booktitle}{\emph{International Conference on Machine Learning}}.
  PMLR, \bibinfo{pages}{5986--5995}.
\newblock


\bibitem[Ta et~al\mbox{.}(2022a)]%
        {ta2022gan}
\bibfield{author}{\bibinfo{person}{Hoang~Thang Ta}, \bibinfo{person}{Abu
  Bakar~Siddiqur Rahman}, \bibinfo{person}{Lotfollah Najjar}, {and}
  \bibinfo{person}{Alexander Gelbukh}.} \bibinfo{year}{2022}\natexlab{a}.
\newblock \showarticletitle{GAN-BERT: Adversarial Learning for Detection of
  Aggressive and Violent Incidents from Social Media}. In
  \bibinfo{booktitle}{\emph{Proceedings of the Iberian Languages Evaluation
  Forum (IberLEF 2022), CEUR Workshop Proceedings. CEUR-WS. org}}.
\newblock


\bibitem[Ta et~al\mbox{.}(2022b)]%
        {ta2022multi}
\bibfield{author}{\bibinfo{person}{Hoang~Thang Ta}, \bibinfo{person}{Abu
  Bakar~Siddiqur Rahman}, \bibinfo{person}{Lotfollah Najjar}, {and}
  \bibinfo{person}{AF Gelbukh}.} \bibinfo{year}{2022}\natexlab{b}.
\newblock \showarticletitle{Multi-Task Learning for Detection of Aggressive and
  Violent Incidents from Social Media}. In
  \bibinfo{booktitle}{\emph{Proceedings of the 2022 Iberian Languages
  Evaluation Forum, IberLEF}}.
\newblock


\bibitem[Tanase et~al\mbox{.}(2020)]%
        {Tanase2020DetectingAI}
\bibfield{author}{\bibinfo{person}{Mircea-Adrian Tanase},
  \bibinfo{person}{George-Eduard Zaharia}, \bibinfo{person}{Dumitru-Clementin
  Cercel}, {and} \bibinfo{person}{M. Dascalu}.}
  \bibinfo{year}{2020}\natexlab{}.
\newblock \showarticletitle{Detecting Aggressiveness in Mexican Spanish Social
  Media Content by Fine-Tuning Transformer-Based Models}. In
  \bibinfo{booktitle}{\emph{IberLEF@SEPLN}}.
\newblock


\bibitem[Tawalbeh et~al\mbox{.}(2020)]%
        {tawalbeh2020saja}
\bibfield{author}{\bibinfo{person}{Saja Tawalbeh}, \bibinfo{person}{Mahmoud
  Hammad}, {and} \bibinfo{person}{AL-Smadi Mohammad}.}
  \bibinfo{year}{2020}\natexlab{}.
\newblock \showarticletitle{Saja at trac 2020 shared task: Transfer learning
  for aggressive identification with xgboost}. In
  \bibinfo{booktitle}{\emph{Proceedings of the Second Workshop on Trolling,
  Aggression and Cyberbullying}}. \bibinfo{pages}{99--105}.
\newblock


\bibitem[Terizi et~al\mbox{.}(2021)]%
        {terizi2021modeling}
\bibfield{author}{\bibinfo{person}{Chrysoula Terizi}, \bibinfo{person}{Despoina
  Chatzakou}, \bibinfo{person}{Evaggelia Pitoura}, \bibinfo{person}{Panayiotis
  Tsaparas}, {and} \bibinfo{person}{Nicolas Kourtellis}.}
  \bibinfo{year}{2021}\natexlab{}.
\newblock \showarticletitle{Modeling aggression propagation on social media}.
\newblock \bibinfo{journal}{\emph{Online Social Networks and Media}}
  \bibinfo{volume}{24} (\bibinfo{year}{2021}), \bibinfo{pages}{100137}.
\newblock


\bibitem[Tommasel et~al\mbox{.}(2019)]%
        {Tommasel2019AnES}
\bibfield{author}{\bibinfo{person}{Antonela Tommasel}, \bibinfo{person}{J.
  Rodriguez}, {and} \bibinfo{person}{Daniela Godoy}.}
  \bibinfo{year}{2019}\natexlab{}.
\newblock \showarticletitle{An experimental study on feature engineering and
  learning approaches for aggression detection in social media}.
\newblock \bibinfo{journal}{\emph{Inteligencia Artif.}}  \bibinfo{volume}{22}
  (\bibinfo{year}{2019}), \bibinfo{pages}{81--100}.
\newblock


\bibitem[Tommasel et~al\mbox{.}(2018)]%
        {tommasel2018textual}
\bibfield{author}{\bibinfo{person}{Antonela Tommasel},
  \bibinfo{person}{Juan~Manuel Rodriguez}, {and} \bibinfo{person}{Daniela
  Godoy}.} \bibinfo{year}{2018}\natexlab{}.
\newblock \showarticletitle{Textual aggression detection through deep
  learning}. In \bibinfo{booktitle}{\emph{Proceedings of the first workshop on
  trolling, aggression and cyberbullying (TRAC-2018)}}.
  \bibinfo{pages}{177--187}.
\newblock


\bibitem[Tonglin et~al\mbox{.}(2018)]%
        {tonglin2018effect}
\bibfield{author}{\bibinfo{person}{JIN Tonglin}, \bibinfo{person}{LU Guizhi},
  \bibinfo{person}{ZHANG Lu}, \bibinfo{person}{WU Yuntena}, {and}
  \bibinfo{person}{JIN Xiangzhong}.} \bibinfo{year}{2018}\natexlab{}.
\newblock \showarticletitle{The effect of violent exposure on online aggressive
  behavior of college students: The role of ruminative responses and internet
  moral.}
\newblock \bibinfo{journal}{\emph{Acta Psychologica Sinica}}
  (\bibinfo{year}{2018}).
\newblock


\bibitem[Tonja et~al\mbox{.}(2022)]%
        {tonja2022detection}
\bibfield{author}{\bibinfo{person}{Atnafu~Lambebo Tonja},
  \bibinfo{person}{Muhammad Arif}, \bibinfo{person}{Olga Kolesnikova},
  \bibinfo{person}{Alexander Gelbukh}, {and} \bibinfo{person}{Grigori
  Sidorov}.} \bibinfo{year}{2022}\natexlab{}.
\newblock \showarticletitle{Detection of aggressive and violent incidents from
  social media in spanish using pretrained language model}. In
  \bibinfo{booktitle}{\emph{Proceedings of the Iberian Languages Evaluation
  Forum (IberLEF 2022), CEUR Workshop Proceedings. CEUR-WS. org}}.
\newblock


\bibitem[Torregrosa et~al\mbox{.}(2022)]%
        {torregrosa2022mixed}
\bibfield{author}{\bibinfo{person}{Javier Torregrosa}, \bibinfo{person}{Sergio
  D’Antonio-Maceiras}, \bibinfo{person}{Guillermo Villar-Rodr{\'\i}guez},
  \bibinfo{person}{Amir Hussain}, \bibinfo{person}{Erik Cambria}, {and}
  \bibinfo{person}{David Camacho}.} \bibinfo{year}{2022}\natexlab{}.
\newblock \showarticletitle{A mixed approach for aggressive political discourse
  analysis on Twitter}.
\newblock \bibinfo{journal}{\emph{Cognitive computation}}
  (\bibinfo{year}{2022}), \bibinfo{pages}{1--26}.
\newblock


\bibitem[Vaswani et~al\mbox{.}(2017)]%
        {vaswani2017}
\bibfield{author}{\bibinfo{person}{Ashish Vaswani}, \bibinfo{person}{Noam
  Shazeer}, \bibinfo{person}{Niki Parmar}, \bibinfo{person}{Jakob Uszkoreit},
  \bibinfo{person}{Llion Jones}, \bibinfo{person}{Aidan~N. Gomez},
  \bibinfo{person}{\L{}ukasz Kaiser}, {and} \bibinfo{person}{Illia
  Polosukhin}.} \bibinfo{year}{2017}\natexlab{}.
\newblock \showarticletitle{Attention is all you need}. In
  \bibinfo{booktitle}{\emph{Proceedings of the 31st International Conference on
  Neural Information Processing Systems}} (Long Beach, California, USA)
  \emph{(\bibinfo{series}{NIPS'17})}. \bibinfo{publisher}{Curran Associates
  Inc.}, \bibinfo{address}{Red Hook, NY, USA}, \bibinfo{pages}{6000–6010}.
\newblock
\showISBNx{9781510860964}


\bibitem[Vazsonyi et~al\mbox{.}(2012)]%
        {vazsonyi2012cyberbullying}
\bibfield{author}{\bibinfo{person}{Alexander~T Vazsonyi}, \bibinfo{person}{Hana
  Machackova}, \bibinfo{person}{Anna Sevcikova}, \bibinfo{person}{David
  Smahel}, {and} \bibinfo{person}{Alena Cerna}.}
  \bibinfo{year}{2012}\natexlab{}.
\newblock \showarticletitle{Cyberbullying in context: Direct and indirect
  effects by low self-control across 25 European countries}.
\newblock \bibinfo{journal}{\emph{European Journal of Developmental
  Psychology}} \bibinfo{volume}{9}, \bibinfo{number}{2} (\bibinfo{year}{2012}),
  \bibinfo{pages}{210--227}.
\newblock


\bibitem[Vladimirou et~al\mbox{.}(2021)]%
        {vladimirou2021aggressive}
\bibfield{author}{\bibinfo{person}{Dimitra Vladimirou},
  \bibinfo{person}{Juliane House}, {and} \bibinfo{person}{D{\'a}niel~Z
  K{\'a}d{\'a}r}.} \bibinfo{year}{2021}\natexlab{}.
\newblock \showarticletitle{Aggressive complaining on social media: the case
  of\# MuckyMerton}.
\newblock \bibinfo{journal}{\emph{Journal of Pragmatics}}
  \bibinfo{volume}{177} (\bibinfo{year}{2021}), \bibinfo{pages}{51--64}.
\newblock


\bibitem[V{\"o}llink et~al\mbox{.}(2015)]%
        {vollink2015cyberbullying}
\bibfield{author}{\bibinfo{person}{Trijntje V{\"o}llink},
  \bibinfo{person}{Francine Dehue}, {and} \bibinfo{person}{Conor Mc~Guckin}.}
  \bibinfo{year}{2015}\natexlab{}.
\newblock \showarticletitle{Cyberbullying: From theory to intervention}.
\newblock  (\bibinfo{year}{2015}).
\newblock


\bibitem[Walther(2008)]%
        {walther2008social}
\bibfield{author}{\bibinfo{person}{Joseph~B Walther}.}
  \bibinfo{year}{2008}\natexlab{}.
\newblock \showarticletitle{Social information processing theory}.
\newblock \bibinfo{journal}{\emph{Engaging theories in interpersonal
  communication: Multiple perspectives}}  \bibinfo{volume}{391}
  (\bibinfo{year}{2008}).
\newblock


\bibitem[Waseem(2016)]%
        {waseem2016you}
\bibfield{author}{\bibinfo{person}{Zeerak Waseem}.}
  \bibinfo{year}{2016}\natexlab{}.
\newblock \showarticletitle{Are you a racist or am i seeing things? annotator
  influence on hate speech detection on twitter}. In
  \bibinfo{booktitle}{\emph{Proceedings of the first workshop on NLP and
  computational social science}}. \bibinfo{pages}{138--142}.
\newblock


\bibitem[Weingartner and Stahel(2019)]%
        {weingartner2019online}
\bibfield{author}{\bibinfo{person}{Sebastian Weingartner} {and}
  \bibinfo{person}{Lea Stahel}.} \bibinfo{year}{2019}\natexlab{}.
\newblock \showarticletitle{Online aggression from a sociological perspective:
  An integrative view on determinants and possible countermeasures}. In
  \bibinfo{booktitle}{\emph{Proceedings of the Third Workshop on Abusive
  Language Online}}. \bibinfo{pages}{181--187}.
\newblock


\bibitem[Wohlin(2014)]%
        {Wohlin2014}
\bibfield{author}{\bibinfo{person}{Claes Wohlin}.}
  \bibinfo{year}{2014}\natexlab{}.
\newblock \showarticletitle{Guidelines for snowballing in systematic literature
  studies and a replication in software engineering}. In
  \bibinfo{booktitle}{\emph{Proceedings of the 18th International Conference on
  Evaluation and Assessment in Software Engineering}} (London, England, United
  Kingdom) \emph{(\bibinfo{series}{EASE '14})}. \bibinfo{publisher}{Association
  for Computing Machinery}, \bibinfo{address}{New York, NY, USA}, Article
  \bibinfo{articleno}{38}, \bibinfo{numpages}{10}~pages.
\newblock
\showISBNx{9781450324762}
\urldef\tempurl%
\url{https://doi.org/10.1145/2601248.2601268}
\showDOI{\tempurl}


\bibitem[Wong et~al\mbox{.}(2022)]%
        {wong2022association}
\bibfield{author}{\bibinfo{person}{Natalie Wong}, \bibinfo{person}{Takuya
  Yanagida}, \bibinfo{person}{Christiane Spiel}, {and} \bibinfo{person}{Daniel
  Graf}.} \bibinfo{year}{2022}\natexlab{}.
\newblock \showarticletitle{The association between appetitive aggression and
  social media addiction mediated by cyberbullying: the moderating role of
  inclusive norms}.
\newblock \bibinfo{journal}{\emph{International journal of environmental
  research and public health}} \bibinfo{volume}{19}, \bibinfo{number}{16}
  (\bibinfo{year}{2022}), \bibinfo{pages}{9956}.
\newblock


\bibitem[Wright(2020)]%
        {wright2020role}
\bibfield{author}{\bibinfo{person}{Michelle~F Wright}.}
  \bibinfo{year}{2020}\natexlab{}.
\newblock \showarticletitle{The role of technologies, behaviors, gender, and
  gender stereotype traits in adolescents’ cyber aggression}.
\newblock \bibinfo{journal}{\emph{Journal of interpersonal violence}}
  \bibinfo{volume}{35}, \bibinfo{number}{7-8} (\bibinfo{year}{2020}),
  \bibinfo{pages}{1719--1738}.
\newblock


\bibitem[Wu and Dredze(2020)]%
        {wu-dredze-2020-languages}
\bibfield{author}{\bibinfo{person}{Shijie Wu} {and} \bibinfo{person}{Mark
  Dredze}.} \bibinfo{year}{2020}\natexlab{}.
\newblock \showarticletitle{Are All Languages Created Equal in Multilingual
  {BERT}?}. In \bibinfo{booktitle}{\emph{Proceedings of the 5th Workshop on
  Representation Learning for NLP}},
  \bibfield{editor}{\bibinfo{person}{Spandana Gella}, \bibinfo{person}{Johannes
  Welbl}, \bibinfo{person}{Marek Rei}, \bibinfo{person}{Fabio Petroni},
  \bibinfo{person}{Patrick Lewis}, \bibinfo{person}{Emma Strubell},
  \bibinfo{person}{Minjoon Seo}, {and} \bibinfo{person}{Hannaneh Hajishirzi}}
  (Eds.). \bibinfo{publisher}{Association for Computational Linguistics},
  \bibinfo{address}{Online}, \bibinfo{pages}{120--130}.
\newblock
\urldef\tempurl%
\url{https://doi.org/10.18653/v1/2020.repl4nlp-1.16}
\showDOI{\tempurl}


\bibitem[Zeng et~al\mbox{.}(2023)]%
        {zeng2023effects}
\bibfield{author}{\bibinfo{person}{Ke Zeng}, \bibinfo{person}{Feizhen Cao},
  \bibinfo{person}{Yajun Wu}, \bibinfo{person}{Manhua Zhang}, {and}
  \bibinfo{person}{Xinfang Ding}.} \bibinfo{year}{2023}\natexlab{}.
\newblock \showarticletitle{Effects of interpretation bias modification on
  hostile attribution bias and reactive cyber-aggression in Chinese
  adolescents: a randomized controlled trial}.
\newblock \bibinfo{journal}{\emph{Current Psychology}} (\bibinfo{year}{2023}),
  \bibinfo{pages}{1--14}.
\newblock


\bibitem[Zhao and Gao(2012)]%
        {zhao2012reliability}
\bibfield{author}{\bibinfo{person}{F Zhao} {and} \bibinfo{person}{WB Gao}.}
  \bibinfo{year}{2012}\natexlab{}.
\newblock \showarticletitle{Reliability and validity of the adolescent online
  aggressive behavior scale}.
\newblock \bibinfo{journal}{\emph{Chinese Mental Health Journal}}
  \bibinfo{volume}{26}, \bibinfo{number}{6} (\bibinfo{year}{2012}),
  \bibinfo{pages}{439--444}.
\newblock


\bibitem[Zhao et~al\mbox{.}(2021)]%
        {zhao2021comparative}
\bibfield{author}{\bibinfo{person}{Zhixue Zhao}, \bibinfo{person}{Ziqi Zhang},
  {and} \bibinfo{person}{Frank Hopfgartner}.} \bibinfo{year}{2021}\natexlab{}.
\newblock \showarticletitle{A comparative study of using pre-trained language
  models for toxic comment classification}. In
  \bibinfo{booktitle}{\emph{Companion Proceedings of the Web Conference 2021}}.
  \bibinfo{pages}{500--507}.
\newblock


\bibitem[Zimmerman and Ybarra(2016)]%
        {zimmerman2016online}
\bibfield{author}{\bibinfo{person}{Adam~G Zimmerman} {and}
  \bibinfo{person}{Gabriel~J Ybarra}.} \bibinfo{year}{2016}\natexlab{}.
\newblock \showarticletitle{Online aggression: The influences of anonymity and
  social modeling.}
\newblock \bibinfo{journal}{\emph{Psychology of Popular Media Culture}}
  \bibinfo{volume}{5}, \bibinfo{number}{2} (\bibinfo{year}{2016}),
  \bibinfo{pages}{181}.
\newblock


\end{thebibliography}

\vspace*{-0.3\baselineskip}
\appendix
\section{Abbreviations}
\label{abbreviations}
\vspace*{-0.3\baselineskip}
{
\scriptsize
\setlength\LTleft{0pt}
\setlength\LTright{0pt}
\begin{longtable}[c]{llll}
\endhead

Acc & Accuracy & m-BERT & Multilingual BERT \\
AG & Aggression & MAG & Mediam aggression \\
ARF & Adaptive Random Forest & ML & Machine learning \\
AUC & Area Under Curve & Mn & Maithili \\
Avg & Average & MTL & Multi Tasking Learning \\
BanglaBERT & BERT pre-trained on Bangla & MultiCNNPooling & \begin{tabular}[c]{@{}l@{}}Multi-Convolutional Neural \\ Network with Pooling\end{tabular} \\
BERT & \begin{tabular}[c]{@{}l@{}}Bidirectional Encoder \\ Representations from \\ Transformers\end{tabular} & N & No \\
BFFO & Binary Firefly Optimization & NAG & Non-aggression \\
BiGRU & Bidirectional GRU & NB & Naive Bayes \\
BiLSTM & Bidirectional LSTM & NER & Named Entity Recognition \\
Bn & Bangla & NetzDG & Network Enforcement Act \\
CAG & Covert Aggression & NKB & Network Knowledge Base \\
Char & Character & NLP & Natural Language Processing \\
CK & Cohen’s kappa & OAG & Overt Aggression \\
CM & Code-mixed & OneR & One Rule \\
CNN & Convolutional Neural Networks & OSM & Online Social Media \\
COVID-19 & COronaVIrus Disease of 2019 & P & Precision \\
CS & Computer Science & Ph. No. & Phone Number \\
DA & Data Augmentation & POFMA & \begin{tabular}[c]{@{}l@{}}Protection from Online \\ Falsehoods and Manipulation \\ Act\end{tabular} \\
DistilBERT & Distilled versions of BERT & POS & Part-of-speech \\
DistilBETO & \begin{tabular}[c]{@{}l@{}}DistilBERT pre-trained \\ on Spanish\end{tabular} & PSMU & Problematic Social Media \\
DNN & Deep Neural Network & R & Recall \\
En & English & RBF & Radial Basis Function \\
Es & Spanish & RF & Random Forest \\
Fb & Facebook & RNN & Recurrent Neural Network \\
FK & Fleiss’ kappa & RoBERTa & Robustly Optimized BERT \\
GAM & General Aggression Model & Ru & Russian \\
GAN & Generative Adversarial Learning & RuBERT & BERT pre-trained on Russian \\
GBM & Gradient Boosting Classifier & SIPT & \begin{tabular}[c]{@{}l@{}}Social Information Processing \\ Theory\end{tabular} \\
GEN & Gendered & SLR & Streaming Logistic Regression \\
Glove & Global Vector & SMO & \begin{tabular}[c]{@{}l@{}}Sequential Minimal \\ Optimization\end{tabular} \\
GPT-2 & \begin{tabular}[c]{@{}l@{}}Generative Pretrained \\ Transformer 2\end{tabular} & SVM & Support Vector Machine \\
GRU & Gated Recurrent Unit & SOTA & State-Of-The-Art \\
HAG & High aggression & Tel & Telegram \\
HAL & Hierarchical attention network & TF-IDF & \begin{tabular}[c]{@{}l@{}}Term Frequency-Inverse \\ Document Frequency\end{tabular} \\
Hi & Hindi & TTR & Type-Token Ratio \\
Hing-BERT & \begin{tabular}[c]{@{}l@{}}BERT pre-trained on code-mixed \\ Hindi-English\end{tabular} & Tw & Twitter \\
HT & Hoeffding Tree & URL & Uniform Resource Locator \\
IAA & Inter-Annotator Agreement & USE & Universal Sentence Encoder \\
Insta & Instagram & VGG16 & Visual Geometry Group 16 \\
It & Italian & W-F1 & Weighted F1 Score \\
KA & Krippendorff’s alpha & XGBoost & eXtreme Gradient Boosting \\
LIWC & \begin{tabular}[c]{@{}l@{}}Linguistic Inquiry and \\ Word Count\end{tabular} & XLMRoBERTa & \begin{tabular}[c]{@{}l@{}}Cross-lingual Language Model \\ RoBERTa\end{tabular} \\
LR & Logistic Regression & Y & Yes \\
LSTM & Long Short-Term Memory & Yt & YouTube \\
\end{longtable}
}

\end{document}